\documentclass{article}

\usepackage[preprint]{neurips_2024}

\usepackage[utf8]{inputenc} 
\usepackage[T1]{fontenc}    
\usepackage{hyperref}       
\usepackage{url}            
\usepackage{booktabs}       
\usepackage{amsfonts}       
\usepackage{nicefrac}       
\usepackage{microtype}      
\usepackage{xcolor}         
\usepackage{amsmath}
\usepackage{dsfont}
\usepackage{mathtools}
\usepackage{graphicx}
\usepackage{subcaption}
\usepackage{svg}
\usepackage{adjustbox}
\usepackage{rotating}
\usepackage{multirow}
\usepackage{floatrow}
\usepackage{caption}
\usepackage[skip=0.5ex]{subcaption}
\newcommand{\stopgradient}{\mathrlap{\nabla}\hspace{0.5pt}/}

\title{LP-3DGS: Learning to Prune 3D Gaussian Splatting}

\author{%
  Zhaoliang Zhang \\
  Johns Hopkins University\\
  Baltimore, MD 21218 \\
  \texttt{zzhan288@jh.edu} \\
  \And
  Tianchen Song \\
  Johns Hopkins University \\
  Baltimore, MD 21218 \\
  \texttt{tsong15@jh.edu} \\
  \And
  Yongjae Lee \\
  Johns Hopkins University \\
  Baltimore, MD 21218 \\
  \texttt{ylee236@jh.edu} \\
  \And
Li Yang \\
  University of North Carolina at Charlotte \\
  Charlotte, NC 28223 \\
  \texttt{lyang50@uncc.edu} \\
  \And
  Cheng Peng \\
  Johns Hopkins University \\
  Baltimore, MD 21218 \\
  \texttt{cpeng26@jhu.edu} \\
  \And
  Rama Chellappa \\
  Johns Hopkins University \\
  Baltimore, MD 21218 \\
  \texttt{rchella4@jhu.edu} \\
  \And
  Deliang Fan \\
  Johns Hopkins University \\
  Baltimore, MD 21218 \\
  \texttt{dfan10@jhu.edu} \\
}

\begin{document}

\maketitle

\begin{abstract}
  Recently, 3D Gaussian Splatting (3DGS) has become one of the mainstream methodologies for novel view synthesis (NVS) due to its high quality and fast rendering speed. However, as a point-based scene representation, 3DGS potentially generates a large number of Gaussians to fit the scene, leading to high memory usage. 
  Improvements that have been proposed require either an empirical and preset pruning ratio or importance score threshold to prune the point cloud. Such hyperparamter requires multiple rounds of training to optimize and achieve the maximum pruning ratio, while maintaining the rendering quality for each scene.  
  In this work, we propose learning-to-prune 3DGS (LP-3DGS), where a trainable binary mask is applied to the importance score that can find optimal pruning ratio automatically. Instead of using the traditional straight-through estimator (STE) method to approximate the binary mask gradient, we redesign the masking function to leverage the Gumbel-Sigmoid method, making it differentiable and compatible with the existing training process of 3DGS. Extensive experiments have shown that LP-3DGS consistently produces a good balance that is both efficient and high quality.

\end{abstract}

\section{Introduction}
Novel view synthesis (NVS) takes images and their corresponding camera poses as input and seek to render new images from different camera poses after 3D scene reconstruction. Neural Radiance Fields (NeRF) (\citet{mildenhall2021nerf}) uses multi-layer perceptron (MLP) to implicitly represent the scene, fetching the transparency and color of a point from the MLPs. NeRF has gained considerable attention in the NVS community due to its simple implementation and excellent performance. However, in order to get a point in the space, NeRF needs to perform an MLP inference. Each pixel rendered requires processing a ray and there are many sample points on each ray. Consequently, rendering an image requires a large amount of MLP inference operations. Thus, rendering speed becomes a major drawback of NeRF method.

Besides NeRF, explicit 3D representations are also widely used. Compared with NeRF, the advantage of point-based scene representation is that modern GPU rendering is well supported, enabling fast render speed. 3D Gaussian Splatting (3DGS) (\citet{kerbl20233d}) achieves good quality and high rendering speed, making it a hot topic in the community. 3DGS uses 3D Gaussian models with color parameters to fit the scene and develops a training framework to optimize the model parameters. However, the number of points required to reconstruct the scene is huge, usually in the millions. In practice, each point has dozens of floating point parameters, which makes 3DGS a memory-intensive method. 

\begin{figure}[htbp]
    \centering

    \makebox[0.32\textwidth]{\textbf{PSNR $\uparrow$}}
    \makebox[0.32\textwidth]{\textbf{SSIM $\uparrow$}}
    \makebox[0.32\textwidth]{\textbf{LPIPS $\downarrow$}}
    
    \begin{subfigure}[b]{0.32\textwidth}
        \includegraphics[width=\textwidth]{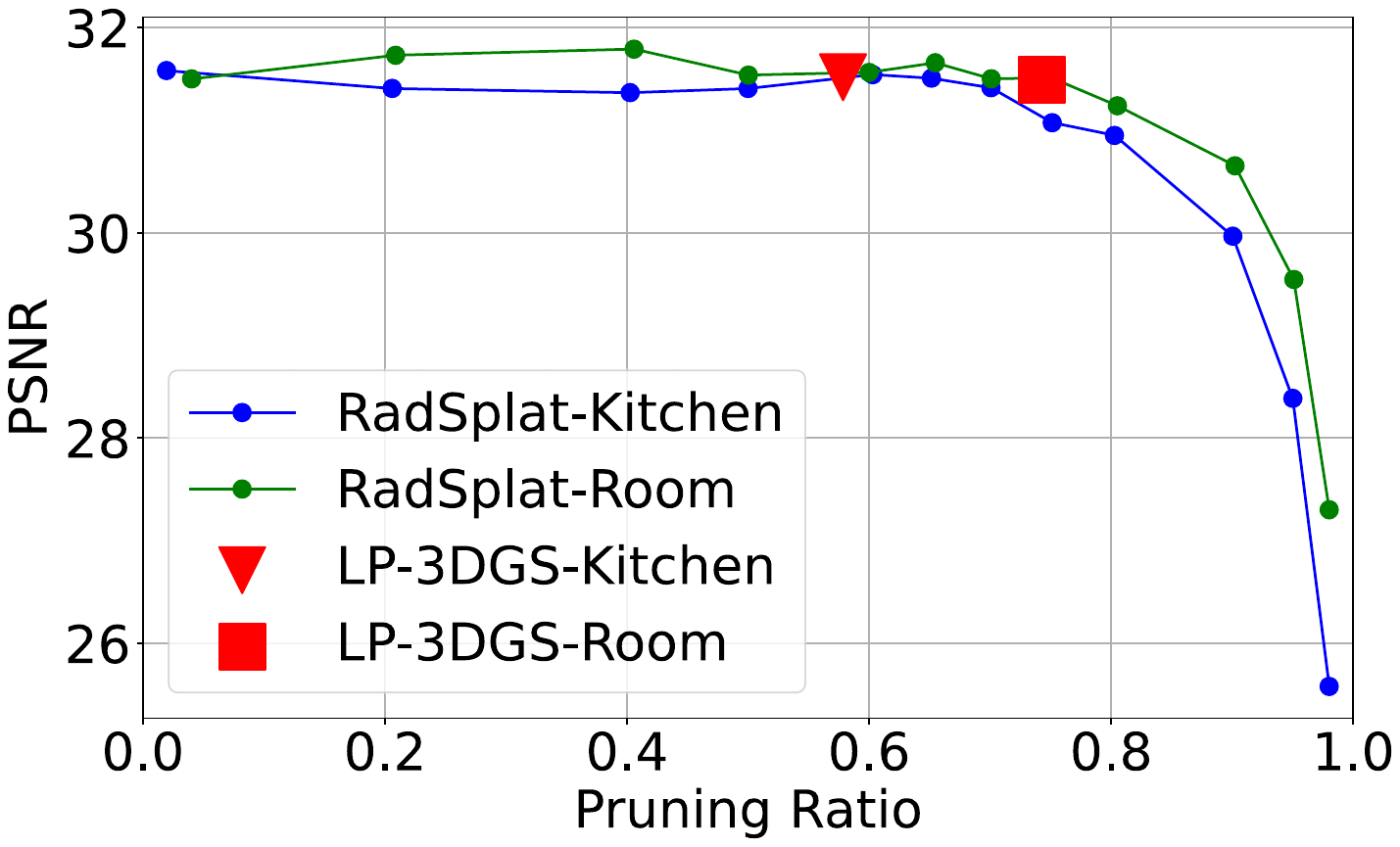}
    \end{subfigure}
    \begin{subfigure}[b]{0.32\textwidth}
        \includegraphics[width=\textwidth]{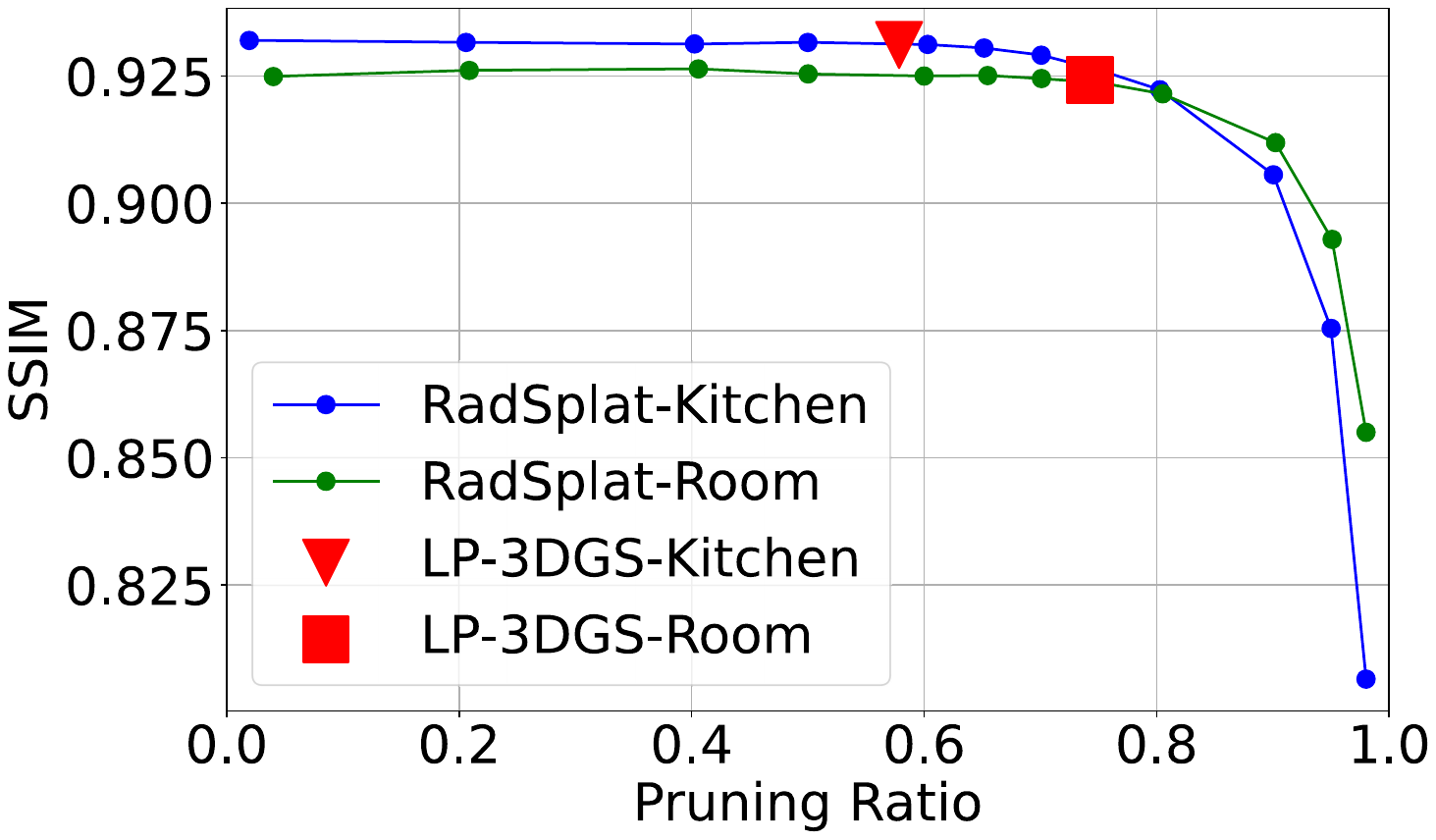}
    \end{subfigure}
    \begin{subfigure}[b]{0.32\textwidth}
        \includegraphics[width=\textwidth]{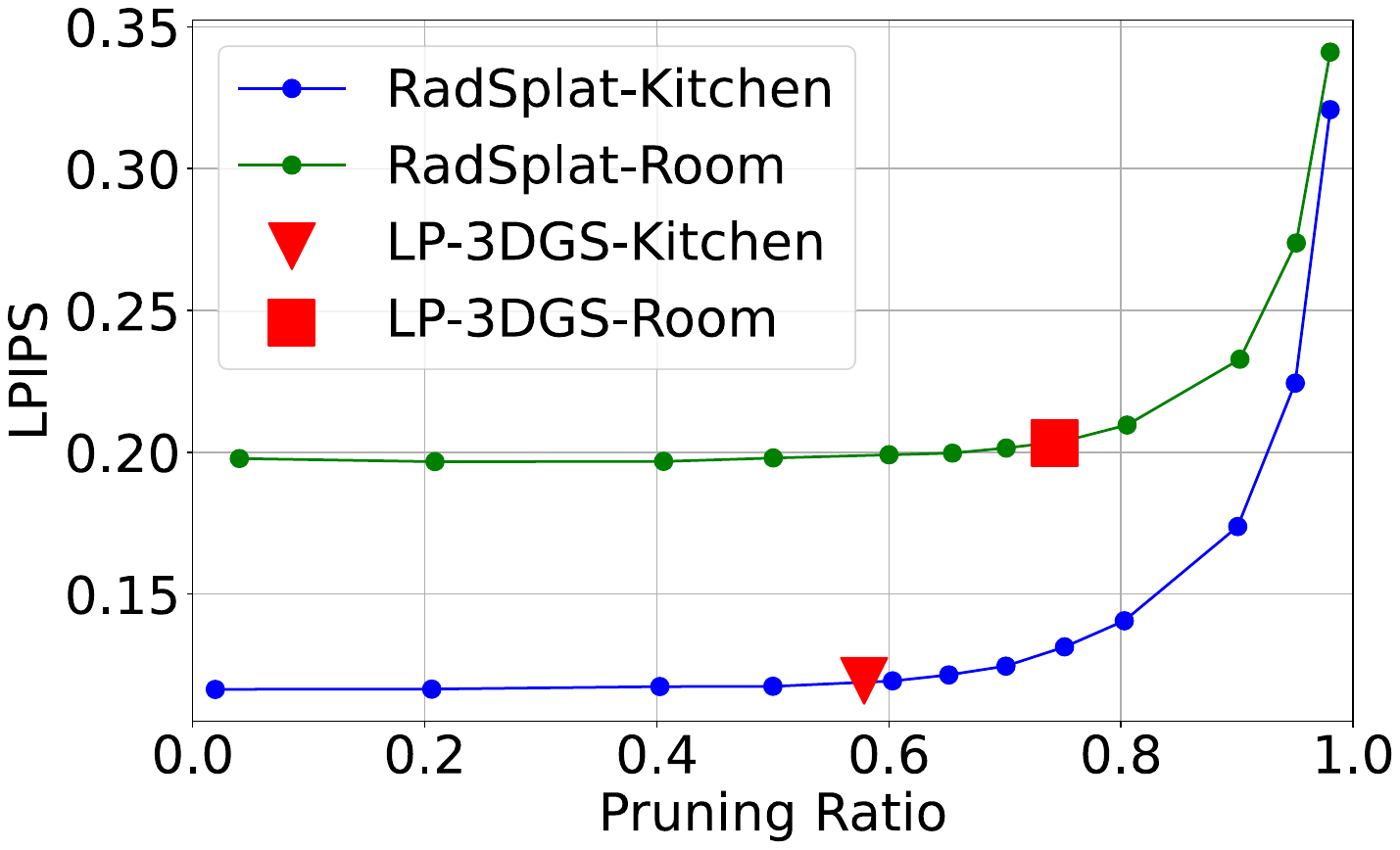}
    \end{subfigure}

    \caption{The performance changes with the pruning ratio of RadSplat on the MipNeRF360 dataset \textit{Kitchen} and \textit{Room} scenes are shown in \textcolor{blue}{blue} and \textcolor{green}{green} lines, respectively. \textcolor{red}{Red} triangles and squares represent the results of LP-3DGS on the importance score of RadSplat. \textbf{LP-3DGS is able to find the optimal pruning ratio in one training session instead of requiring dozens of attempts to find the best hyperparameter}.}
    \label{fig:performance_with_ratio}
\end{figure}

Some recent works have tried to mitigate this problem by pruning the points, such as LightGaussian (\citet{fan2023lightgaussian}), RadSplat (\citet{niemeyer2024radsplat}), and Mini-Splatting (\citet{fang2024mini}). These methods follow a similar pruning approach through defining an \textit{importance score} for each Gaussian point and prune the points with such importance score below than a preset empirical threshold. However, a major drawback of these methods is that such preset threshold needs to be manually tuned through multiple rounds of training process to identify the optimal pruning ratio to minimize the number of Gaussian points while keeping the rendering quality. To make it even worse, such optimal number of points may vary depending on different scenes, which requires manual pruning ratio searching for each scene. 
For example, the blue and red lines in Figure \ref{fig:performance_with_ratio} show the rendering quality of \textit{Kitchen} and \textit{Room} scenes, respectively, in MipNeRF360 scenes (\citet{barron2022mip}), with sweeping 12 different pruning ratios (i.e., 12 rounds of training) following the prior RadSplat (\citet{niemeyer2024radsplat}) and Mini-Splatting (\citet{fang2024mini}) method. It could be clearly seen that a smaller pruning ratio will not hamper the rendering quality, and the rendering quality will start to decrease with much more aggressive pruning ratios. While, both scenes exist an optimal pruning ratio region that could maximize the pruning ratio and maintain the rendering quality. It could also be seen that such optimal pruning ratio region is different for these two scenes. 


In this paper, we propose a learning-to-prune 3DGS (LP-3DGS) methodology where a trainable mask is applied to the importance score. Note that, it is compatible with different types of importance scores defined in prior works. Instead of a preset threshold to determine the 3DGS model pruning ratio, as shown by the red triangle symbol in the Figure \ref{fig:performance_with_ratio}, our method aims to integrate with existing 3DGS model training process to learn the optimal pruning ratio for minimizing the model size while maintaining the rendering quality. Since the traditional hard threshold based binary masking function is not differentiable, a recent prior work, Compact3D (\citet{lee2023compact}), leverages the popular straight through estimator (STE) (\citet{bengio2013estimating}) to bypass the mask gradient for adaption to the backpropagation process. Such method inevitably leads to non-optimal learned pruning ratio. In contrast, in our work, we propose to redesign the masking function leveraging the Gumbel-Sigmoid activation function to make the whole masking function differentiable and integrate with the existing training process of 3DGS. As a result, our LP-3DGS could minimize the number of Gaussian points automatically for each scene with only one-time training.    

In summary, the technical contributions of our work are:
\begin{itemize}
    \item To address the effortful 3DGS optimal pruning ratio tuning, we propose a learning-to-prune 3DGS (LP-3DGS) methodology that leverages the differentiable Gumbel-Sigmoid activation function to embed a trainable mask with different types of existing importance scores designed for pruning redundant Gaussian points. As a result, instead of fixed model size, LP-3DGS could learn an optimal Gaussian point size for individual scene with only one-time training.   
    \item We conducted comprehensive experiments on state-of-the-art (SoTA) 3D scene datasets, including MipNeRF360 (\citet{barron2022mip}), NeRF-Synthetic (\citet{mildenhall2021nerf}), and Tanks \& Temples (\citet{knapitsch2017tanks}). We compared our method with SoTA pruning methods such as RadSplat (\citet{niemeyer2024radsplat}), Mini-Splatting (\citet{fang2024mini}), and Compact3D (\citet{lee2023compact}). The experimental results show that our method can enable the model to learn the optimal pruning ratio and that our trainable mask method performs better than the STE mask.
\end{itemize}

\section{Related Work}
\paragraph{Neural radiance fields (NeRFs)} NeRFs (\citet{mildenhall2021nerf}) targets to represent the scene in multilayer perceptrons (MLPs) based on multi-view image inputs, enabling high-quality novel view synthesis. Due to its advancement, numerous follow-up works improved it in either rendering quality (\citet{barron2021mip,barron2022mip}) or efficiency(\citet{muller2022instant,chen2022tensorf,fridovich2022plenoxels}).

Although NeRF models demonstrate impressive rendering capabilities across numerous benchmarks, and considerable efforts have been made to enhance training and inference efficiency, they typically still face challenges in achieving fast training and real-time rendering.


\paragraph{Radiance Field Based On Points.} In addition to implicit representations, several works have focused on volumetric point-based methods for 3D presentation (\citet{gross2011point}). Inspired by neural network concepts, (\citet{aliev2020neural}) introduced a neural point-based approach to streamline the construction process. Point-NeRF (\citet{ding2024point}) further applied points for volumetric representation, enhancing the effectiveness of point-based methods in radiance field modeling.

\paragraph{Gaussian Splatting} 3D Gaussian Splatting (3DGS) (\citet{kerbl20233d}) represents a significant advancement in novel view synthesis, utilizing 3D Gaussians as primitives to explicitly represent scenes. This approach achieves state-of-the-art rendering quality and speed while maintaining relatively short training time. A series of 
 methods have been introduced to improve the rendering quality through using regularization for better optimization, including depth map (\citet{chung2023depth,li2024dngaussian}), surface alignment (\citet{guedon2023sugar, li2024geogaussian}) and rendered image frequency (\citet{zhang2024fregs}). However, the extensive number of Gaussians required for scene representation often results in a model that is too large for efficient storage. Recent research has focused on compression methods to enhance the efficiency of this representation. Notably, several studies (\citet{fan2023lightgaussian,fang2024mini,niemeyer2024radsplat}) have proposed using predefined scores as pruning criteria to keep Gaussians that significantly contribute to rendering quality. Compact3D (\citet{lee2023compact}) introduces a method that applies a trainable mask on scale and opacity to each Gaussian and utilizes a straight-through estimator (\citet{bengio2013estimating}) for gradient updates. LightGaussian (\citet{fan2023lightgaussian}) employs knowledge distillation to reduce the dimension of spherical harmonics. Additionally, (\citet{fan2023lightgaussian,lee2023compact}) also explored quantization techniques to further compress model storage. The previously proposed pruning methods primarily rely on predefined scores to determine the importance of each Gaussian. These approaches present two main challenges: first, whether the criteria accurately reflect the importance of the Gaussians, and second, the need for a manually selected pruning threshold to decide the level of pruning. In this work, we address these issues by introducing a trainable mask activated by a Gumbel-sigmoid function, applied to the scores derived from prior methods  or directly to the scale and opacity of each Gaussian for more flexibility. Our approach automatically identifies an optimal balance between the pruning ratio and rendering quality, eliminating the need to test on various pruning ratios.

\section{Methodology}
The conventional pruning methods leveraging predefined importance score require pruning ratio as a manually tuned parameter to reduce the size of Gaussian points in 3DGS.
To seeking for the optimal pruning ratio, these methods may need to perform multiple rounds of training for each individual scene, which is inefficient. Thus motivated, we propose a \textbf{learning-to-prune 3DGS (LP-3DGS)} algorithm which learns a binary mask to determine the optimal pruning ratio for each scene automatically. Importantly, the proposed LP-3DGS is compatible with different types of pruning importance score. In this section, we will: 1) introduce the preliminary of the original 3DGS and recap different importance metrics for pruning that are proposed by prior works, and 2) present the proposed learning-to-prune 3DGS algorithm.

\subsection{3DGS Background}
\paragraph{3DGS Parameters} 3DGS is an explicit point-based 3D representation that uses Gaussian points to model the scene. Each point has the following attributes: position $\bf{p} \in \mathbb{R}^3$, opacity $\sigma \in [0, 1]$, scale in 3D $\bf{s} \in \mathbb{R}^3$, rotation presented by 4D quaternions $\bf{q} \in \mathbb{R}^4$ and forth-degree spherical harmonics (SH) coefficients $\bf{k} \in \mathbb{R}^{48}$. In summary, one gaussian point has 59 parameters. The center point $X$ of a Gaussian model is denoted by $\bf{p}$ and covariance matrix $\Sigma$ is denoted by $\bf{s}$ and $\bf{q}$. The SH coefficients model the color as viewed from different directions. The parameters of the Gaussians are optimized through gradient backpropagation of the loss between the rendered images and the ground truth.

\paragraph{Rendering on 3DGS} In order to render an image, the first step is projecting the Gaussians to 2D camera plane by world to camera transform matrix $W$ and Jacobian $J$ of affine approximation of the projective transform. The covariance matrix of projected 2D Gaussian is
\begin{equation}
    \Sigma^{'} = JW\Sigma W^TJ^T
\end{equation}
The projected Gaussians would be rendered as splat (\citet{botsch2005high}), the color of one pixel could be rendered as
\begin{equation}
    \bf{c}_i = \sum_{j=1}^{N} \cdot \bf{c}_j \cdot \alpha_j \cdot T_j \cdot \bf{G}^{2D}_j
\end{equation}
Where $i$ is the pixel index, $j$ is the Gaussian index and $N$ is the number of the Gaussians in the ray. $c_j$ is the color of the Gaussian calculated by SH coefficients, $\alpha_j = (1 - exp^{-\sigma_j \delta_j})$, $\sigma_j$ is the opacity of the point and $\delta_j$ is the interval between points. $T_j = \prod_{k=1}^{j-1}(1 - \alpha_k)$ is the transmittance from the start of rendering to this point. $\bf{G}^{2D}_j$ is the 2D Gaussian distribution. 

\paragraph{Adaptive Density Control of 3DGS} At the start of training, the Gaussians are initialized using Structure-from-Motion (SfM) sparse points. To make the Gaussians fit the scene better, 3DGS applies an adaptive density control strategy to adjust the number of Gaussians. Periodically, 3DGS will grow Gaussians in areas that are not well reconstructed, a process called "densification." Simultaneously, Gaussians with low opacity will be pruned. 

\subsection{Importance Metrics for Pruning}
A straightforward way to prune the Gaussians is by sorting them based on a defined importance score and then removing the less important ones. As a result, one of the main objective of prior 3DGS pruning works is to define the importance metric. 

RadSplat (\citet{niemeyer2024radsplat}) defines the importance score as the maximum contribution along all rays of the training images, written as
\begin{equation}
    \bf{S}_i = \max_{I_f\in \mathcal{I}_f, r\in I_f} \alpha_i^r \cdot T_i^r
\end{equation}
Where $\alpha_i^r \cdot T_i^r$ is the contribution of Gaussian $G_i$ along ray $r$. 
RadSplat performs pruning by applying a binary mask according to the importance score, where the mask value for Gaussian $G_i$ is
\begin{equation}
    m_i = m(\bf{S}_i) = \mathds{1}(\bf{S}_i < t_{prune})
\end{equation}
Where $t_{prune} \in [0,1]$ is the threshold of score magnitude for pruning, $\mathds{1} [\cdot]$ is the indicator function.

Another recent work, Mini-Splatting (\citet{fang2024mini}), uses the cumulative weight of the Gaussian as the importance score, which can be formulated as:
\begin{equation}
    \bf{S}_i = \sum_{j=1}^{K} \omega_{ij}
\end{equation}
Where K is the total number of rays intersected with Gaussian $G_i$, $\omega_{ij}$ is the color weight of Gaussian $G_i$ on the $j$-th ray. 

\begin{figure}[t]
  \centering
  \includegraphics[width=1.0\textwidth]{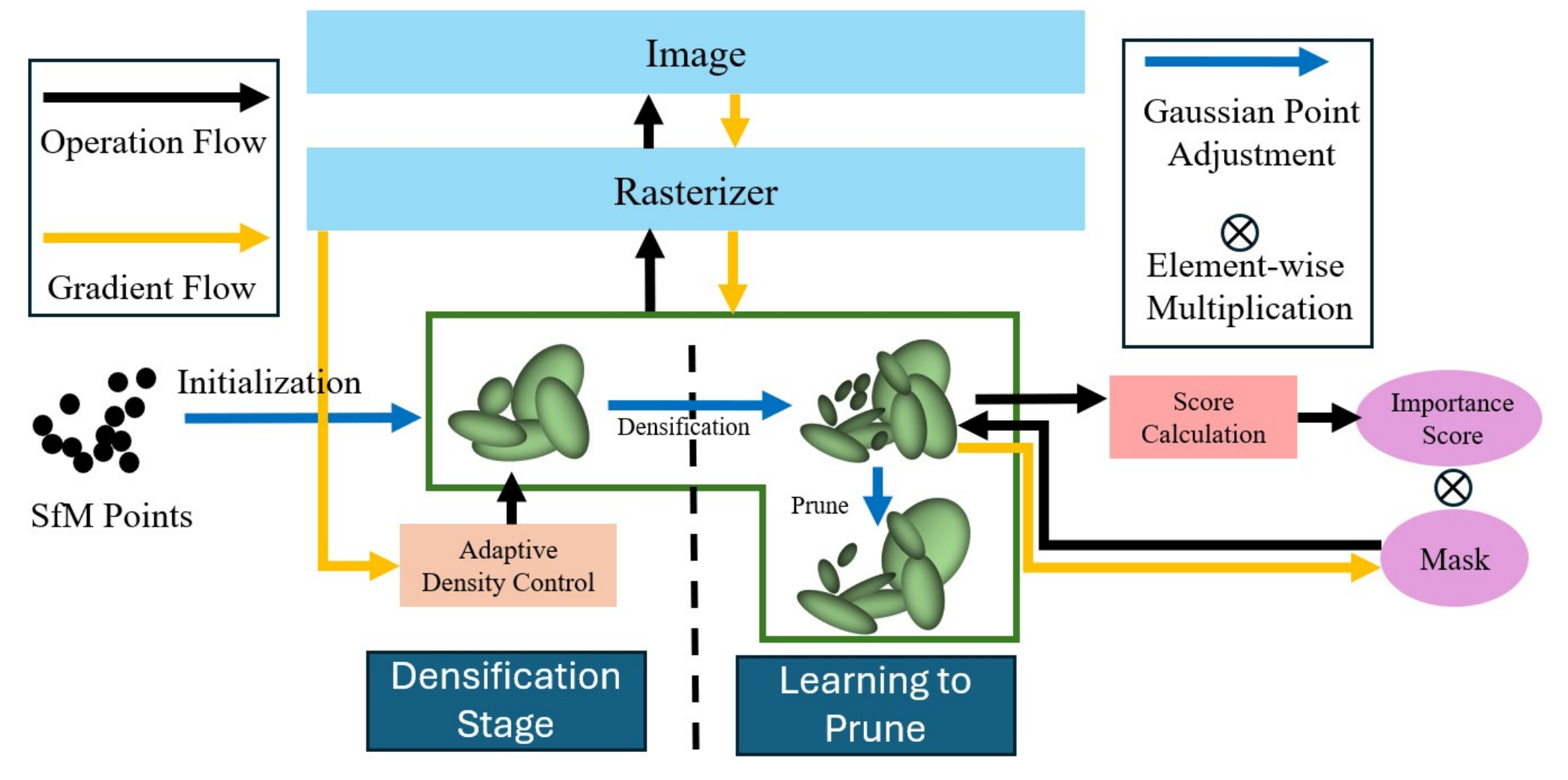}
  \caption{Overall learning process of the proposed LP-3DGS.}
  \label{fig:overview}
\end{figure}

\subsection{Learning-to-Prune 3DGS}


The overall LP-3DGS learning process is shown in the Figure \ref{fig:overview}. In general, it mainly can be divided into two stages:1) densification stage, and 2) learning-to-prune stage. Following the original 3DGS, densification stage applies an adaptive density control strategy to gradually increase the number of Gaussians. As revealed by prior pruning works~(\citet{lee2023compact,niemeyer2024radsplat}), 3DGS exists redundant Gaussians significantly. Subsequently, in the learning-to-prune stage, the proposed LP-3DGS learns a trainable mask upon  prior defined importance metric to compress the number of Gaussians with an optimal pruning ratio automatically.
Specifically, to learn a binary mask, we first initialize a real-value mask  $m_i$ for each point $i$, and then adopt the Gumbel-sigmoid technique to binarize the mask value differentially.

\begin{figure}[htbp]
    \centering
    \begin{subfigure}[b]{0.45\textwidth}
        \centering
        \includegraphics[width=\textwidth]{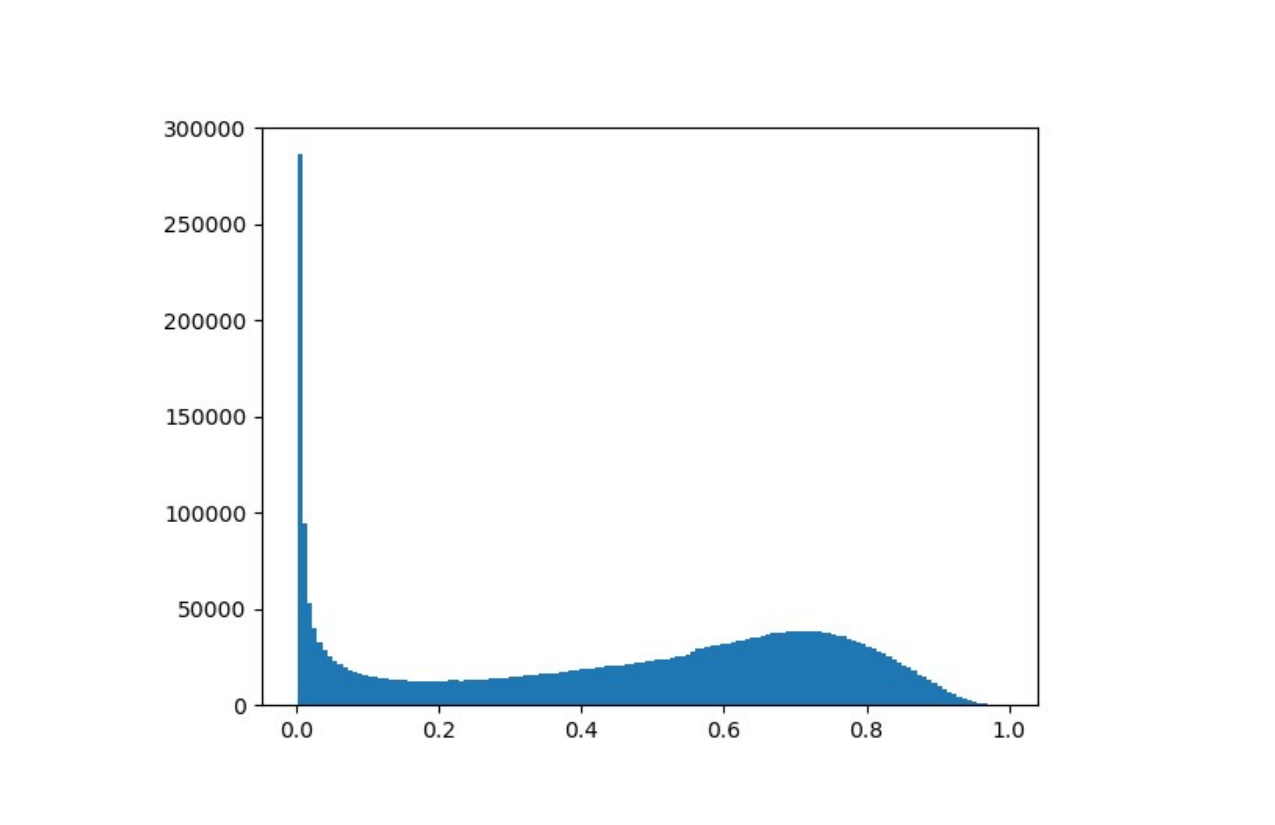}
        \caption{Mask values after Sigmoid activation}
        \label{fig:sigmoid}
    \end{subfigure}
    \hfill
    \begin{subfigure}[b]{0.45\textwidth}
        \centering
        \includegraphics[width=\textwidth]{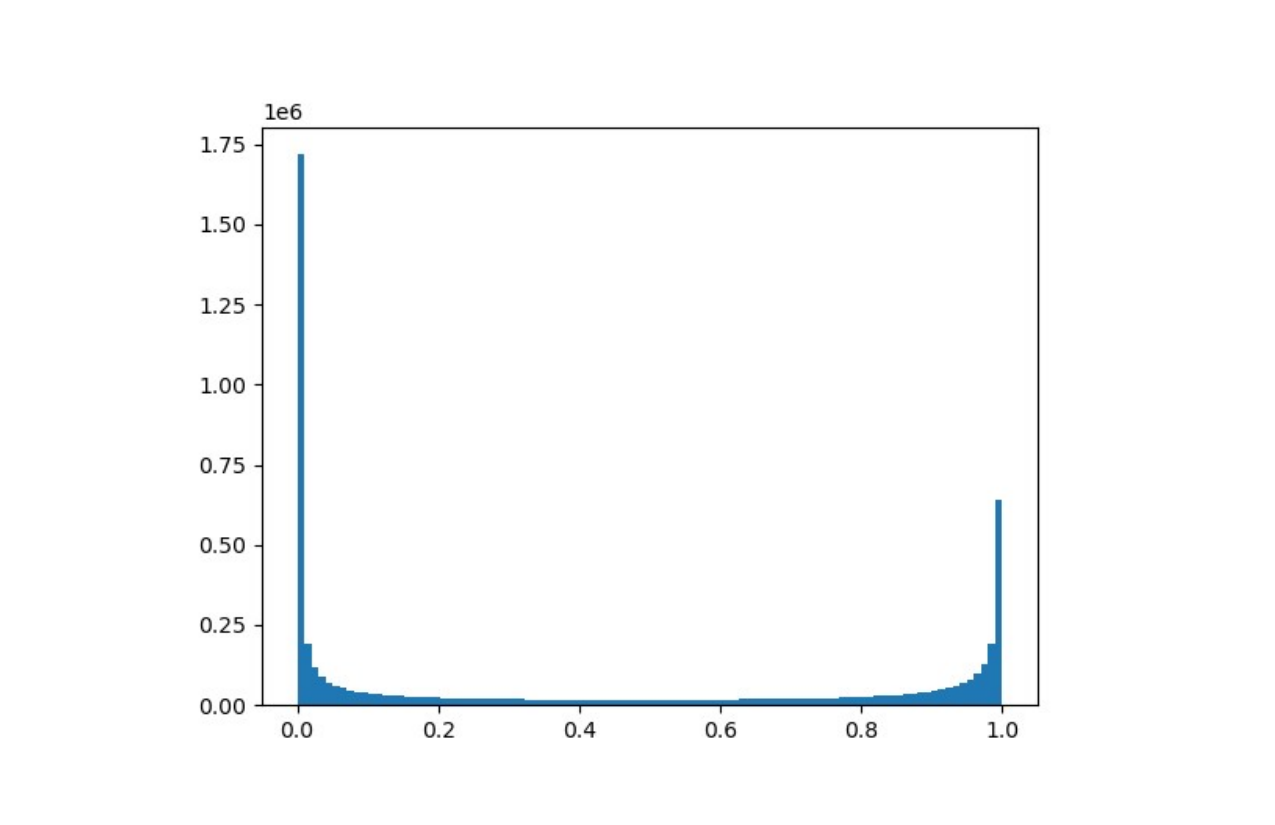}
        \caption{Mask values after Gumbel-Sigmoid activation}
        \label{fig:gumbel_sigmoid}
    \end{subfigure}
    \caption{Comparison between Sigmoid and Gumbel-Sigmoid. The Gumbel-Sigmoid function pushes the values closer to 0 or 1 and is a good approximation of a binarized mask.}
    \label{fig:activations}
\end{figure}

\paragraph{Gumbel-Sigmoid based Trainable Mask}
The binarization operation for real-value mask in pruning usually involves a hard threshold function, determining the binary mask should be 0 or 1. However, such hard threshold function is not differential during backpropagation. To solve this issue, popular straight
through estimator (STE) method (\citet{bengio2013estimating}) is widely used which skips the gradient of the threshold function during backpropagation. Such process may lead to a gap between trainable real-value mask and binary mask. As shown in Figure \ref{fig:activations}(a), the trainable mask values exist certain ratios cross the whole range from 0 to 1 after Sigmoid function, which could be inaccurate when further converting to binary mask via a hard threshold function. To better optimize the trainable mask towards binary values, we propose to apply Gumbel-Sigmoid function to learn the binary mask.

The Gumbel distribution is used to model the extreme value distribution and generate samples from the categorical distribution (\citet{gumbel1954statistical}). This property is then utilized to create the Gumbel-Softmax (\citet{jang2016categorical}), a differentiable categorical distribution sampling function. The sample of one category is given by:

\begin{equation}
y_i = \frac{\exp((\log(\pi_i) + g_i)/\tau)}{\sum_{j=1}^{k} \exp((\log(\pi_j) + g_j)/\tau)}
\end{equation}
Where $\tau$ is the input adjustment parameter, $g_i$ is sample from Gumbel distribution.
Inspired by the Gumbel-Softmax, we treat learning the binary mask of each point as a two-class category problem. Thus, we replace the Softmax function to Sigmoid function, referring to Gumbel-Sigmoid:
\begin{equation}
\label{equ:gumbel_sigmoid}
    gs(m) = \frac{exp((\log(m) + g_0)/\tau)}{exp((\log(m)+g_0)/T)+exp(g_1/\tau)} = \frac{1}{1+exp(-(\log(m) + g_0 - g_1)/\tau)}
\end{equation}

By using such Gumbel-Sigmoid function, the output value is either close to 0 or 1, as shown in Figure \ref{fig:activations}, and thus can be utilized as an approximation of a binary masking function. More importantly, this function remains differentiable, thus can be integrated during backpropagation. 

Moreover, to prune the selected Guassians practically according to the learned binary mask, we further apply the mask value on opacity, which can be mathematically formulated as
\begin{equation}
\label{equ:apply_mask}
    o_{im} = o_i * gs(m_i*S_i)
\end{equation}
where $S_i$ is the defined importance score of each Gaussian point. The closer the mask value is to 0, the less corresponding Gaussian point contributes to the rendering. 
In practice, after learning the trainable mask, a one-time pruning is applied to the corresponding Gaussian points with mask value of 0. 

\paragraph{Sparsity regularization} In order to compress the model as much as possible, we apply a L1 regularization term~(\citet{lee2023compact}) to encourage the trainable mask to be sparse, which can be formulated as:
\begin{equation}
    R_{mask} = \frac{1}{N} \sum_{i=1}^{N} \left| m_i \right|
\end{equation}
Upon that, the final loss function is defined as:
\begin{equation}
    L = (1-\lambda_{ssim})*L_{L1} + \lambda_{ssim}*L_{ssim} + \lambda_m*R_{mask}
\end{equation}
$L_{L1}$ is the L1 loss between rendered image and ground truth.
$L_{ssim}$ is the ssim loss. $\lambda_{ssim}$ and $\lambda_{m}$ are two coefficients.


Moreover,  
we find that the trainable mask can be effectively learned in just a few hundred iterations, compared to the thousands needed for the overall training process. In practice, the mask learning function is activated for only 500 iterations. Once the mask values are learned, we follow the 3DGS training setup to further fine-tune the pruned model, maintaining the same total number of training iterations.
The detailed hyper parameters are described in the later experiment section.

\section{Experiments}
\label{sec:experiments}
\subsection{Experimental Settings}
\label{sec:exp_settings}
\paragraph{Dataset and Baseline} 
We test our method on two most popular real-world datasets: the MipNeRF360 dataset (\citet{barron2022mip}), which contains 9 scenes, and the \textit{Train} and \textit{Truck} scenes from the Tanks \& Temples dataset (\citet{knapitsch2017tanks}). We also test on the NeRF-Synthetic dataset (\citet{mildenhall2021nerf}), which includes 8 synthetic scenes. In this section, we only list the results on  MipNeRF360 dataset, rest of them are listed in appendix \ref{appendix}. 
In this paper, we use the SoTA RadSplat (\citet{niemeyer2024radsplat}) and Mini-Splatting (\citet{fang2024mini}) as the baselines, which propose two different pruning importance scores. First, we test the performance of these two methods under different pruning ratios. Since neither method is open-sourced at the time of writing, we reproduced them based on the provided equations. For each pruning ratio, we calculate the corresponding threshold based on the magnitude of the importance scores and prune the Gaussians with scores below the threshold. Note that, each pruning ratio requires one round of training. 
We use peak signal-to-noise ratio (PSNR), structural similarity index measure (SSIM), and Learned Perceptual Image Patch Similarity (LPIPS) (\citet{zhang2018unreasonable}) as rendering evaluation metrics.

\paragraph{Implement Details}
The machine running the experiments has an AMD 5955WX processor and two Nvidia A6000 GPUs. It should be noted that our method does not support multi-GPU training. We ran different experiments simultaneously on two GPUs. We train each scene under every setting for 30,000 iterations and train the mask during iterations 19,500 to 20,000, updating the importance score every 20 iterations. The $\tau$ in Equation \ref{equ:gumbel_sigmoid} is 0.5 and the coefficient $\lambda_m$ of mask loss is 5e-4. 

\subsection{Experimental Results}
\label{sec:exp_results}
\begin{figure}[htbp]
    \centering

    \makebox[0.08\textwidth]{} 
    \makebox[0.29\textwidth]{\textbf{PSNR $\uparrow$}}
    \makebox[0.29\textwidth]{\textbf{SSIM $\uparrow$}}
    \makebox[0.29\textwidth]{\textbf{LPIPS $\downarrow$}}
    \par\medskip



    \begin{minipage}[b]{0.05\textwidth}
        \rotatebox[origin=t, y=2.8cm]{90}{\textbf{Kitchen}}
    \end{minipage}
    \begin{minipage}[b]{0.90\textwidth}
    \rotatebox[origin=t, y=1.3cm]{90}{\textbf{RadSplat}}
    \begin{subfigure}[b]{0.31\textwidth}
        \includegraphics[width=\textwidth]{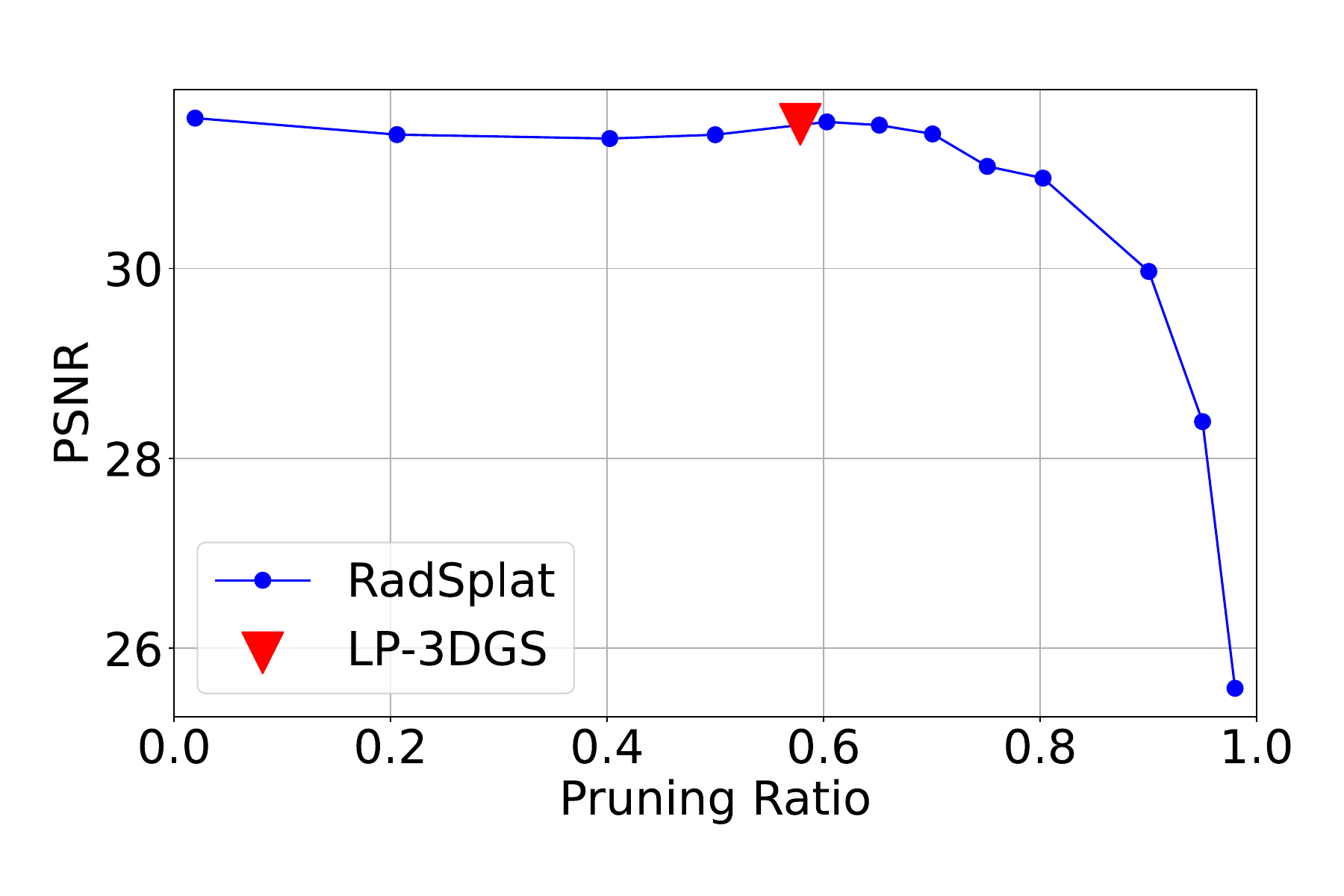}
    \end{subfigure}
    \begin{subfigure}[b]{0.31\textwidth}
        \includegraphics[width=\textwidth]{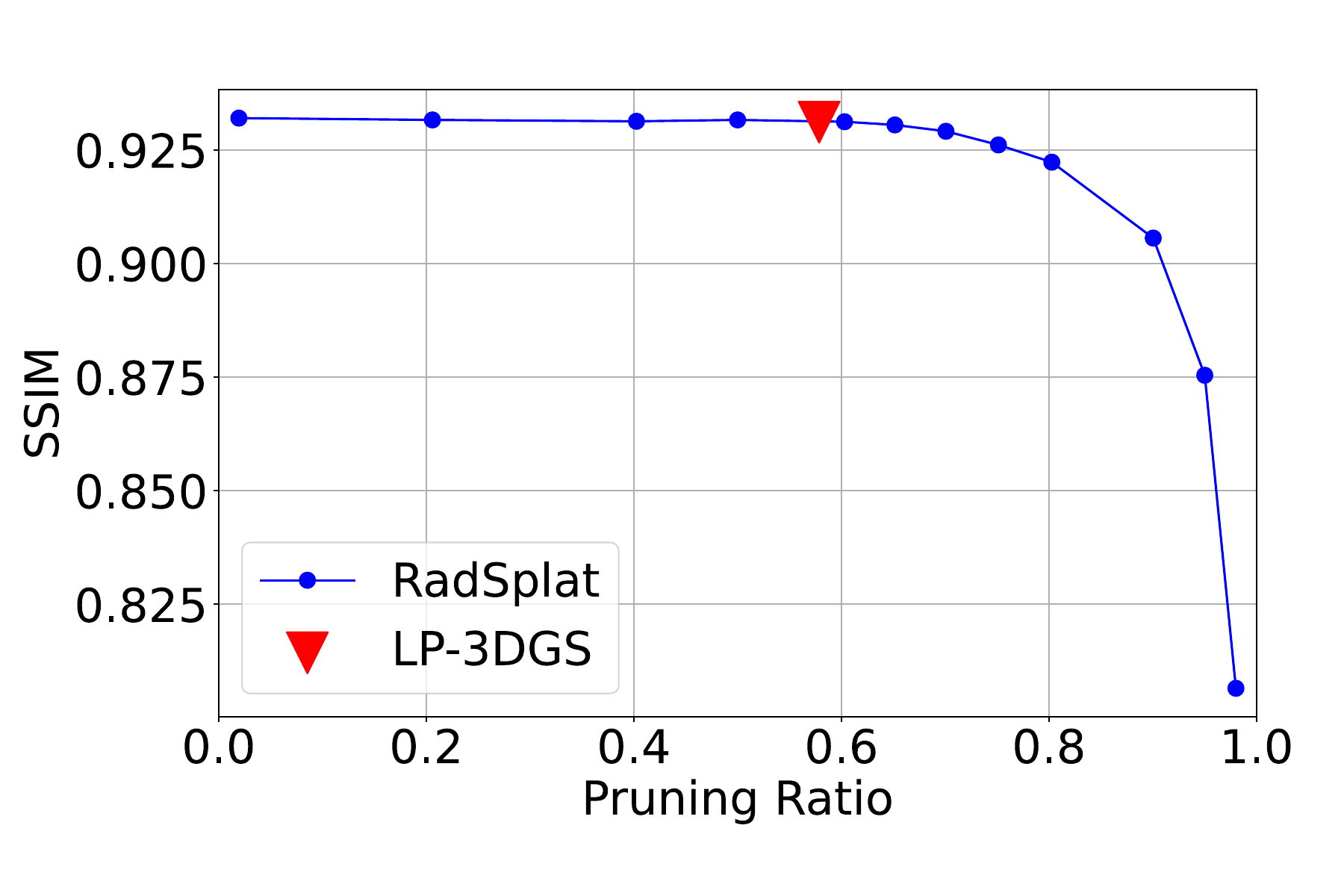}
    \end{subfigure}
    \begin{subfigure}[b]{0.31\textwidth}
        \includegraphics[width=\textwidth]{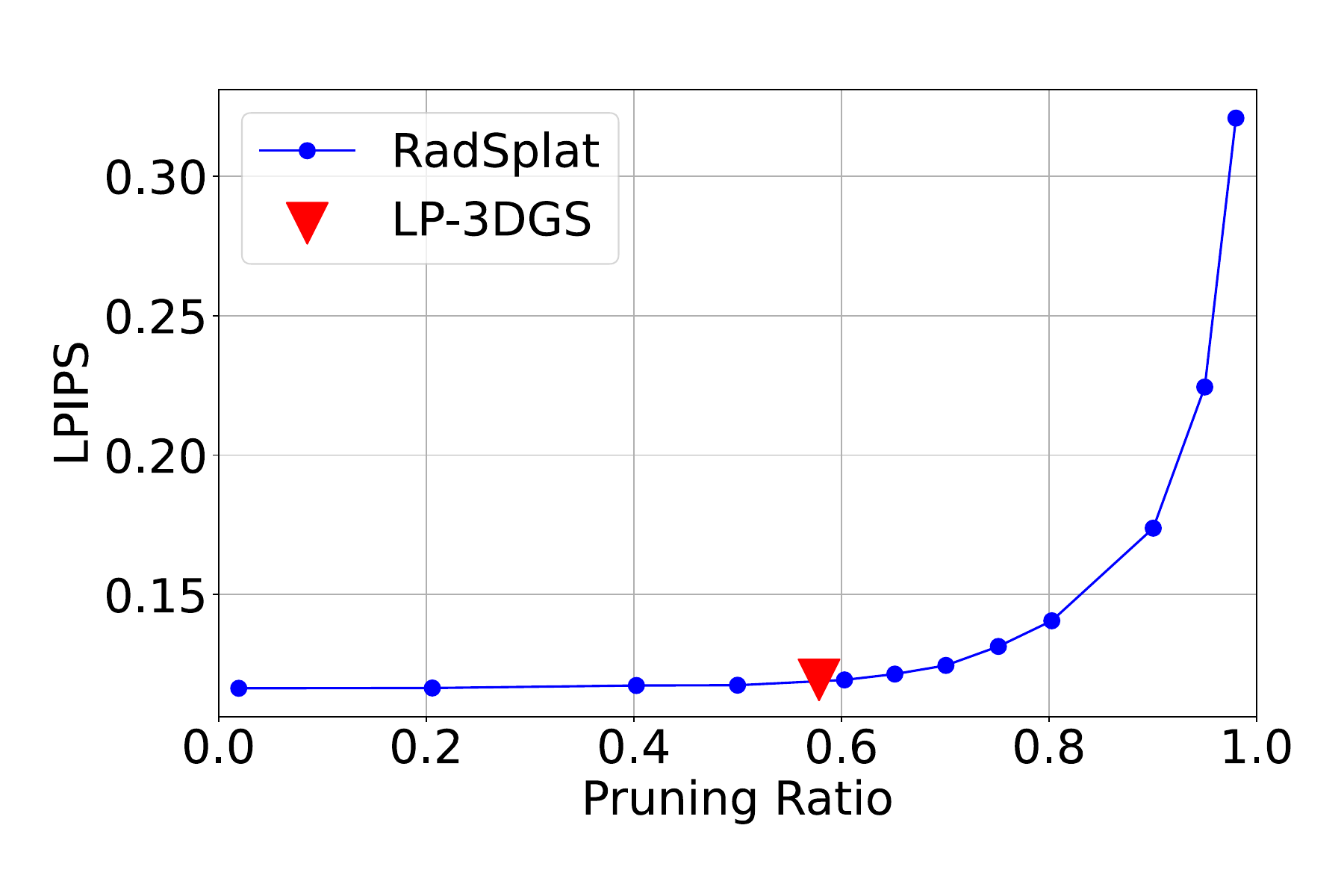}
    \end{subfigure}

    \rotatebox[origin=t, y=1.4cm]{90}{\textbf{Mini-Splatting}}
    \begin{subfigure}[b]{0.31\textwidth}
        \includegraphics[width=\textwidth]{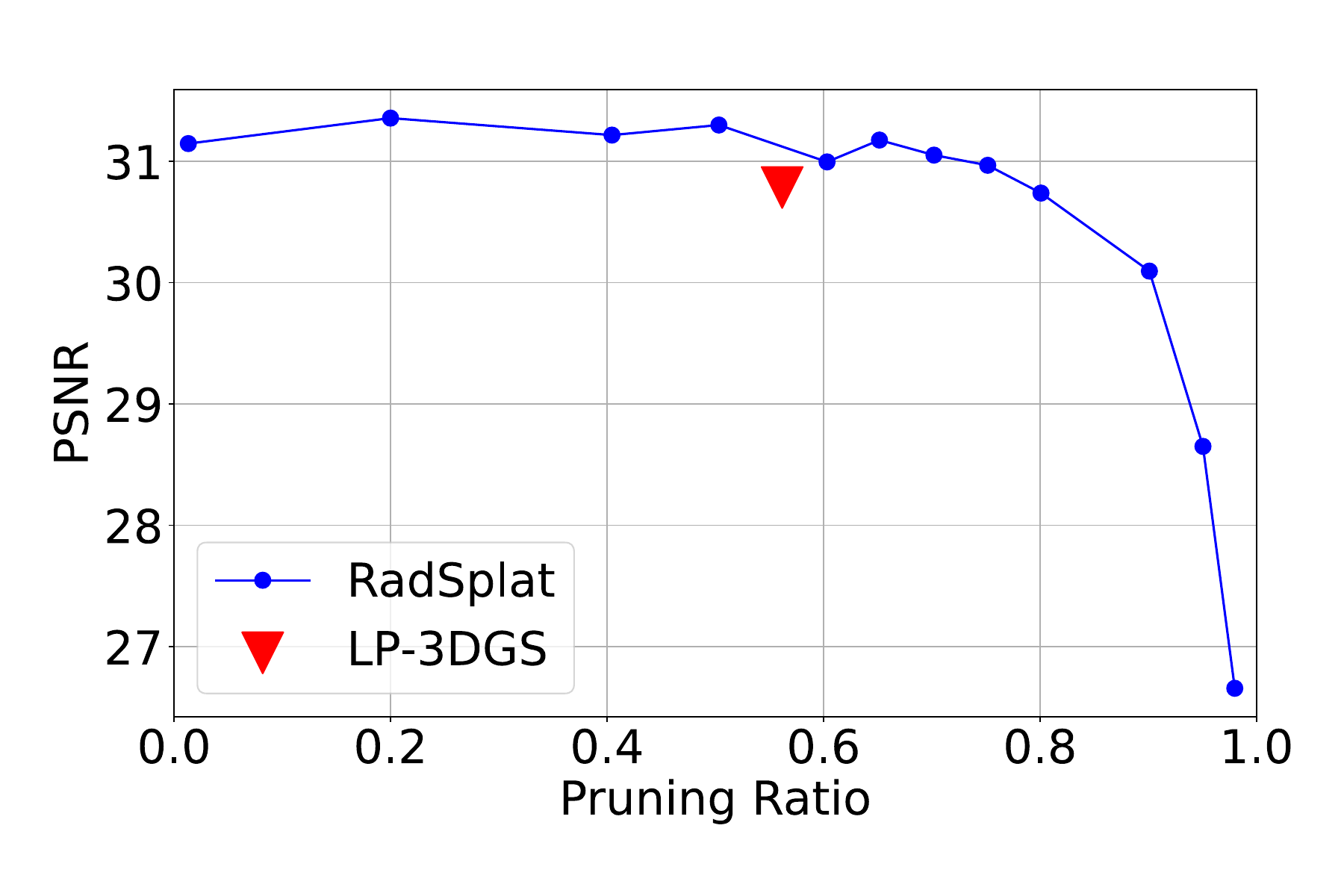}
    \end{subfigure}
    \begin{subfigure}[b]{0.31\textwidth}
        \includegraphics[width=\textwidth]{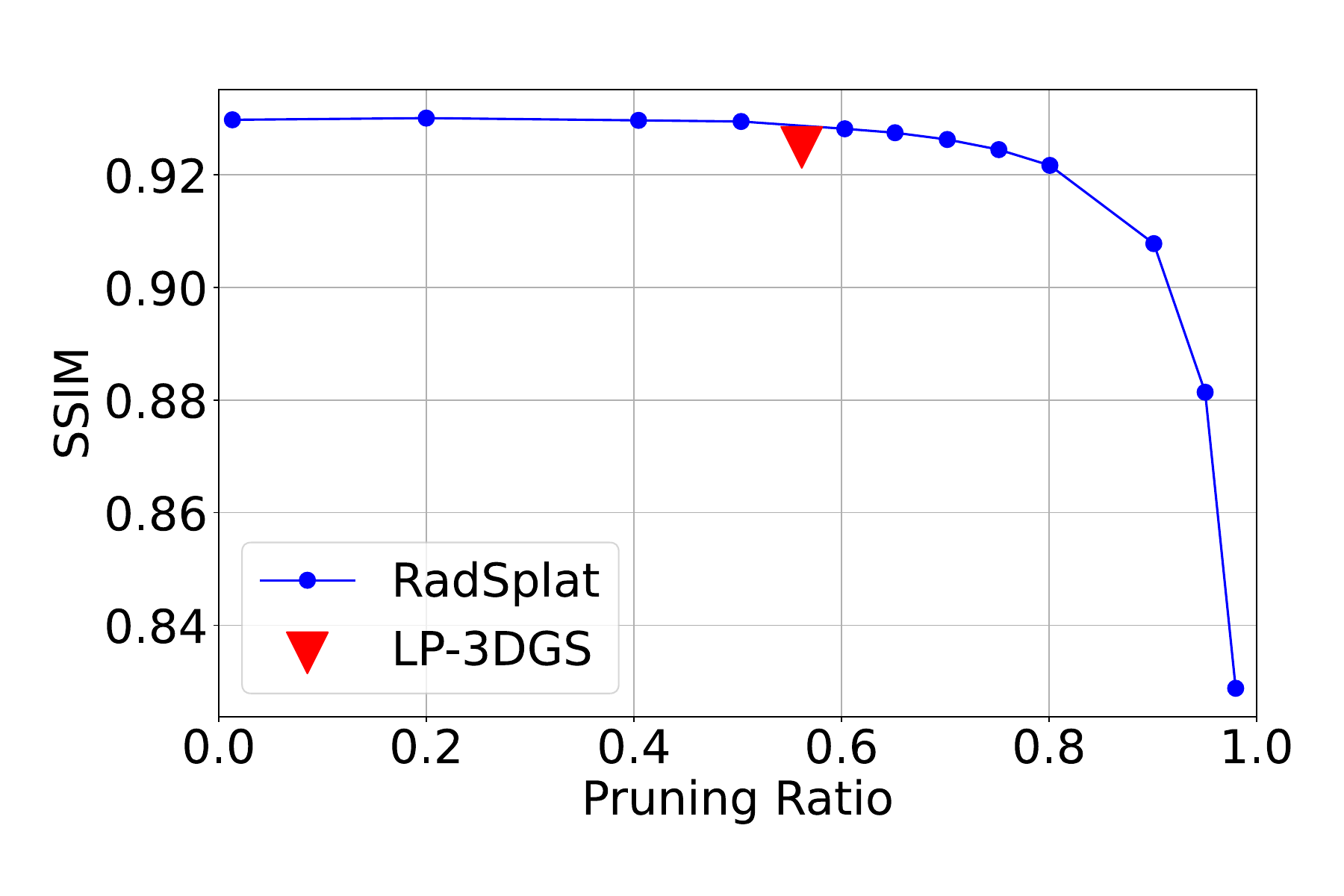}
    \end{subfigure}
    \begin{subfigure}[b]{0.31\textwidth}
        \includegraphics[width=\textwidth]{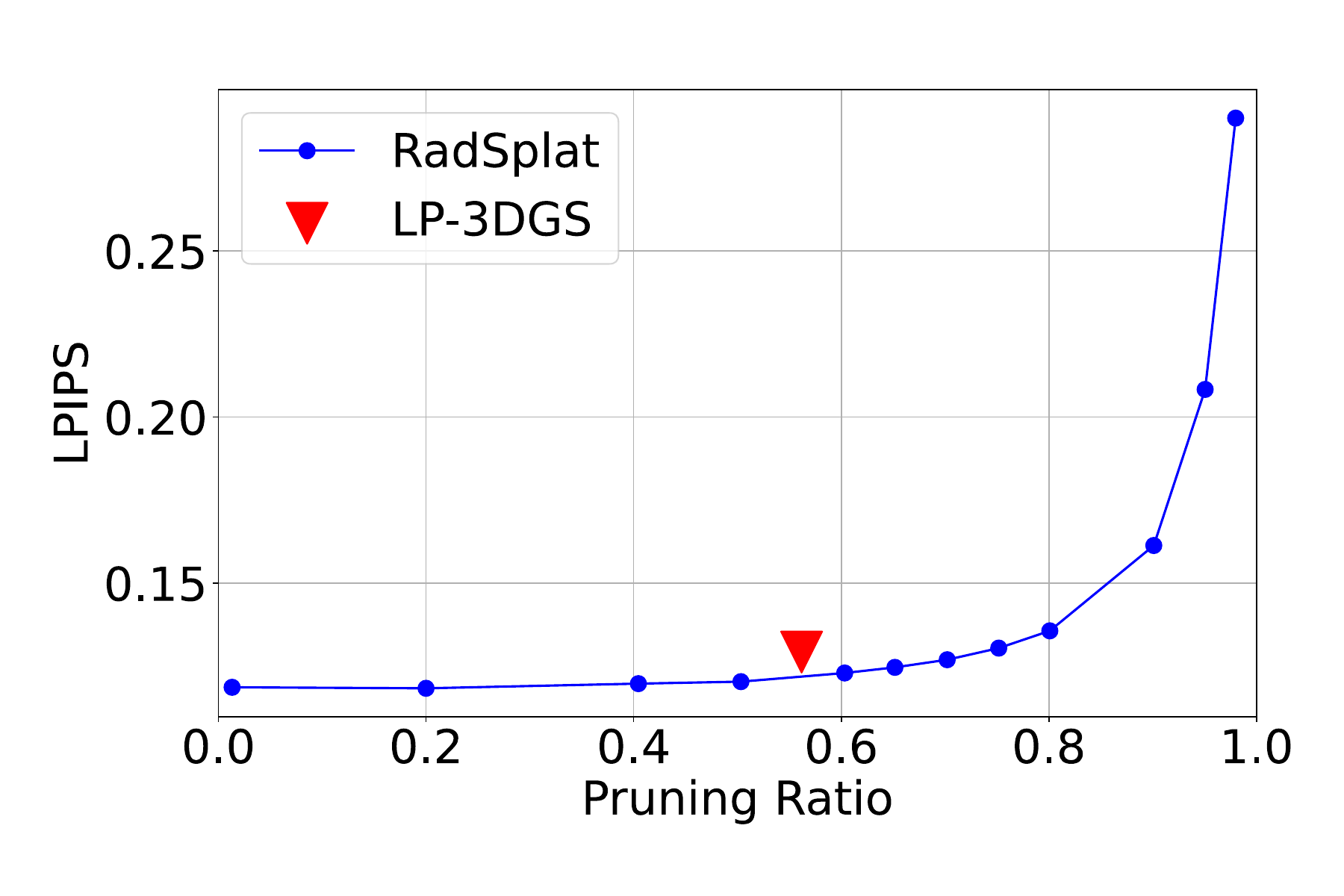}
    \end{subfigure}
    \end{minipage}

    \begin{minipage}[b]{0.05\textwidth}
        \rotatebox[origin=t, y=2.8cm]{90}{\textbf{Room}}
    \end{minipage}
    \begin{minipage}[b]{0.90\textwidth}
    \rotatebox[origin=t, y=1.3cm]{90}{\textbf{RadSplat}}
    \begin{subfigure}[b]{0.31\textwidth}
        \includegraphics[width=\textwidth]{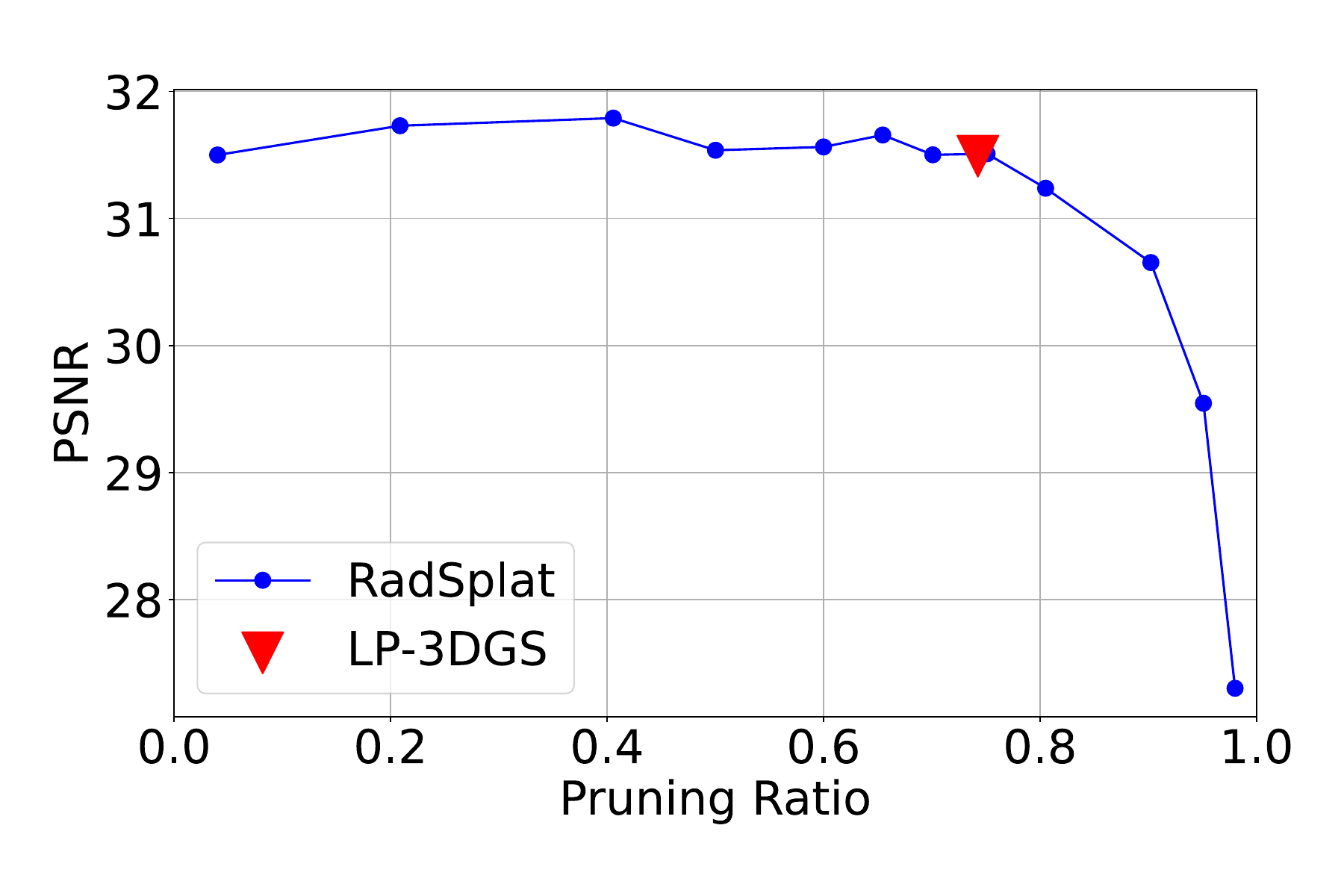}
    \end{subfigure}
    \begin{subfigure}[b]{0.31\textwidth}
        \includegraphics[width=\textwidth]{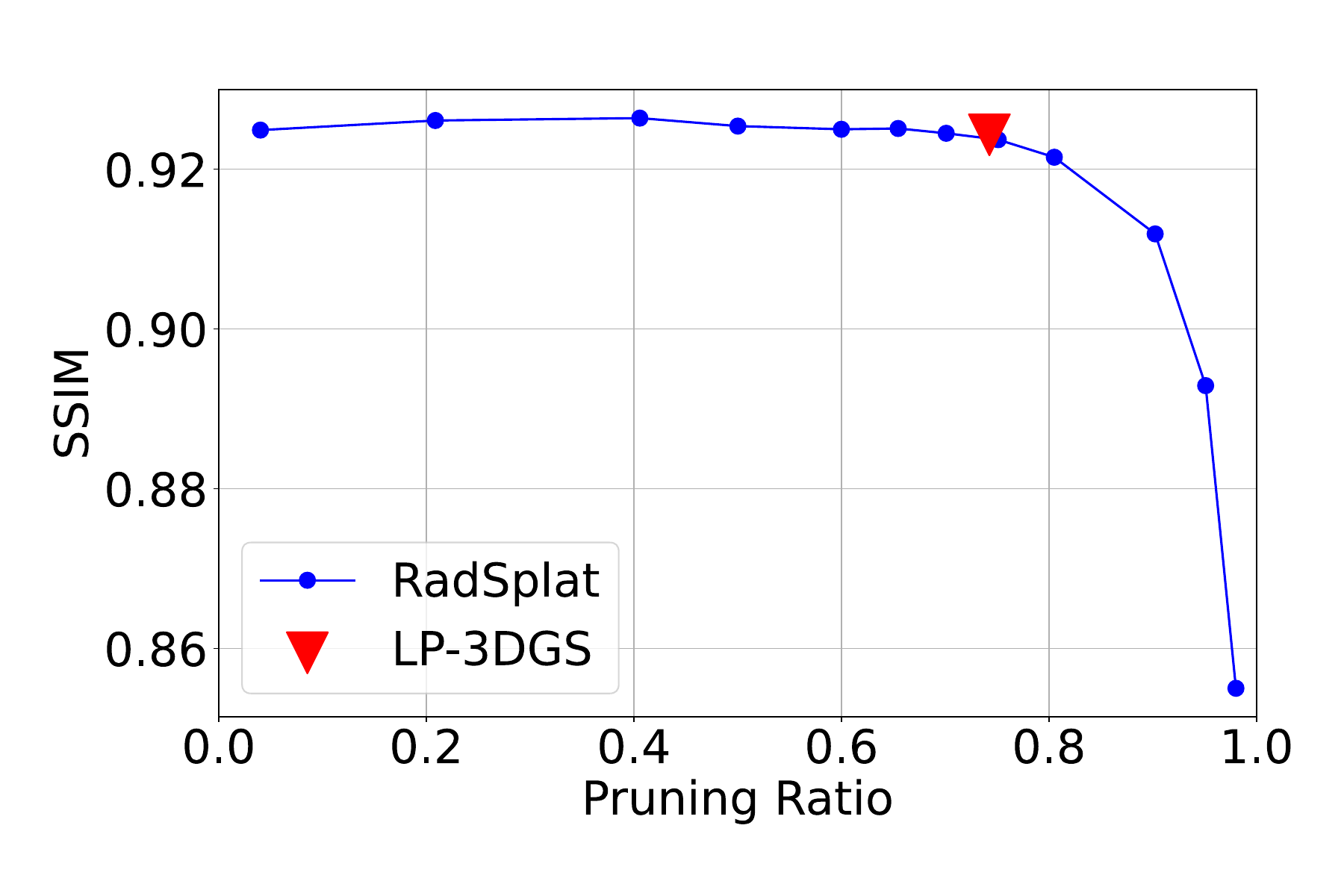}
    \end{subfigure}
    \begin{subfigure}[b]{0.31\textwidth}
        \includegraphics[width=\textwidth]{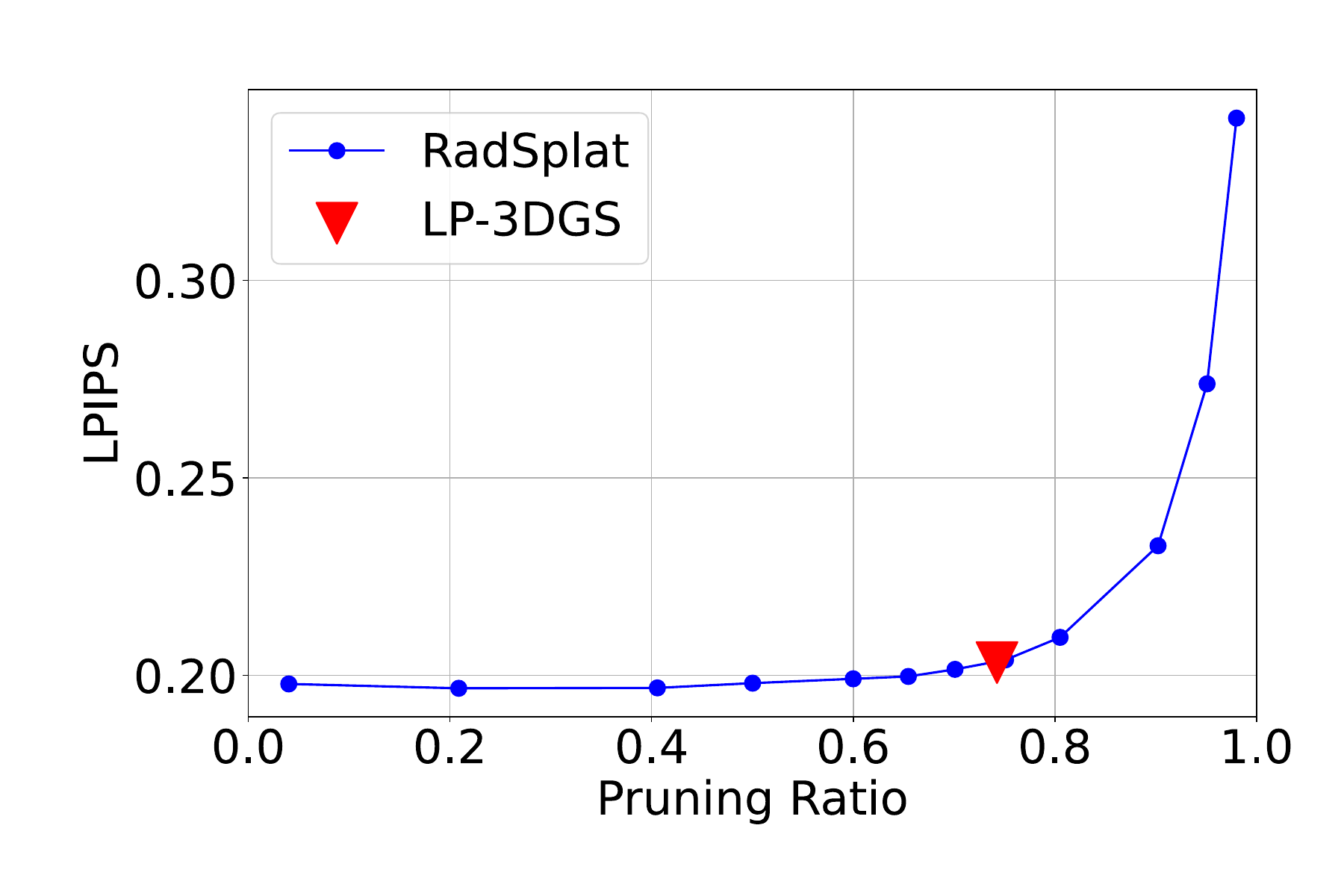}
    \end{subfigure}

    \rotatebox[origin=t, y=1.4cm]{90}{\textbf{Mini-Splatting}}
    \begin{subfigure}[b]{0.31\textwidth}
        \includegraphics[width=\textwidth]{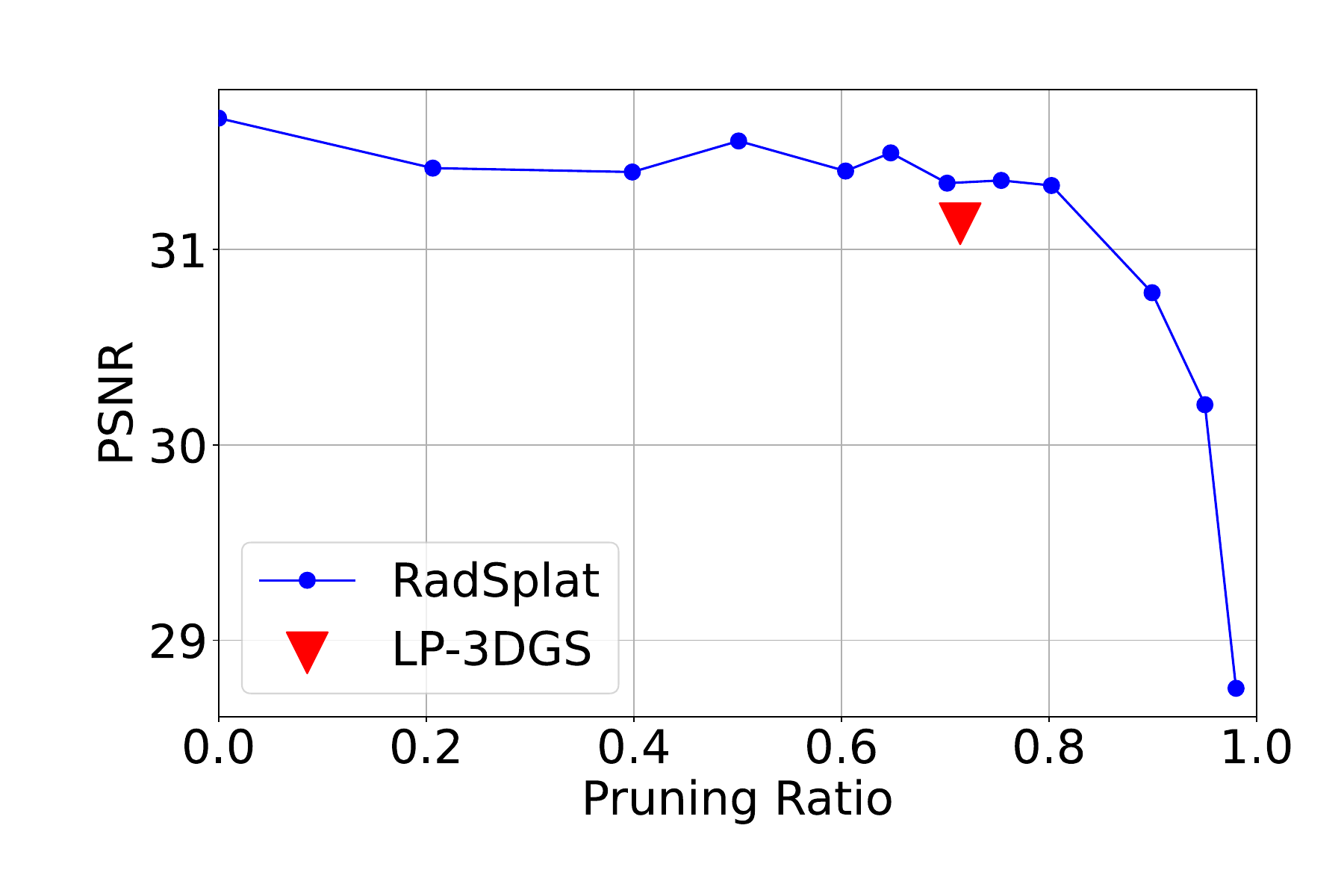}
    \end{subfigure}
    \begin{subfigure}[b]{0.31\textwidth}
        \includegraphics[width=\textwidth]{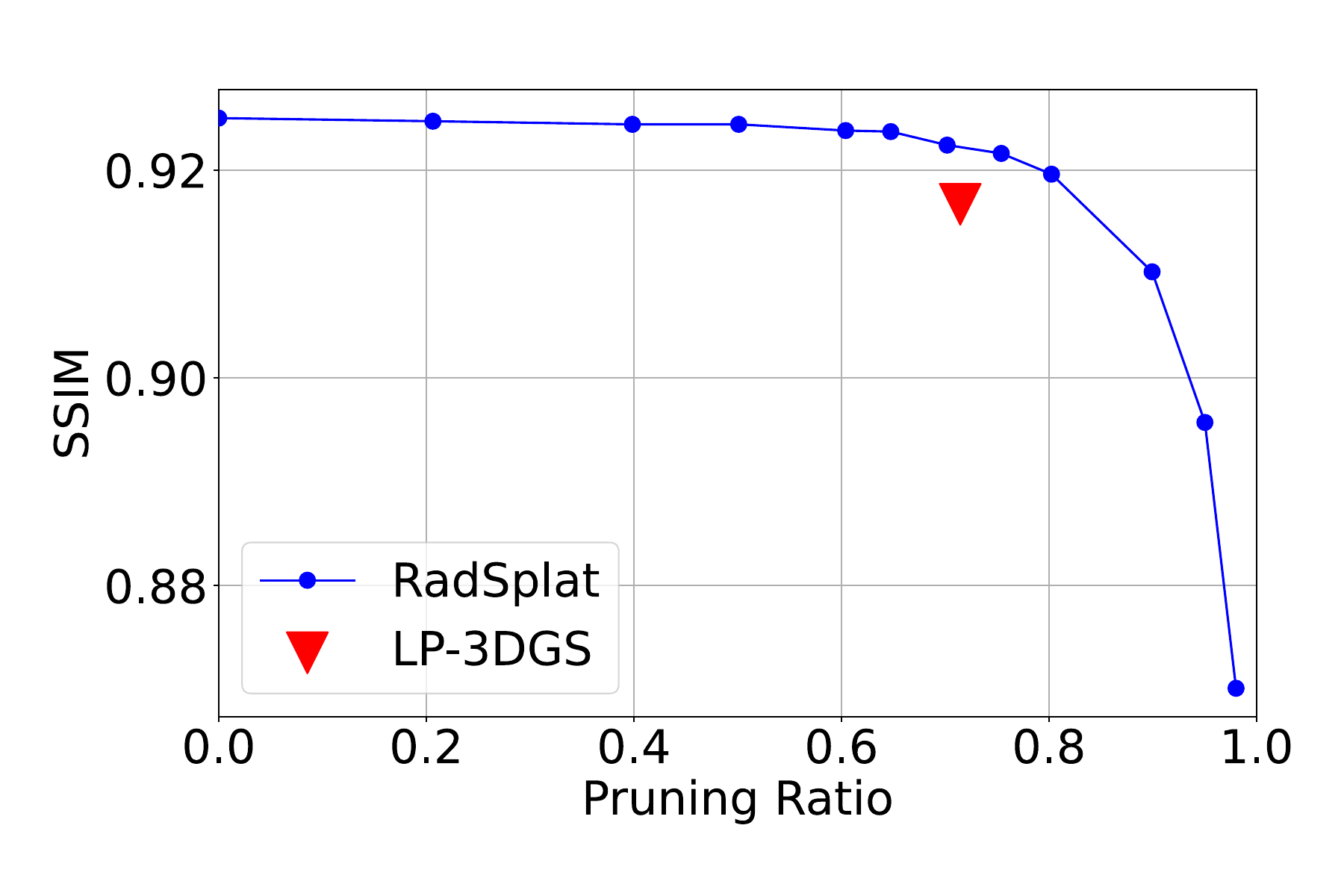}
    \end{subfigure}
    \begin{subfigure}[b]{0.31\textwidth}
        \includegraphics[width=\textwidth]{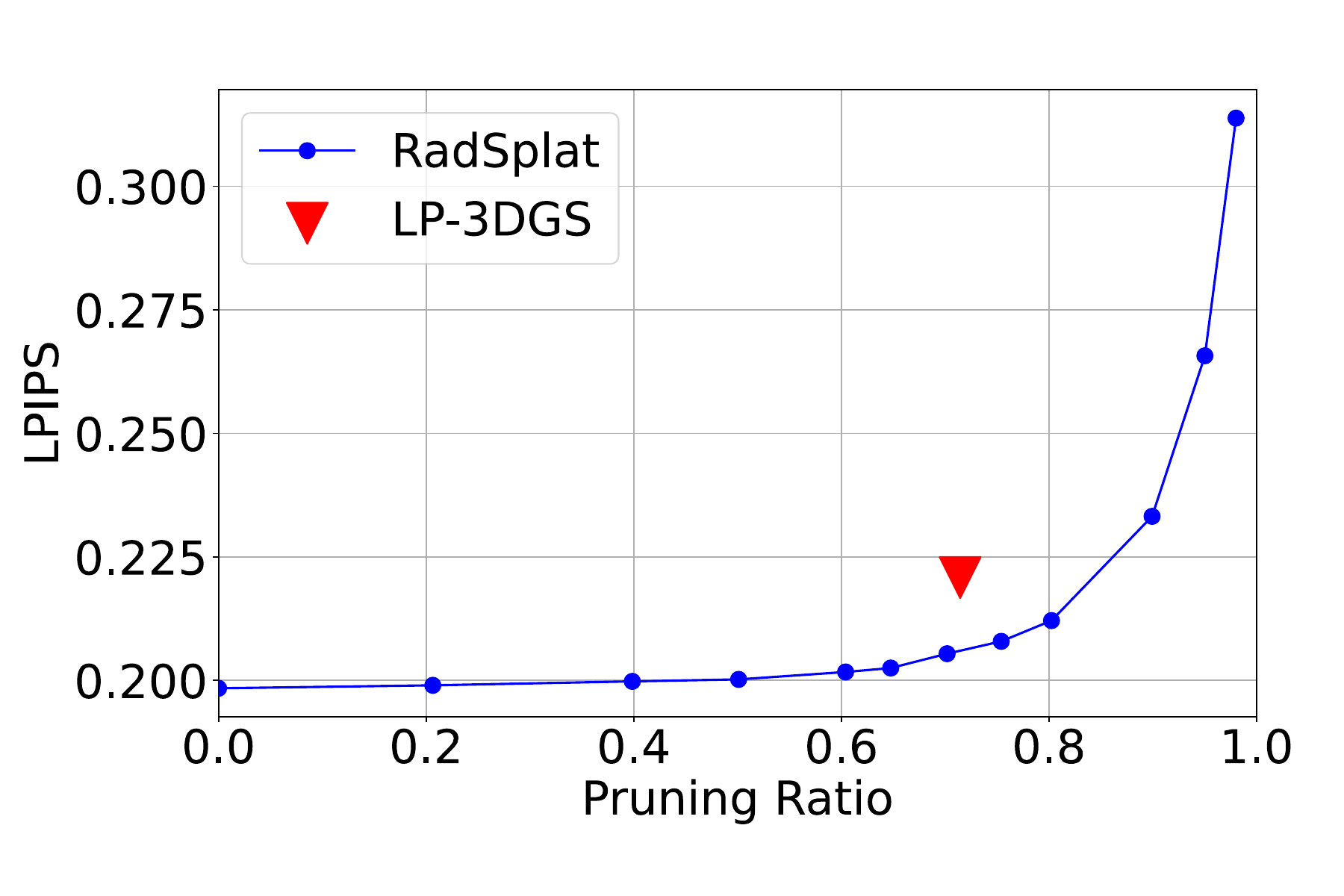}
    \end{subfigure}
    \end{minipage}

    \caption{The performance changes with the pruning ratio in different scenes}
    \label{fig:garden}
\end{figure}

\paragraph{Quantitative Results} The blue lines in Figure \ref{fig:garden} show the results of sweeping pruning ratios using RadSplat and Mini-Splatting for the \textit{Kitchen}, and \textit{Room} scenes. Figures on other scenes of MipNeRF 360 are listed in Appendix \ref{appendix}. The result of the learned LP-3DGS model size and rendering quality is indicated by red triangles.

The quantitative results fluctuate at lower pruning ratios but generally maintain around a certain value up to a specific point. After passing that point, the rendering quality decreases significantly. It's worth noting that such decreasing point varies for different scenes. Instead of manually searching such optimal pruning ratio, it clearly shows our LP-3DGS method could learn the optimal model size embedding with the scene learning process, with only one-time training.
The Table \ref{tab:prune_ratio_list} lists the quantitative results of all scenes in MipNeRF360 dataset. It clearly shows that each scene converge into different model size leveraging our LP-3DGS method with maintaining almost the same rendering quality.
The pruning ratio varies based on what importance score is used, but LP-3DGS could find the optimal pruning ratio on the corresponding score.

\begin{table}[htb]
\caption{The results comparison on the MipNeRF360 dataset shows that LP-3DGS has similar performance after pruning and achieves different pruning ratios for different scenes. This demonstrates LP-3DGS's ability to adaptively find the optimal pruning ratio, maintaining performance while effectively compressing the model.}
\label{tab:prune_ratio_list}
\begin{adjustbox}{max width=\linewidth}
\begin{tabular}{ccccccccccc}
    \toprule
    Scene           & Bicycle       & Bonsai        & Counter       & Kitchen       & Room          & Stump         & Garden        & Flowers & Treehill & AVG \\ 
    \midrule
    Baseline PSNR $\uparrow$   & 25.087  & 32.262 & 29.079  & 31.581  & 31.500 & 26.655 & 27.254 & 21.348  & 22.561 & 27.48  \\
    LP-3DGS (RadSplat Score)   & 25.099  & 32.094 & 28.936  & 31.515  & 31.490 & 26.687 & 27.290 & 21.383  & 22.706  & 27.47 \\
    LP-3DGS (Mini-Splatting Score) & 24.906  & 31.370 & 28.4098  & 30.785  & 31.132 & 26.679 & 27.095 & 21.150  & 22.522  & 27.12  \\ \hline
    Baseline SSIM $\uparrow$   & 0.7464  & 0.9460 & 0.9138  & 0.9320  & 0.9249 & 0.7700 & 0.8557 & 0.5876  & 0.6358 & 0.8125  \\
    LP-3DGS (RadSplat Score)   & 0.7458  & 0.9441 & 0.9120  & 0.9311  & 0.9243 & 0.7714 & 0.8548 & 0.5865  & 0.6381  & 0.8120 \\
    LP-3DGS (Mini-Splatting Score) & 0.7373  & 0.9358 & 0.9017  & 0.9249  & 0.9167 & 0.7677 & 0.8493 & 0.5756  & 0.6336 & 0.8047 \\ \hline
    Baseline LPIPS $\downarrow$& 0.2441  & 0.1799 & 0.1839  & 0.1164  & 0.1978 & 0.2423 & 0.1224 & 0.3601  & 0.3469 & 0.2215  \\
    LP-3DGS (RadSplat Score)   & 0.2516  & 0.1865 & 0.1896  & 0.1194  & 0.2032 & 0.2466 & 0.1270 & 0.3656  & 0.3527 & 0.2269  \\
    LP-3DGS (Mini-Splatting Score) & 0.2642  & 0.2036 & 0.2068  & 0.1292  & 0.2208 & 0.2553 & 0.1353 & 0.3753 &  0.3618 & 0.2391  \\ \hline
    RadSplat Score pruning ratio       & 0.64      & 0.65   & 0.66    & 0.58    & 0.74   & 0.65   & 0.59   & 0.59   & 0.59  &  0.63      \\
    Mini-Splatting Score pruning ratio  & 0.57   & 0.67   & 0.64    & 0.56    & 0.71   & 0.61   & 0.60   & 0.54   & 0.54  &  0.60  \\
    \bottomrule
    \end{tabular}
\end{adjustbox}
\end{table}

\paragraph{Training Cost}
Table \ref{tab:training cost} shows the training cost of LP-3DGS on MipNeRF360 dataset. In our setup, since after 20000th iteration, the model will be pruned based on the learned mask values. The number of Gaussian points will be significantly reduced, which makes the later stages of training take much less time than the non-pruned version. Even with the embedding of the mask learning function, the overall training cost is similar with that of the vanilla 3DGS. In most cases, the peak training memory usage is slightly larger because training the mask requires more GPU memory. However, after pruning, the 3DGS model size becomes much smaller, leading to a significant improvement in rendering speed, measured in terms of FPS.

\begin{table}[thb]
\caption{Training cost and on MipNeRF360 Dataset. Training time of LP-3DGS is similar with baseline but since the model is compressed, the FPS is larger.}
\label{tab:training cost}
\begin{adjustbox}{max width=\linewidth}
\begin{tabular}{cccccccccc}
    \toprule
    Scene           & Bicycle       & Bonsai        & Counter       & Kitchen       & Room          & Stump         & Garden        & Flowers & Treehill \\ 
    \midrule
    3DGS Training time (Minute) & 49      & 34     & 26      & 33      & 27   & 37    & 47     & 33      & 32       \\
    LP-3DGS (RadSplat Score)    & 43      & 27     & 28      & 34      & 30   & 35    & 46     & 34      & 33       \\
    LP-3DGS (Mini-Splatting Score) & 44      & 27     & 28      & 35      & 29   & 34    & 46     & 34      & 33       \\ \hline
    3DGS Peak Memory (GB)   & 14.7    & 8.6    & 9.4     & 9.3     & 10.6 & 12.2  & 15.7   & 10.3    & 9.4      \\ 
    LP-3DGS (RadSplat Score)    & 16.1    & 8.5    & 11.3     & 12.3    & 11.8 & 12.1  & 15.8   & 10.1     & 10.3     \\ 
    LP-3DGS (Mini-Splatting Score) & 15.5    & 8.4    & 12.7    & 13.0    & 13.0 & 12.1  & 15.2   & 9.7     & 9.7     \\ \hline
    3DGS FPS      & 132     & 417    & 421     & 315     & 380  & 164   & 129    & 200     & 205      \\
    LP-3DGS (RadSplat Score)    & 324     & 662    & 670     & 542     & 692  & 371   & 296    & 412     & 411      \\
    LP-3DGS (Mini-Splatting Score) & 290     & 634    & 650     & 507     & 662  & 341   & 252    & 368     & 384      \\
    \bottomrule
    \end{tabular}
\end{adjustbox}
\end{table}

\subsection{Ablation Study}
\label{sec:ablation}
A recent prior work Compact3D (\citet{lee2023compact}) proposes to leverage STE to train a binary mask on opacity and scale of Gaussian parameter. To conduct a fair comparison between STE based mask and our LP-3DGS, we make two ablation studies, one is replacing the STE mask in Compact3D with our method, the other is applying STE mask on importance score of RadSpalt. The formula of STE mask is
\begin{equation}
\label{equ:ste}
    M(m) = \stopgradient \, ( \mathds{1}[f(m) > \epsilon] - f(m) ) + f(m)
\end{equation}
$\stopgradient$ means stop gradients, $\mathds{1} [\cdot]$ is the indicator function and $f(\cdot)$ is sigmoid function, $\epsilon$ is masking threshold. 

\paragraph{Comparison with Compact3D}
We firstly apply Gumbel-sidmoid activated mask, instead of STE mask, on the opacity and scale of gaussians in the same way as proposed in Compact3D. The threshold $\epsilon$ in Equation \ref{equ:ste} and mask loss coefficient follows the default settings in Compact3D. Table \ref{tab:ste_vs_gumbel} shows the comparison between two methods.

\begin{table}[htb]
\caption{Results with/without trainable mask on Gaussian opacity and scale}
\label{tab:ste_vs_gumbel}
\begin{adjustbox}{max width=\linewidth}
\begin{tabular}{cccccccccccc}
    \toprule
    & Scene           & Bicycle       & Bonsai        & Counter       & Kitchen       & Room          & Stump         & Garden        & Flowers & Treehill & AVG  \\ 
    \midrule
    \multirow{2}{*}{PSNR} & Compact3D & 24.846  & 32.19 & 29.066  & 30.867  & 31.489   & 26.408 & 27.026 & 21.187  & 22.479 & 27.284 \\
    & LP-3DGS  & 25.087  & 32.2 & 29.033   & 31.213  & 31.678 & 26.658  & 27.223 &  21.32       &  22.569  &  27.442    \\ \hline
    \multirow{2}{*}{SSIM} & Compact3D  & 0.7292  & 0.9462  & 0.9137  & 0.925  & 0.9251 & 0.7563   & 0.8446 & 0.5773  & 0.6305  & 0.8053\\
    & LP-3DGS  & 0.7438  & 0.9461  & 0.9141  & 0.9305  & 0.9263 & 0.7687 & 0.8547 &  0.5843       &  0.6358  & 0.8116     \\ \hline
    \multirow{2}{*}{LPIPS} & Compact3D  & 0.266  & 0.1815 & 0.1866  & 0.124  & 0.2012 & 0.2615 & 0.1401 & 0.3722  & 0.3555 & 0.2320 \\
    & LP-3DGS & 0.2526  & 0.1833 & 0.1867  & 0.1201  & 0.2013 & 0.2472 & 0.1275 &  0.3668       &  0.3513    &  0.2263  \\ \hline
    \multirow{2}{*}{\#Gaussians} & Compact3D & 2620663    & 666558   & 570126    & 1050079    & 566332   & 1902711   & 2412796   &  1685224       &  2089515   &   1507109 \\
    & LP-3DGS & 2510992    & 542235   & 506391    & 887161    & 479681   & 2014270   & 2836989   &   1747766      &  1804155   &  1481071    \\
    \bottomrule
    \end{tabular}
\end{adjustbox}
\end{table}
For most cases, our LP-3DGS learns a higher pruning ratio, except Stump, Garden and Flowers scene. In terms of rendering quality, our LP-3DGS outperforms compact3D using STE based mask with even smaller model size in most scenes.

\paragraph{STE Mask on Importance Score}
We also apply STE mask on the pruning importance score to compare with our method. The Equation \ref{equ:apply_mask} would be rewriten as
\begin{equation}
    o_{im} = o_i * M(m_i*is)
\end{equation}
where $M$ is shown in Equation \ref{equ:ste}. The same as mentioned before, the parameters for STE mask are default values in Compact3D.

\begin{table}[htb]
\caption{Results using LP-3DGS and STE mask on importance score of RadSplat}
\label{tab:ste_vs_gumbel_radsplat}
\begin{adjustbox}{max width=\linewidth}
\begin{tabular}{ccccccccccccc}
    \toprule
    & Scene           & Bicycle       & Bonsai        & Counter       & Kitchen       & Room          & Stump         & Garden        & Flowers & Treehill & AVG\\ 
    \midrule
    \multirow{2}{*}{PSNR} & LP-3DGS   & 25.099  & 32.094 & 28.936  & 31.515  & 31.490 & 26.687 & 27.290 & 21.383  & 22.706  & 27.470 \\
    & STE mask  & 24.833  & 30.947 & 28.371   & 30.705  & 30.950 & 26.396  & 26.793 & 21.056  &  22.552 & 26.955\\ \hline
    \multirow{2}{*}{SSIM} & LP-3DGS   & 0.7458  & 0.9441 & 0.9120  & 0.9311  & 0.9243 & 0.7714 & 0.8548 & 0.5865  & 0.6381  & 0.8120 \\
    & STE mask  & 0.7231  & 0.9268  & 0.8925  & 0.9162  & 0.9120 & 0.7514 & 0.8289 & 0.5624  &  0.6196 & 0.7922\\ \hline
    \multirow{2}{*}{LPIPS} & LP-3DGS  & 0.2441  & 0.1799 & 0.1839  & 0.1164  & 0.1978 & 0.2423 & 0.1224 & 0.3601  & 0.3469 & 0.2215 \\
    & STE mask & 0.2937  & 0.2194 & 0.2274  & 0.1480  & 0.2334 & 0.2899 & 0.1771 & 0.3988  &  0.3983 & 0.2651\\ \hline
    \multirow{2}{*}{Pruning Ratio} & LP-3DGS  & 0.64      & 0.65   & 0.66    & 0.58    & 0.74   & 0.65   & 0.59   & 0.59   & 0.59  &  0.63 \\
    & STE mask & 0.84    & 0.88   & 0.88    & 0.87    & 0.89   & 0.86   &  0.85  & 0.83 & 0.83 & 0.86\\
    \bottomrule
    \end{tabular}
\end{adjustbox}
\end{table}
Table \ref{tab:ste_vs_gumbel_radsplat} shows that under the same settings, after applying the mask to the importance score, the STE mask compresses the mode too much and the performance drops a lot. Trainable mask keeps the gradient of the mask and the comressed model has a more reasonable size. 

\begin{table}[htb]
\caption{Results using LP-3DGS and STE mask on importance score of Mini-Splatting}
\label{tab:ste_vs_gumbel_minisplat}
\begin{adjustbox}{max width=\linewidth}
\begin{tabular}{ccccccccccccc}
    \toprule
    & Scene           & Bicycle       & Bonsai        & Counter       & Kitchen       & Room          & Stump         & Garden        & Flowers & Treehill & AVG\\ 
    \midrule
    \multirow{2}{*}{PSNR} & LP-3DGS   & 24.906  & 31.370 & 28.4098  & 30.785  & 31.132 & 26.679 & 27.095 & 21.150  & 22.522  & 27.12 \\
    & STE mask  & 24.894  & 30.925 & 28.334   & 30.731  & 31.032 & 26.470  & 26.863 & 20.997  &  22.559 & 26.98\\ \hline
    \multirow{2}{*}{SSIM} & LP-3DGS   & 0.7373  & 0.9358 & 0.9017  & 0.9249  & 0.9167 & 0.7677 & 0.8493 & 0.5756  & 0.6336 & 0.8047 \\
    & STE mask  & 0.7287  & 0.9292  & 0.8978  & 0.9232  & 0.9152 & 0.7562 & 0.8381 & 0.5629  &  0.6247 & 0.7973\\ \hline
    \multirow{2}{*}{LPIPS} & LP-3DGS  & 0.2642  & 0.2036 & 0.2068  & 0.1292  & 0.2208 & 0.2553 & 0.1353 & 0.3753 &  0.3618 & 0.2391 \\
    & STE mask & 0.2821  & 0.2154 & 0.2145  & 0.1330  & 0.2241 & 0.2797 & 0.1564 & 0.3939  &  0.3852 & 0.2538\\ \hline
    \multirow{2}{*}{Pruning Ratio} & LP-3DGS  & 0.57   & 0.67   & 0.64    & 0.56    & 0.71   & 0.61   & 0.60   & 0.54   & 0.54  &  0.60\\
    & STE mask & 0.75    & 0.77   & 0.75    & 0.66    & 0.79   & 0.80   & 0.75   & 0.75 & 0.75 & 0.75\\
    \bottomrule
    \end{tabular}
\end{adjustbox}
\end{table}

\section{Discussion and Conclusion}
\label{sec:conclusion}

\paragraph{Broader Impact and Limitation}
LP-3DGS compresses the 3DGS model to an ideal size in a single run, saving storage and computational resources by eliminating the need for parameter sweeping to find the optimal pruning ratio. However, the limitation of this work is that the rendering quality after pruning varies depending on the definition of importance scores.

\paragraph{Conclusion}
In this paper, we present a novel framework, LP-3DGS, which guide the 3DGS model learn the best model size. The framework applies a trainable mask on the importance score of the gaussian points. The mask only would be trained for a certain period and prune the model once. Our method compressed the model as much as possible without significantly sacrificing performance and is able to achieve the optimal compression rate for different test scenes. Compared with STE mask method, ours achieves better performance.

\newpage

\begin{ack}
This research is based upon work supported by the Office of the Director of National Intelligence (ODNI), Intelligence Advanced Research Projects Activity (IARPA), via IARPA R\&D Contract No. 140D0423C0076. The views and conclusions contained herein are those of the authors and should not be interpreted as necessarily representing the official policies or endorsements, either expressed or implied, of the ODNI, IARPA, or the U.S. Government. The U.S. Government is authorized to reproduce and distribute reprints for Governmental purposes notwithstanding any copyright annotation thereon.

\end{ack}

\bibliographystyle{unsrtnat}
\bibliography{refs} 


\newpage
\appendix

\section{Appendix / supplemental material}
\label{appendix}

\textit{Code: } \url{https://github.com/dexgfsdfdsg/LP-3DGS.git}

\subsection{Experiment Results on MipNeRF360 Dataset}

\begin{figure}[htbp]
    \centering

    \makebox[0.05\textwidth]{} 
    \makebox[0.22\textwidth]{\textbf{Ground Truth}}
    \makebox[0.22\textwidth]{\textbf{3DGS}}
    \makebox[0.22\textwidth]{\textbf{RadSplat}}
    \makebox[0.22\textwidth]{\textbf{Mini-Splatting}}
    \par\medskip

    \rotatebox[origin=t, y=1.3cm]{90}{\textbf{Bicycle}}
    \begin{subfigure}[b]{0.23\textwidth}
        \includegraphics[width=\textwidth]{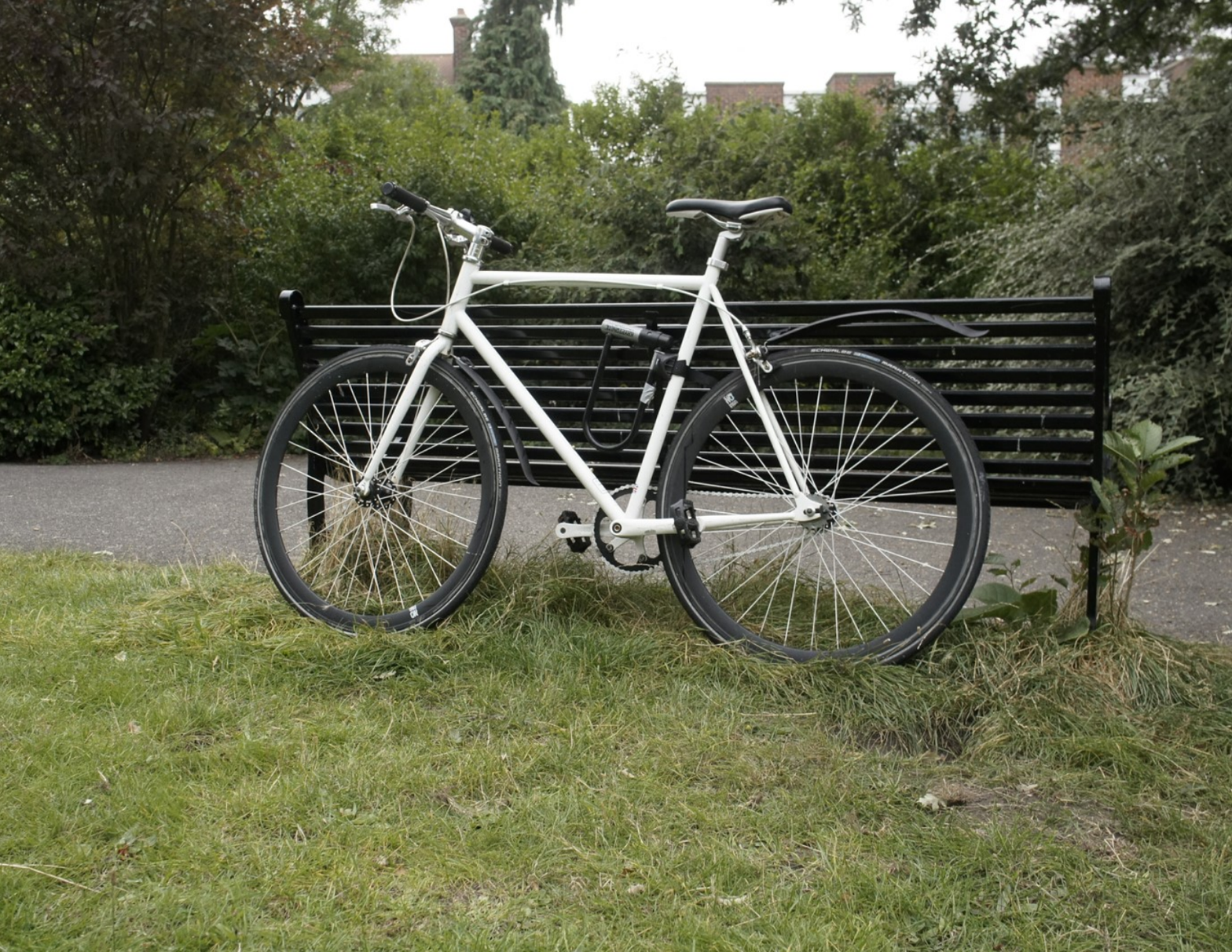}
    \end{subfigure}
    \begin{subfigure}[b]{0.23\textwidth}
        \includegraphics[width=\textwidth]{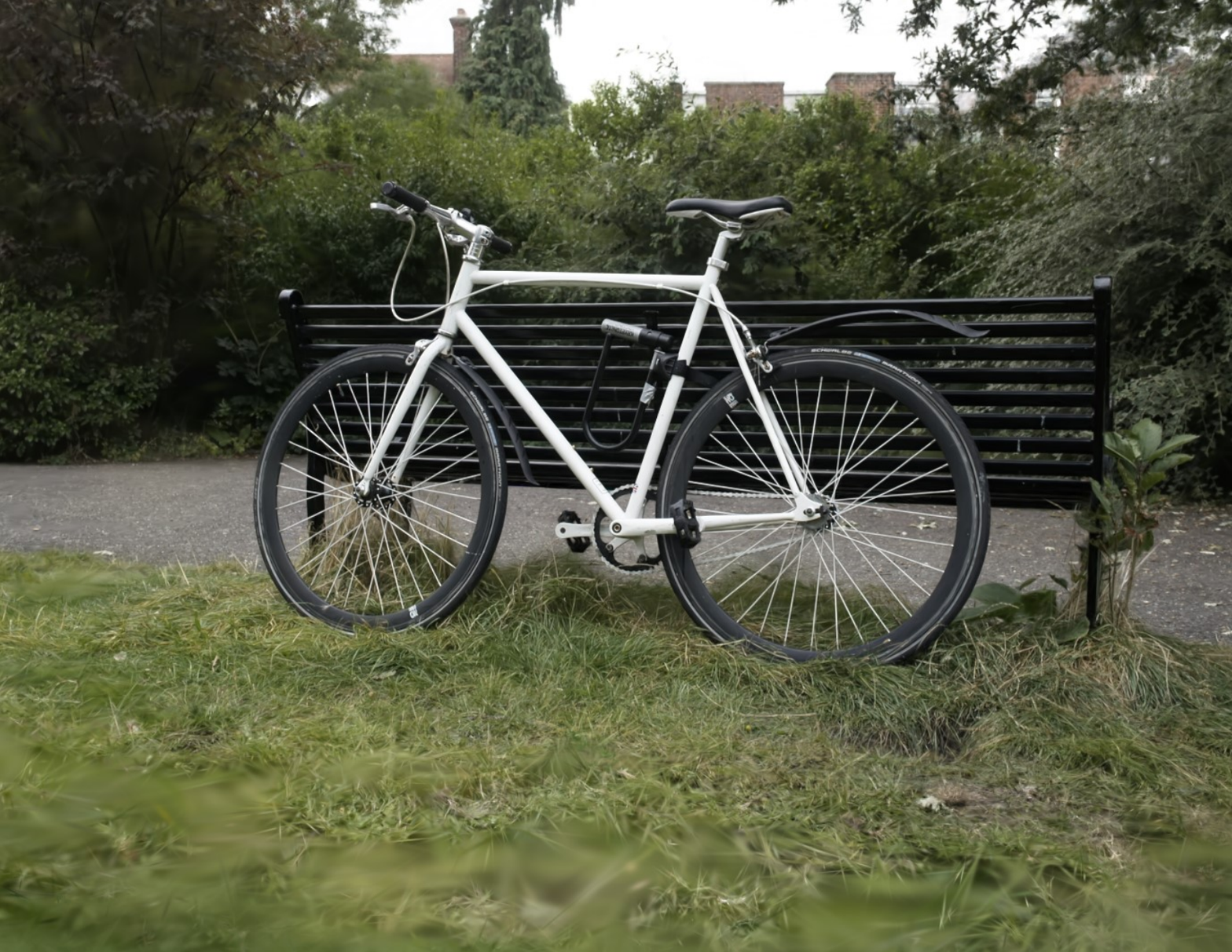}
    \end{subfigure}
    \begin{subfigure}[b]{0.23\textwidth}
        \includegraphics[width=\textwidth]{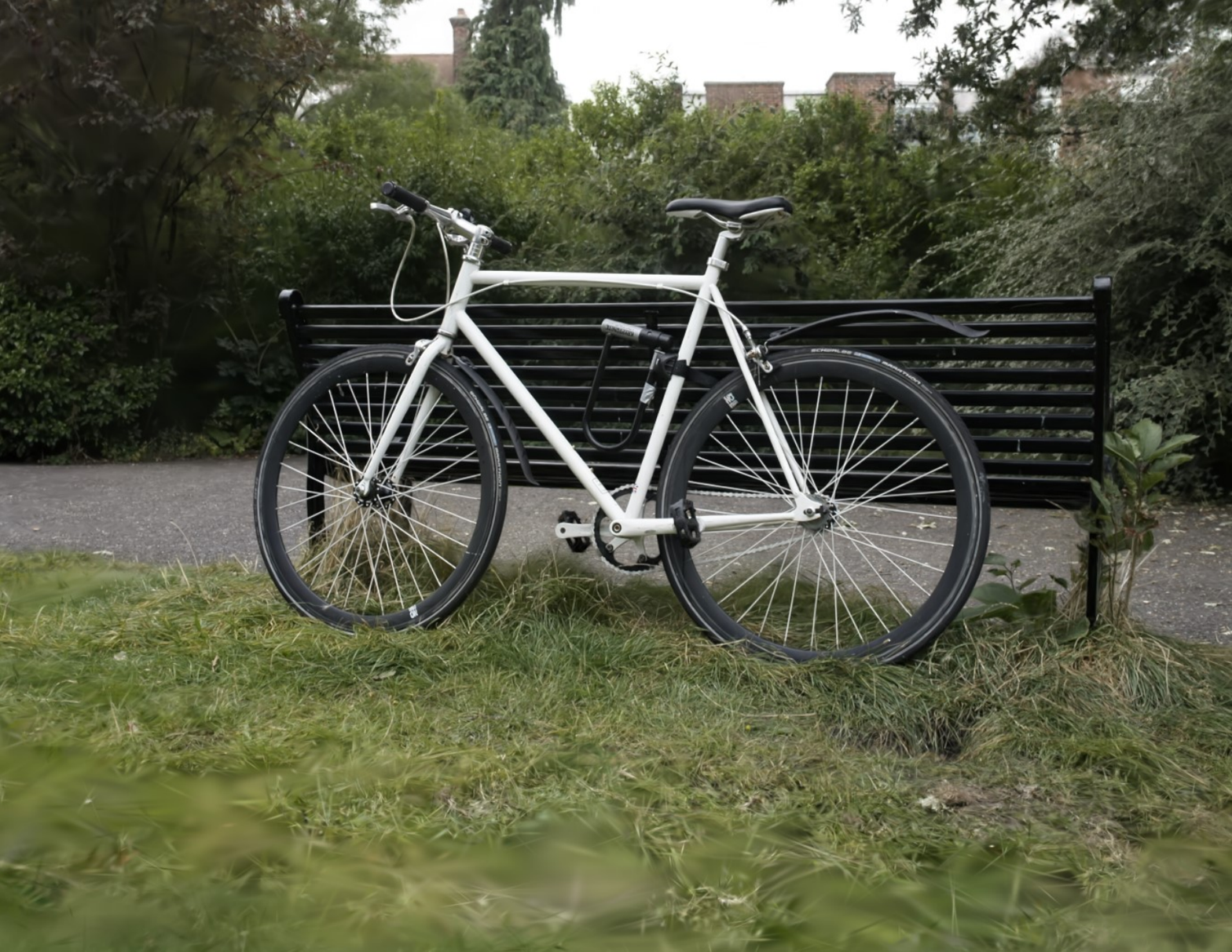}
    \end{subfigure}
    \begin{subfigure}[b]{0.23\textwidth}
        \includegraphics[width=\textwidth]{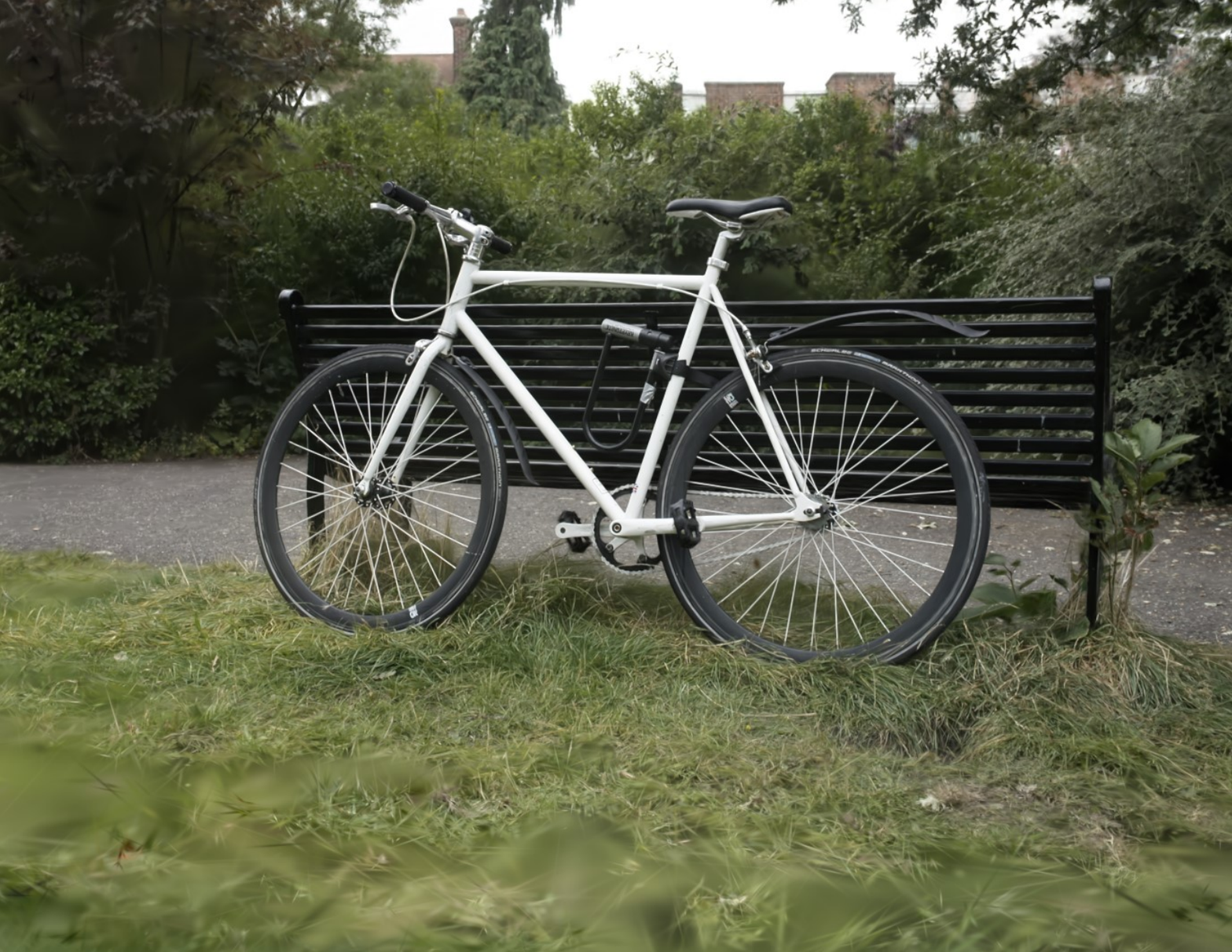}
    \end{subfigure}
    \par\medskip

    \rotatebox[origin=t, y=1.3cm]{90}{\textbf{Bonsai}}
    \begin{subfigure}[b]{0.23\textwidth}
        \includegraphics[width=\textwidth]{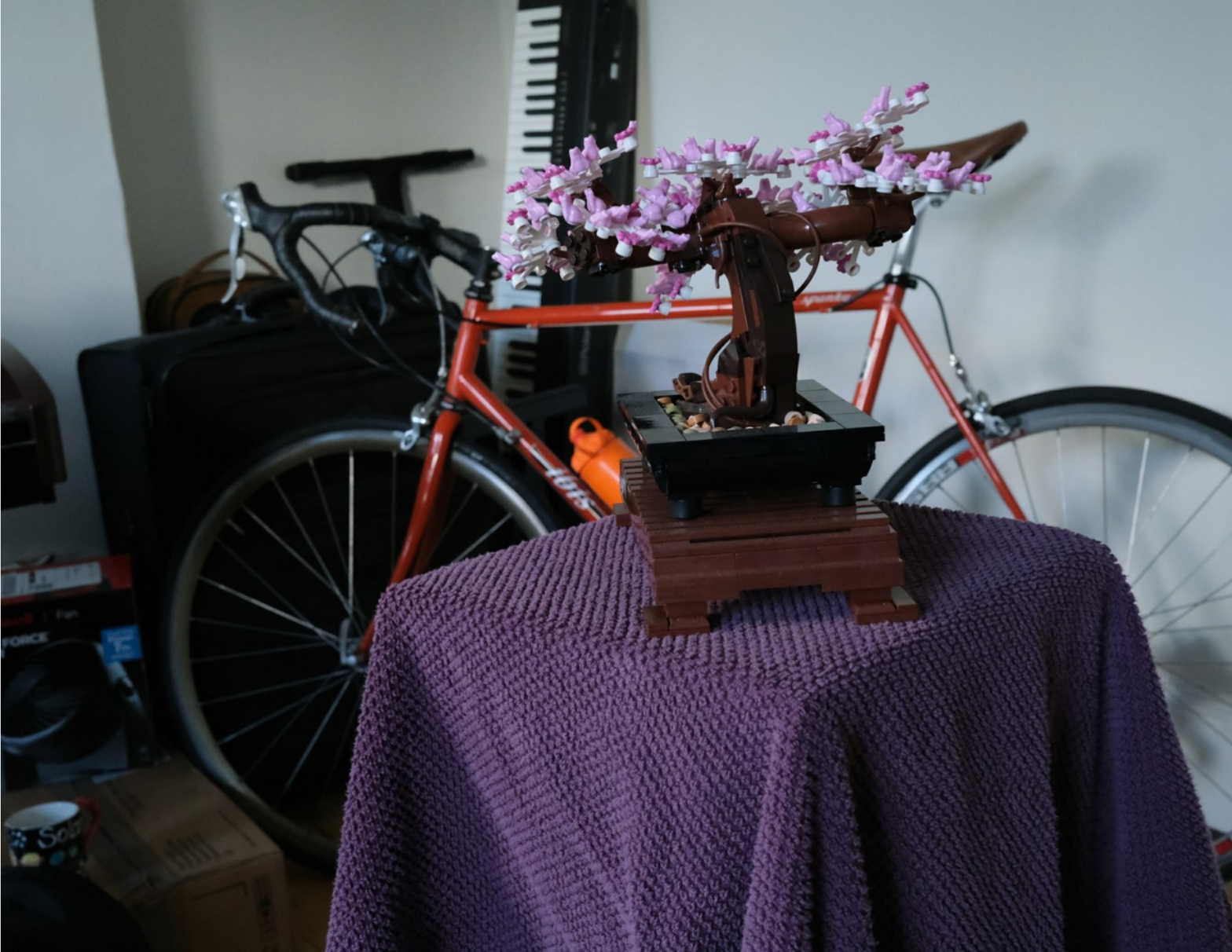}
    \end{subfigure}
    \begin{subfigure}[b]{0.23\textwidth}
        \includegraphics[width=\textwidth]{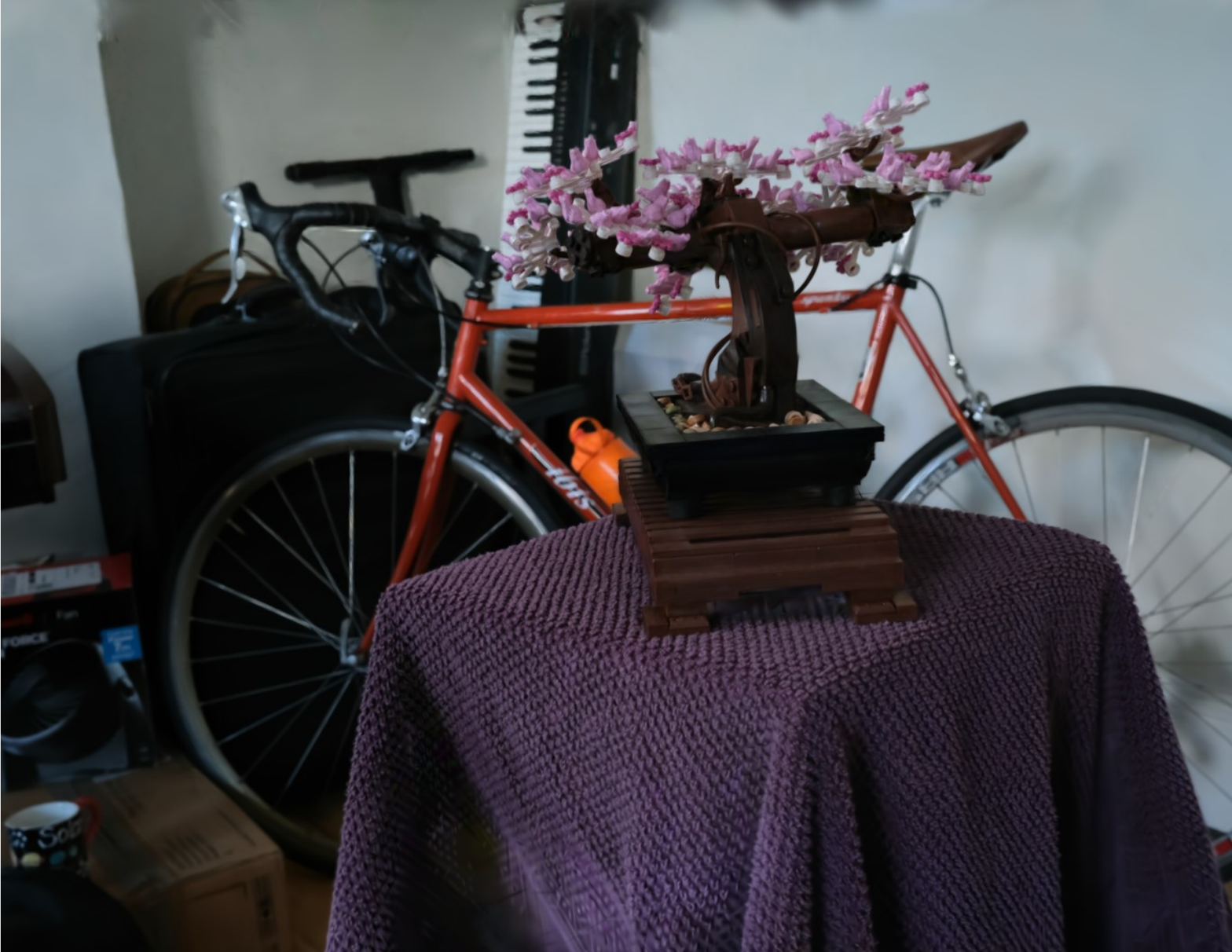}
    \end{subfigure}
    \begin{subfigure}[b]{0.23\textwidth}
        \includegraphics[width=\textwidth]{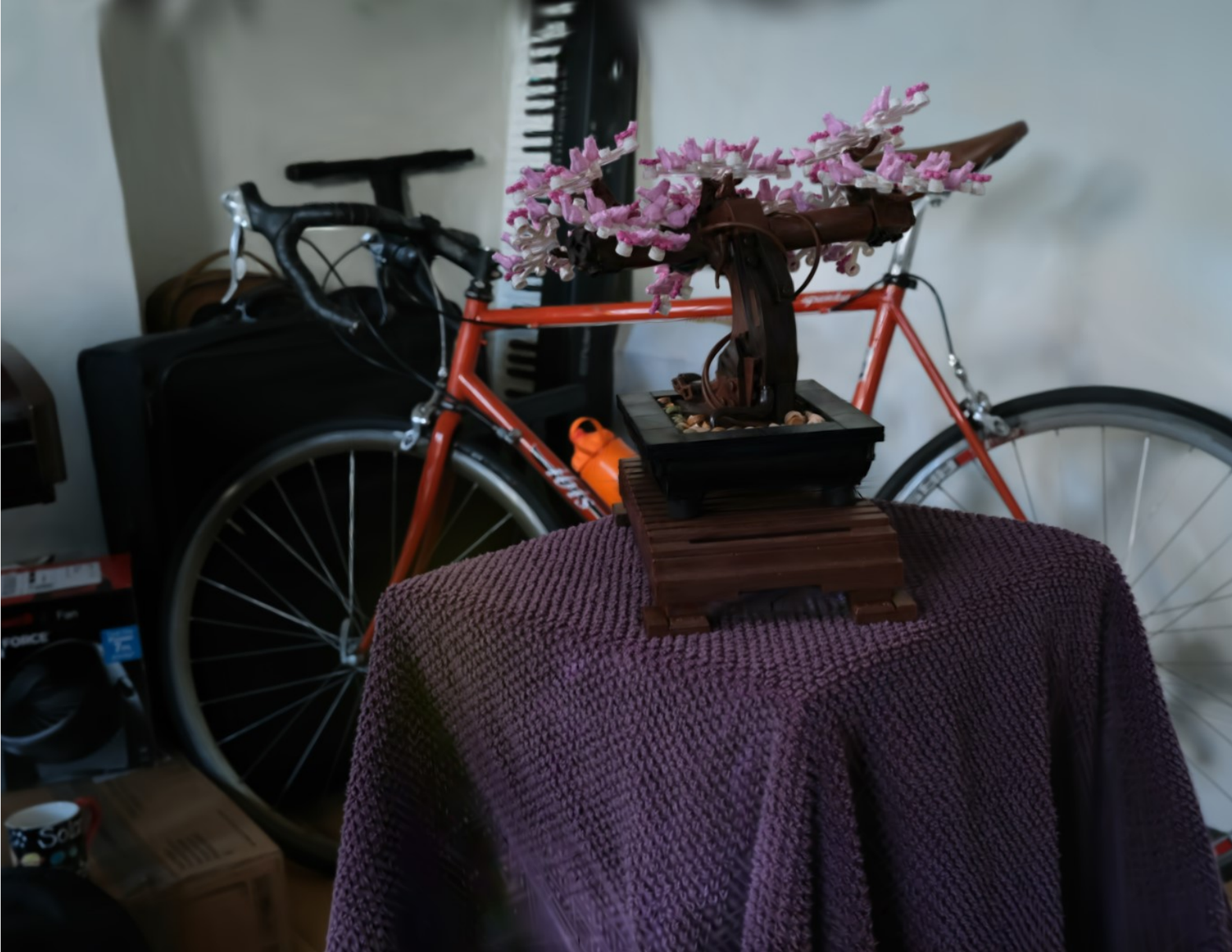}
    \end{subfigure}
    \begin{subfigure}[b]{0.23\textwidth}
        \includegraphics[width=\textwidth]{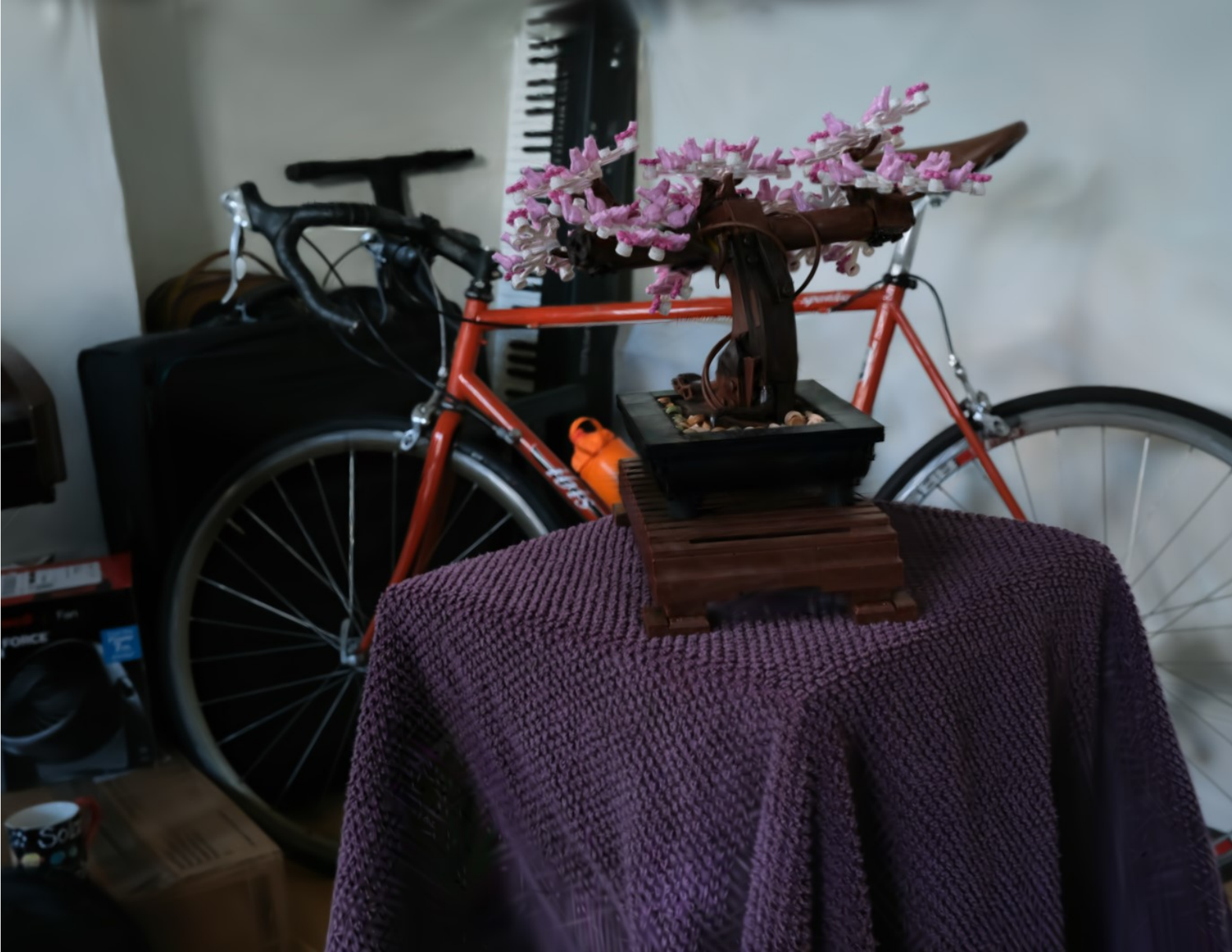}
    \end{subfigure}
    \par\medskip

    \rotatebox[origin=t, y=1.3cm]{90}{\textbf{Counter}}
    \begin{subfigure}[b]{0.23\textwidth}
        \includegraphics[width=\textwidth]{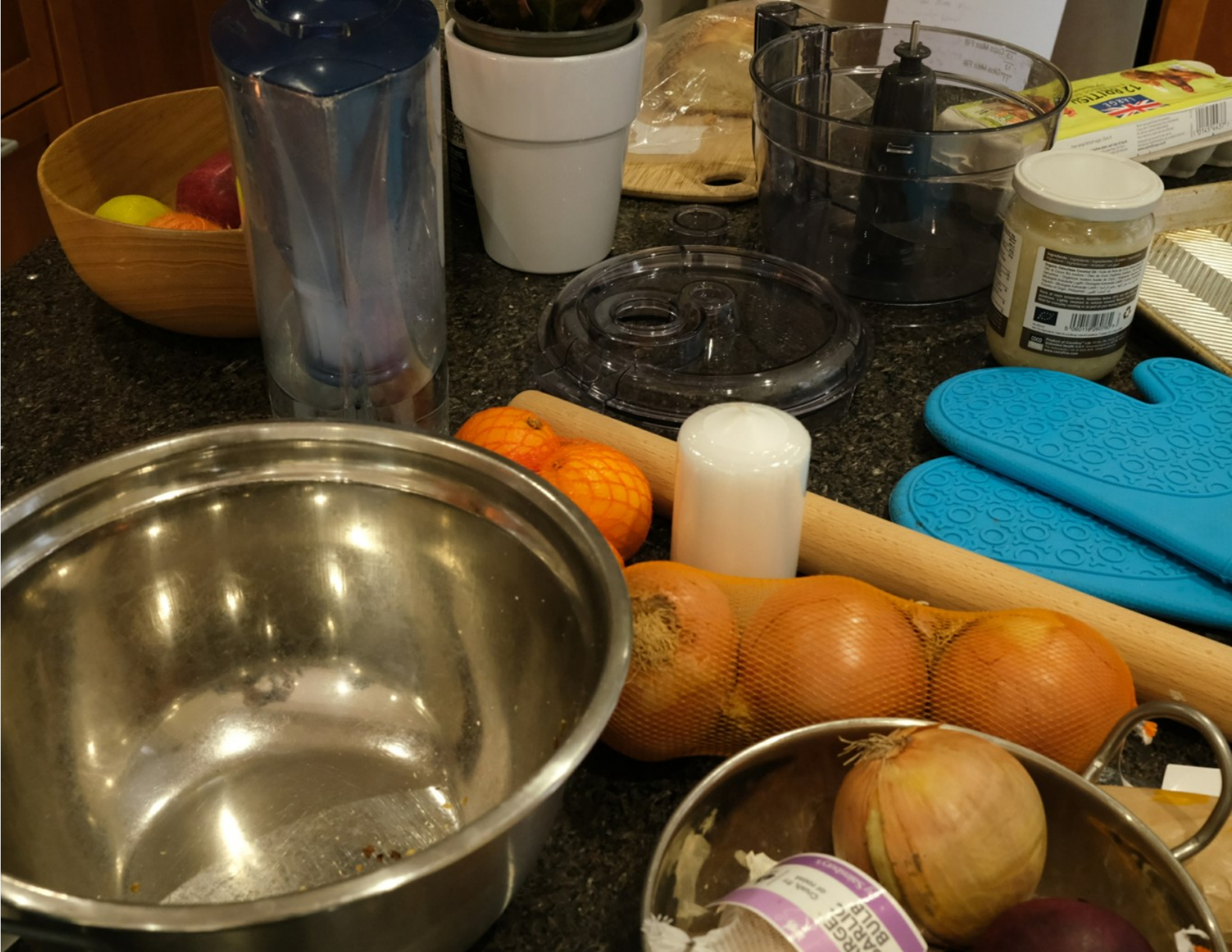}
    \end{subfigure}
    \begin{subfigure}[b]{0.23\textwidth}
        \includegraphics[width=\textwidth]{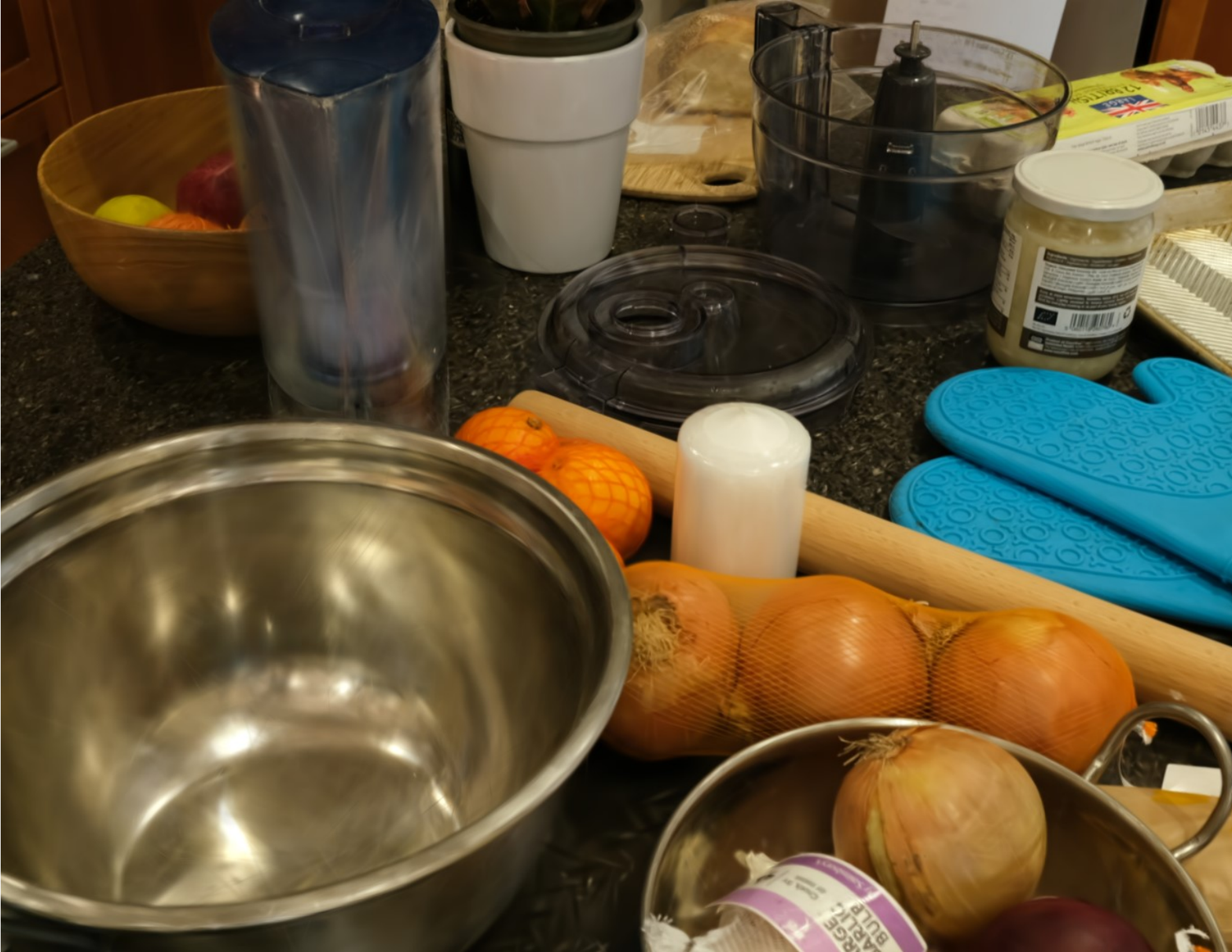}
    \end{subfigure}
    \begin{subfigure}[b]{0.23\textwidth}
        \includegraphics[width=\textwidth]{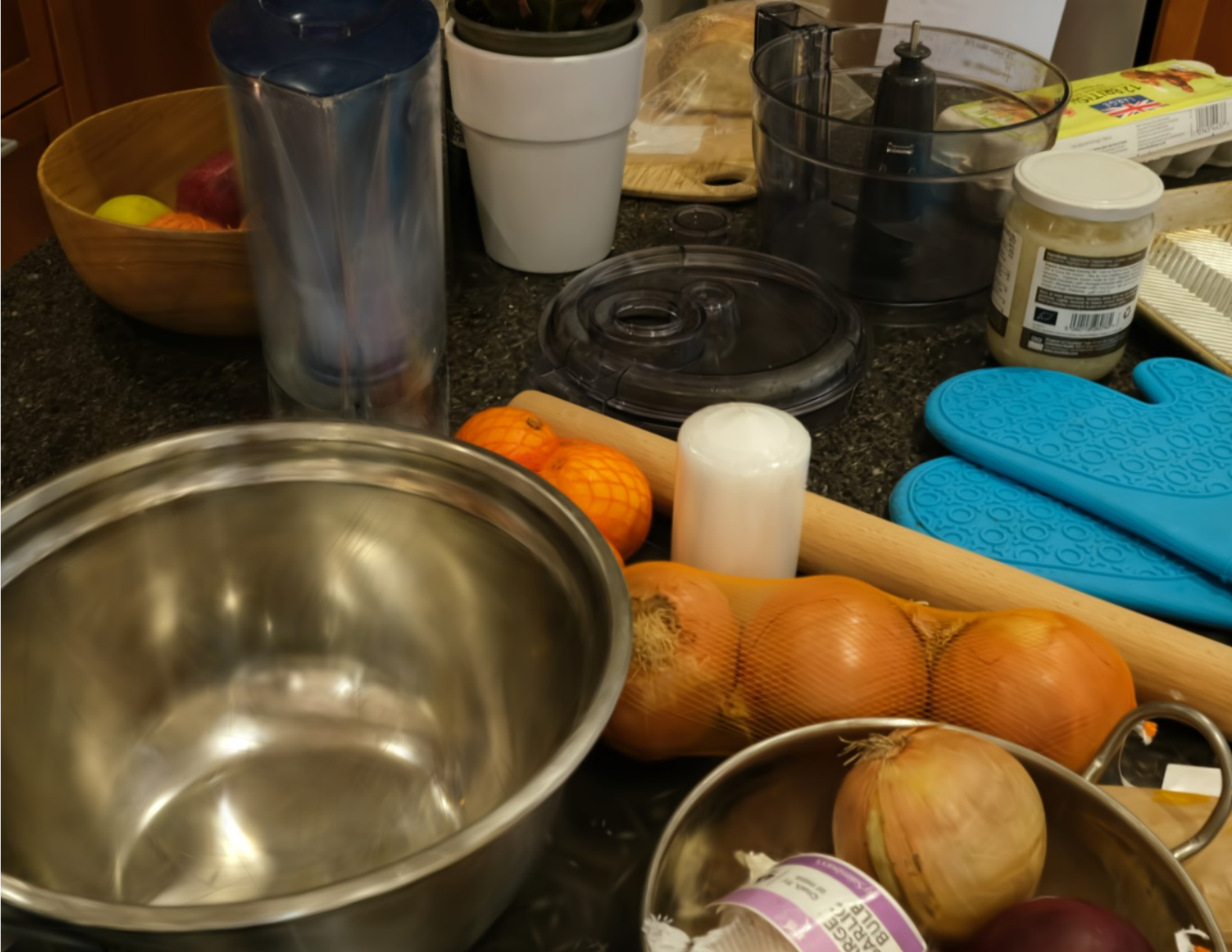}
    \end{subfigure}
    \begin{subfigure}[b]{0.23\textwidth}
        \includegraphics[width=\textwidth]{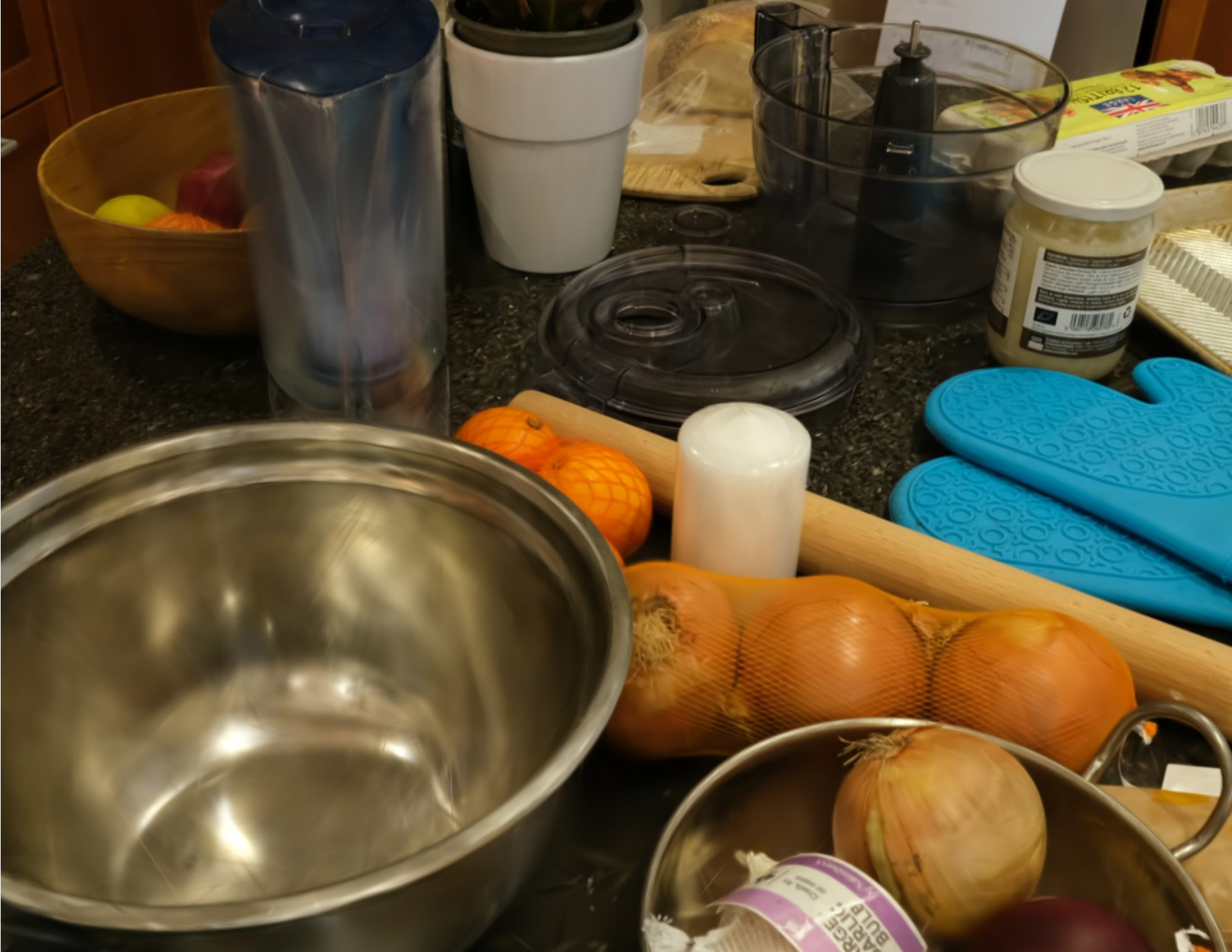}
    \end{subfigure}
    \par\medskip

    \rotatebox[origin=t, y=1.3cm]{90}{\textbf{Kitchen}}
    \begin{subfigure}[b]{0.23\textwidth}
        \includegraphics[width=\textwidth]{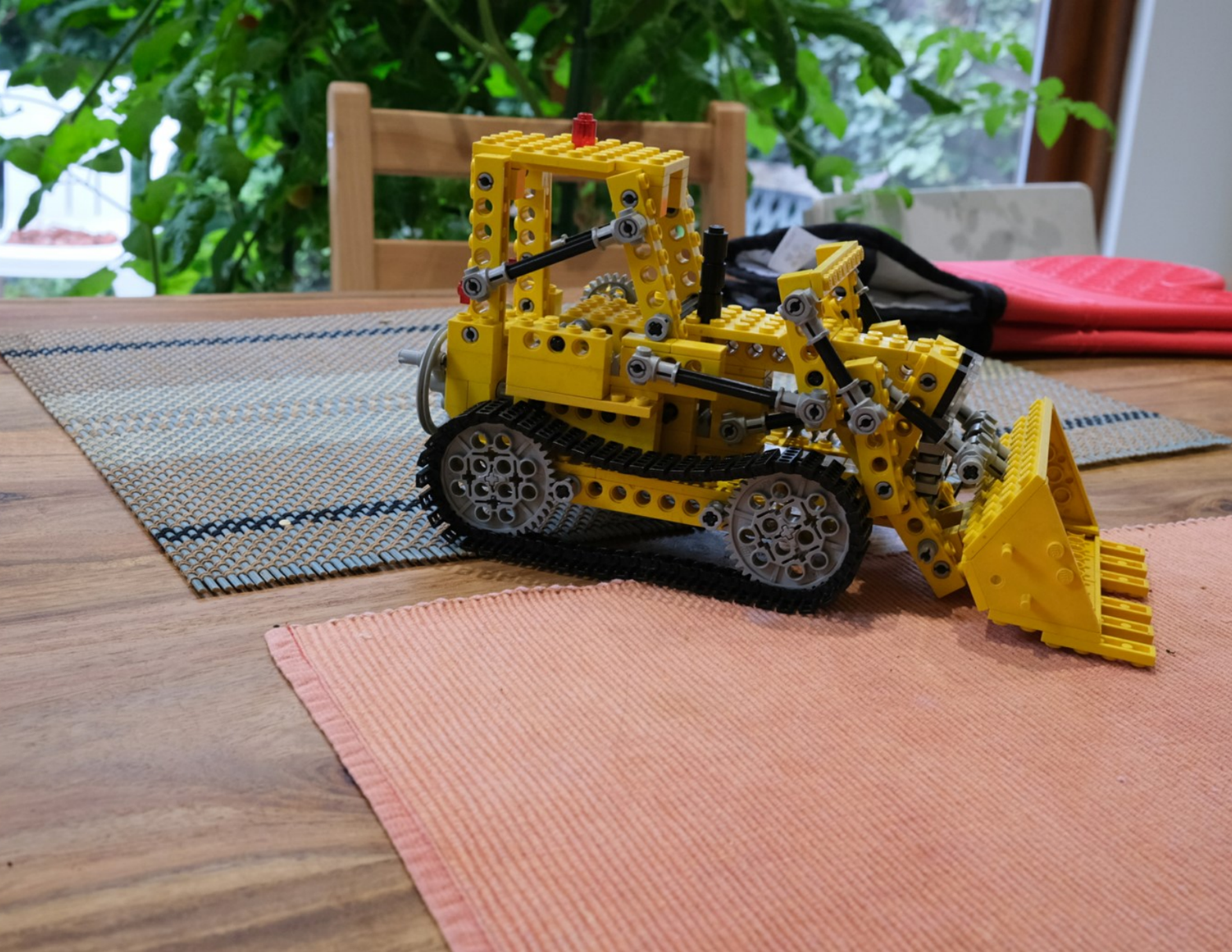}
    \end{subfigure}
    \begin{subfigure}[b]{0.23\textwidth}
        \includegraphics[width=\textwidth]{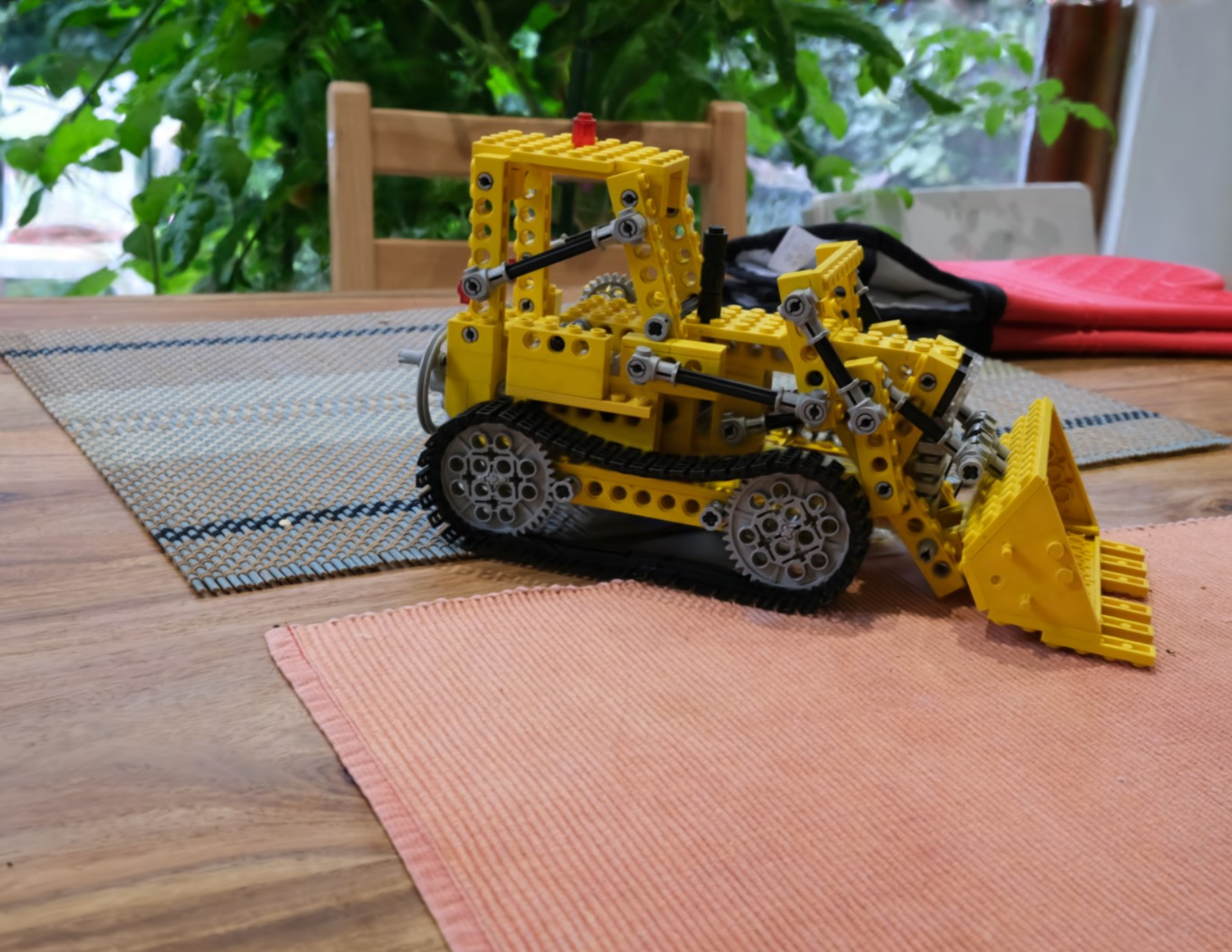}
    \end{subfigure}
    \begin{subfigure}[b]{0.23\textwidth}
        \includegraphics[width=\textwidth]{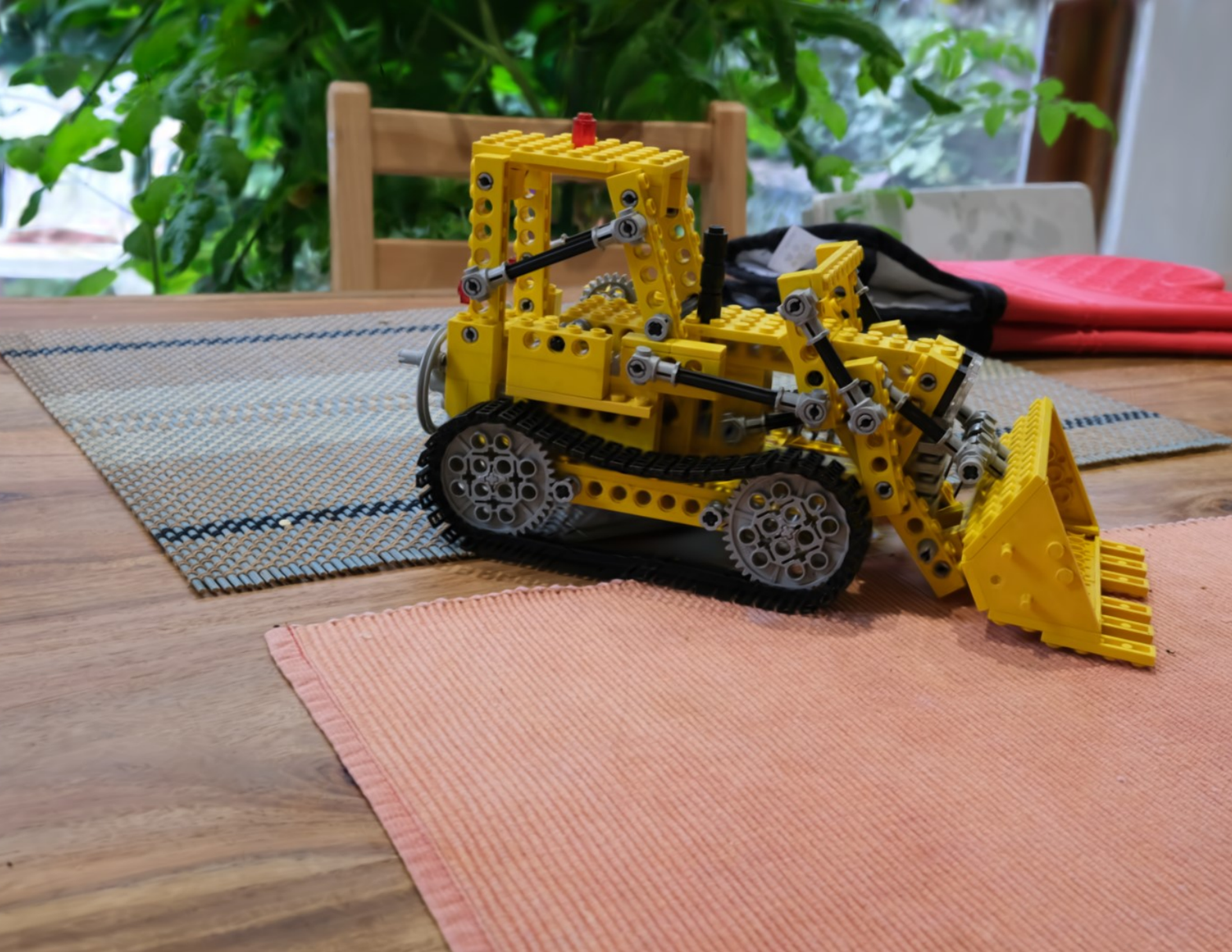}
    \end{subfigure}
    \begin{subfigure}[b]{0.23\textwidth}
        \includegraphics[width=\textwidth]{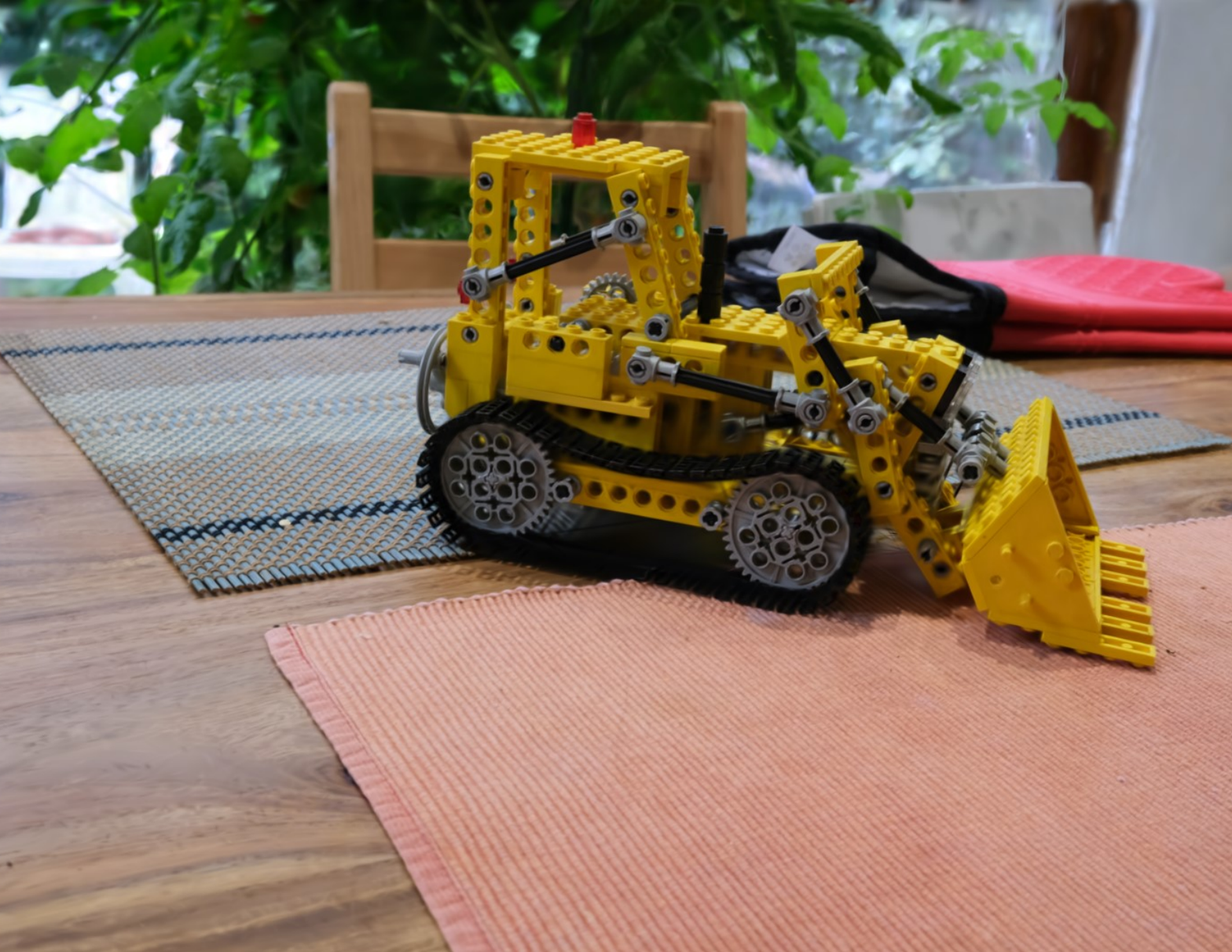}
    \end{subfigure}
    \par\medskip

    \rotatebox[origin=t, y=1.3cm]{90}{\textbf{Room}}
    \begin{subfigure}[b]{0.23\textwidth}
        \includegraphics[width=\textwidth]{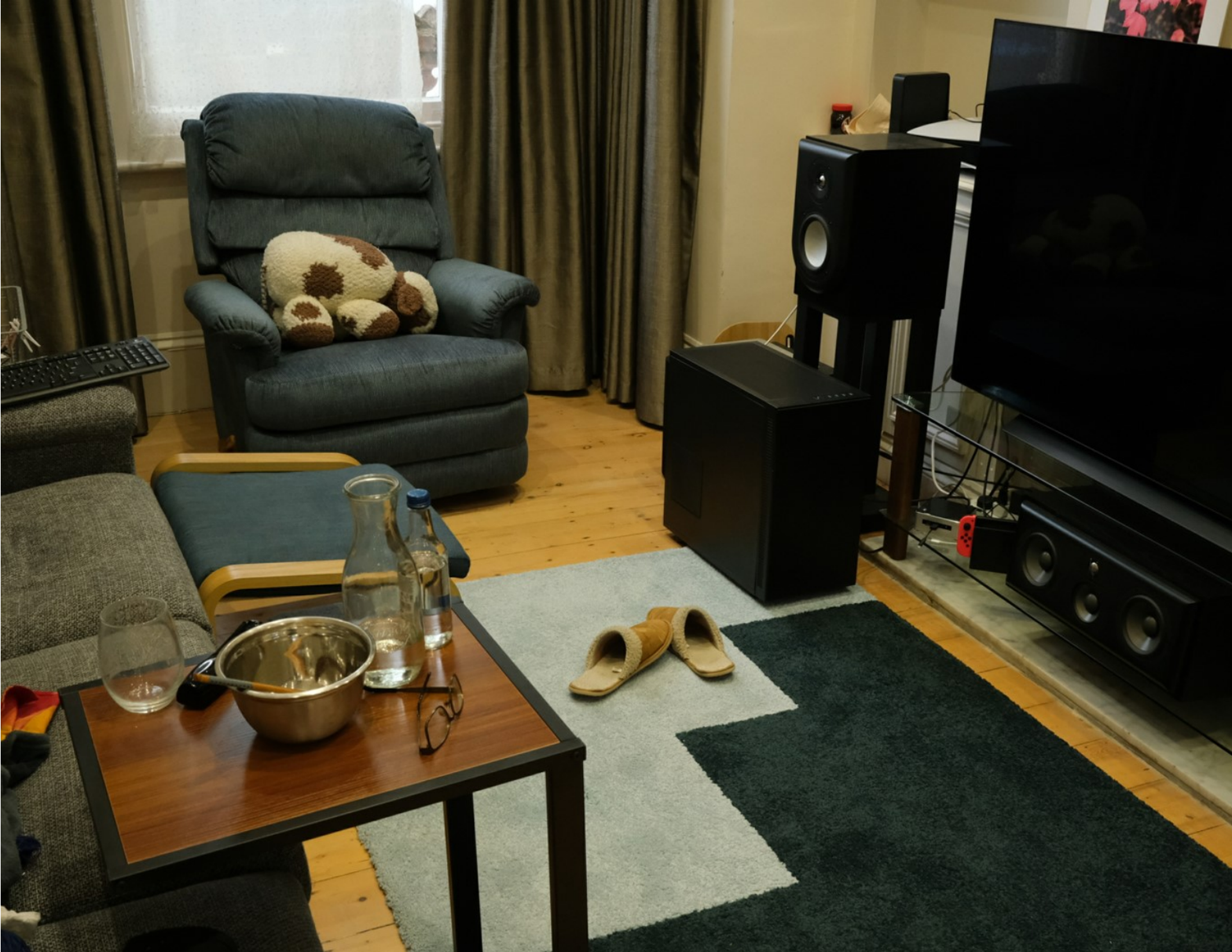}
    \end{subfigure}
    \begin{subfigure}[b]{0.23\textwidth}
        \includegraphics[width=\textwidth]{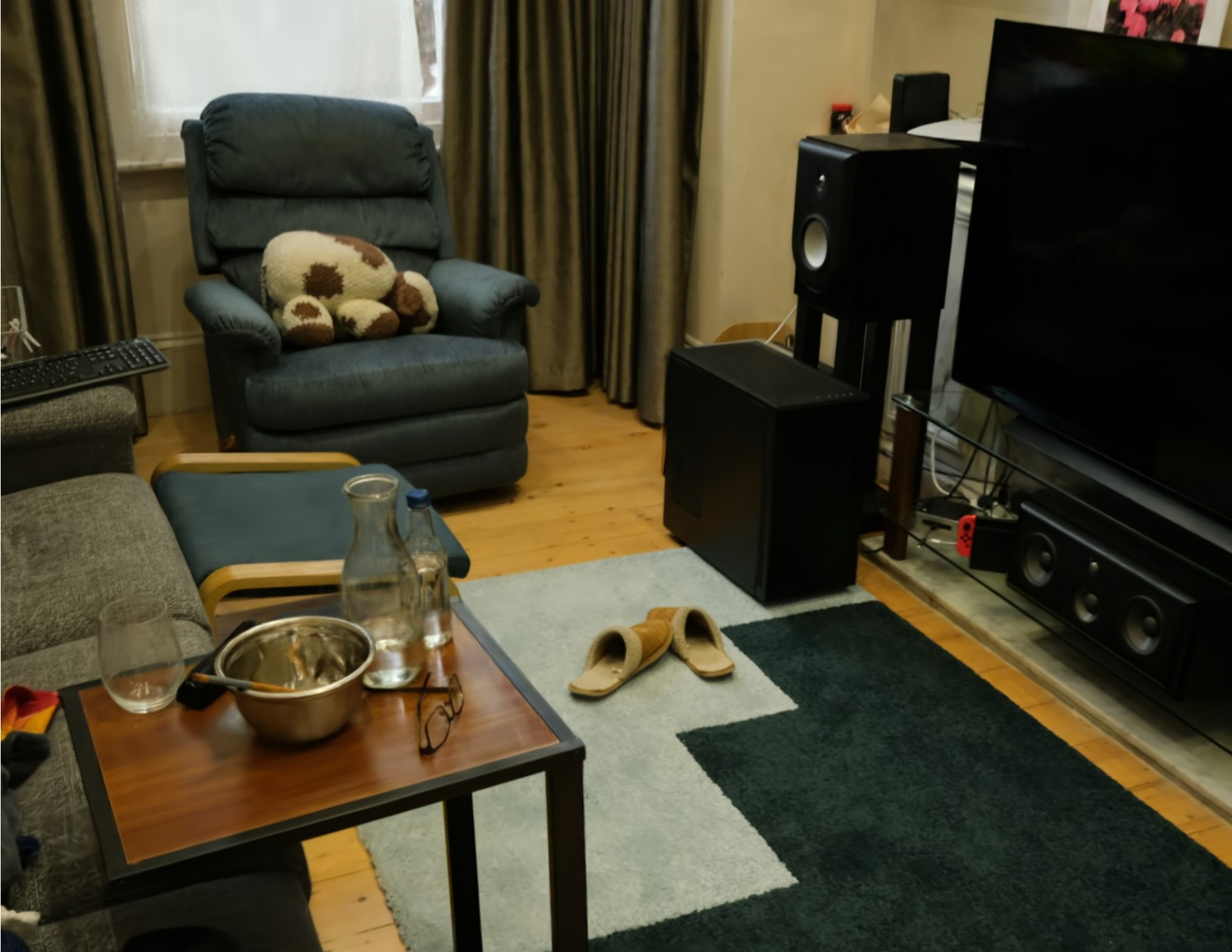}
    \end{subfigure}
    \begin{subfigure}[b]{0.23\textwidth}
        \includegraphics[width=\textwidth]{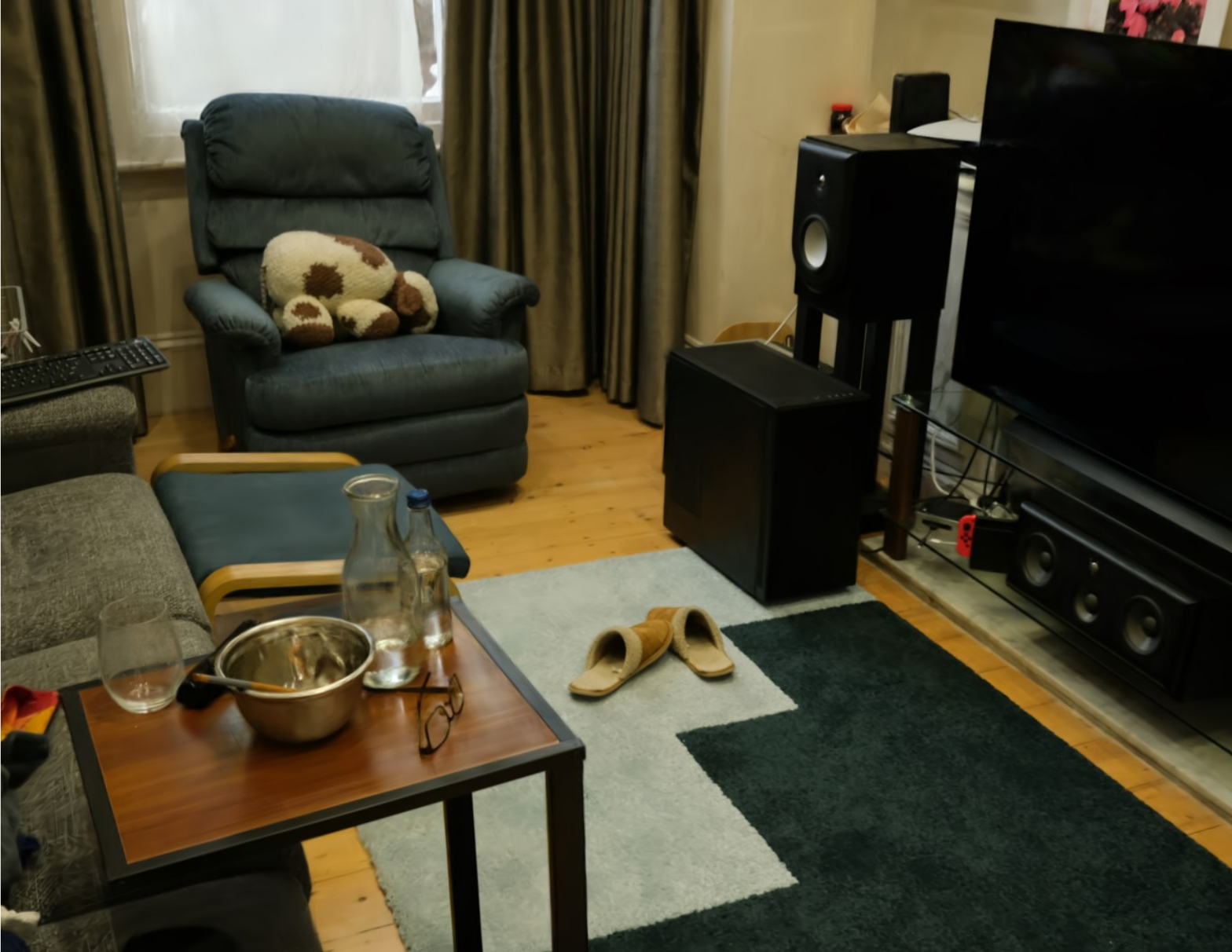}
    \end{subfigure}
    \begin{subfigure}[b]{0.23\textwidth}
        \includegraphics[width=\textwidth]{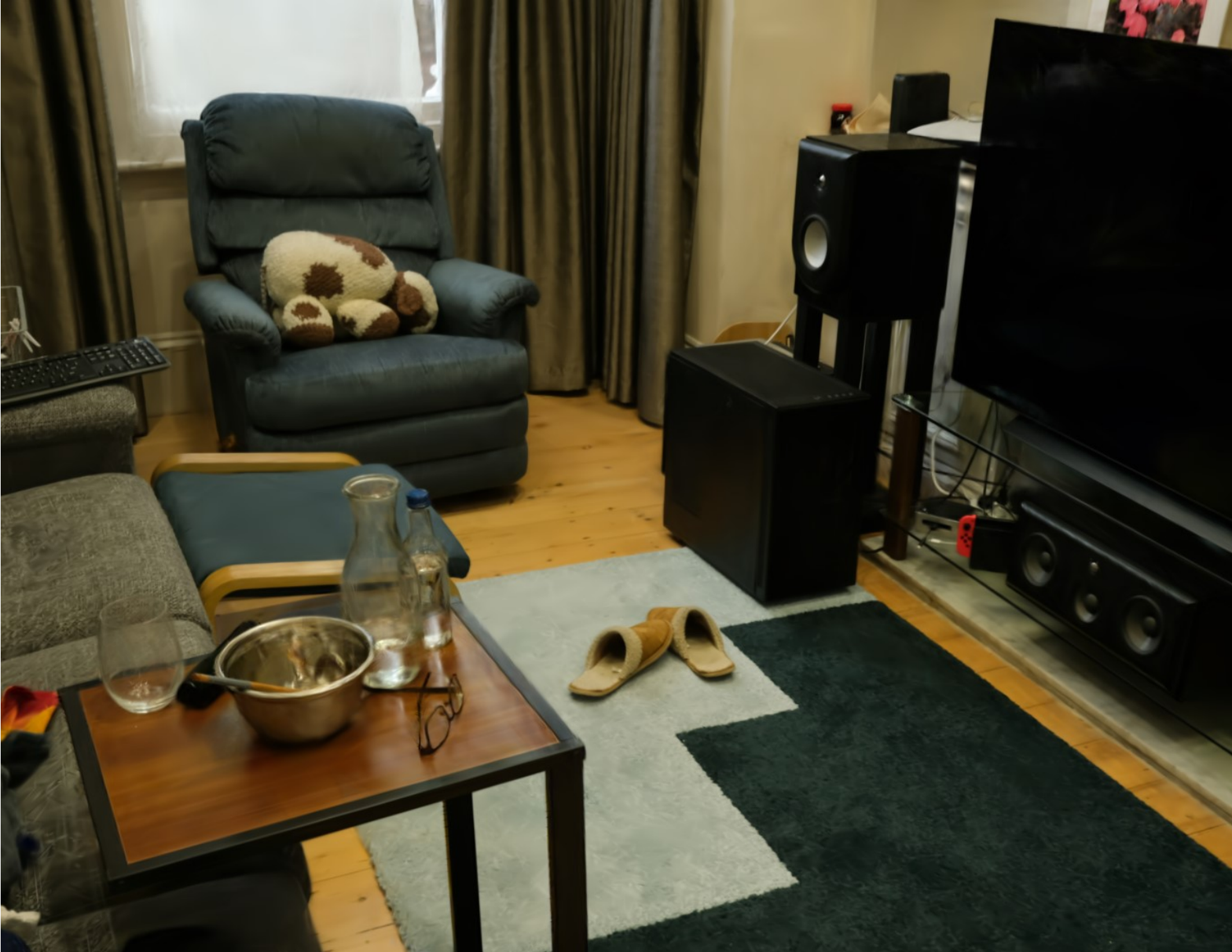}
    \end{subfigure}
    \par\medskip

    \rotatebox[origin=t, y=1.3cm]{90}{\textbf{Stump}}
    \begin{subfigure}[b]{0.23\textwidth}
        \includegraphics[width=\textwidth]{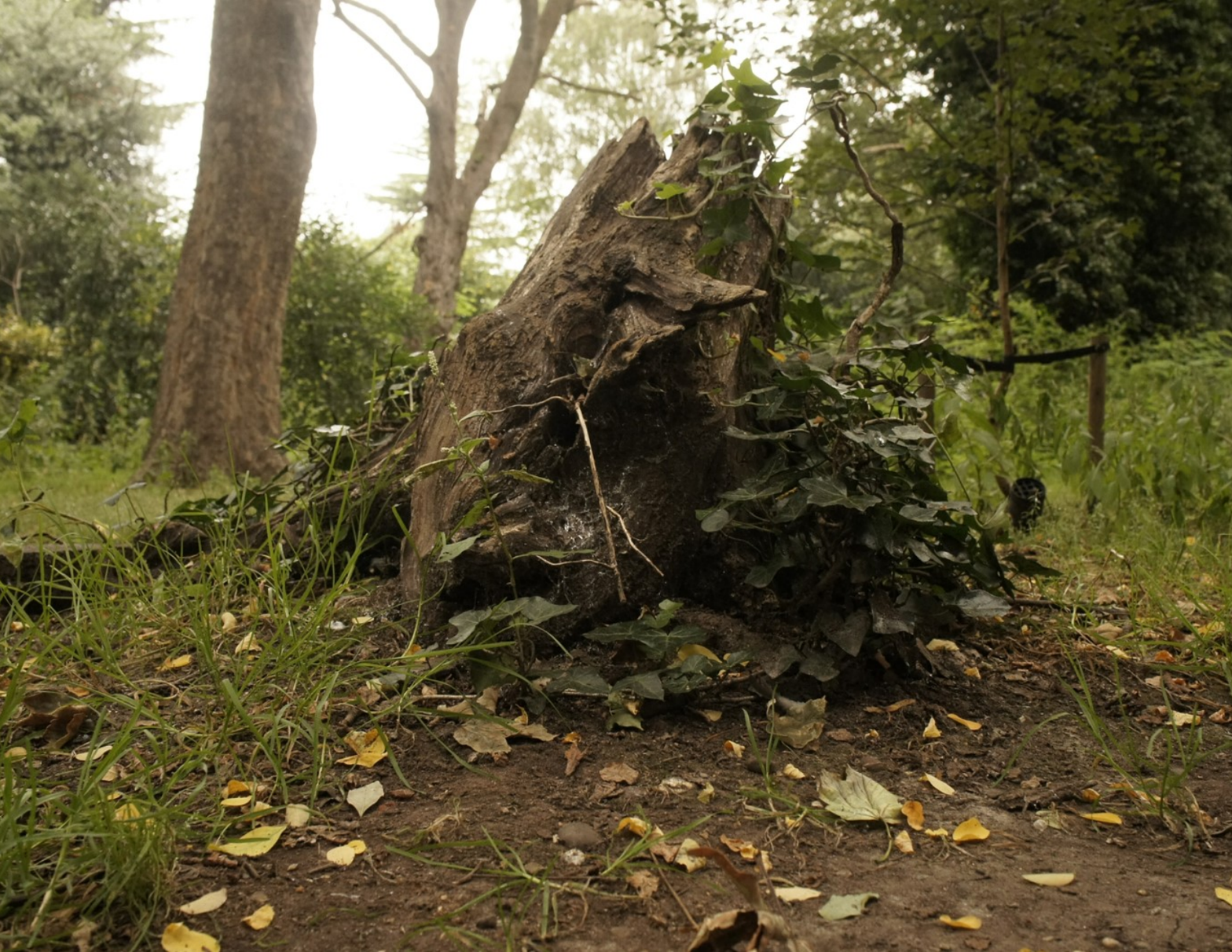}
    \end{subfigure}
    \begin{subfigure}[b]{0.23\textwidth}
        \includegraphics[width=\textwidth]{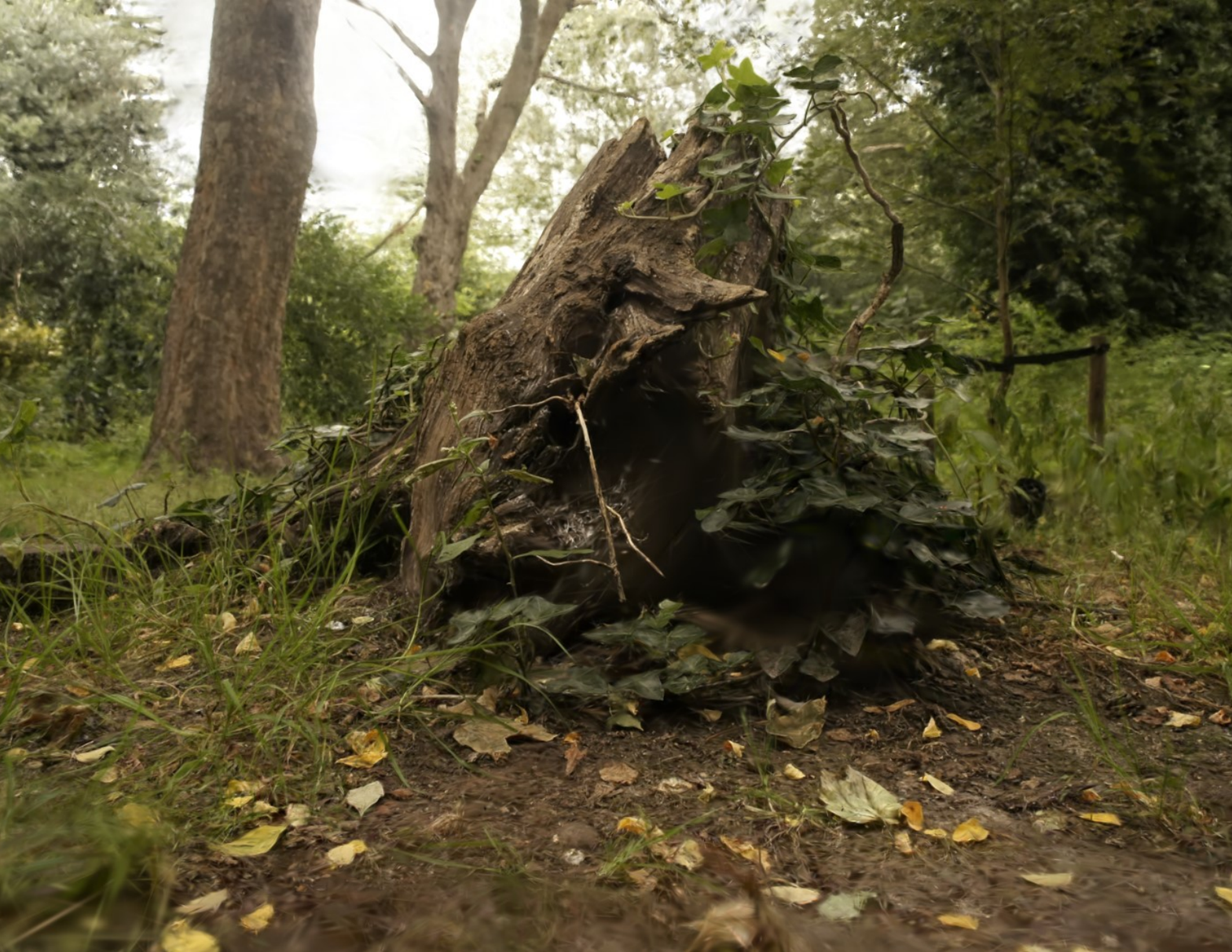}
    \end{subfigure}
    \begin{subfigure}[b]{0.23\textwidth}
        \includegraphics[width=\textwidth]{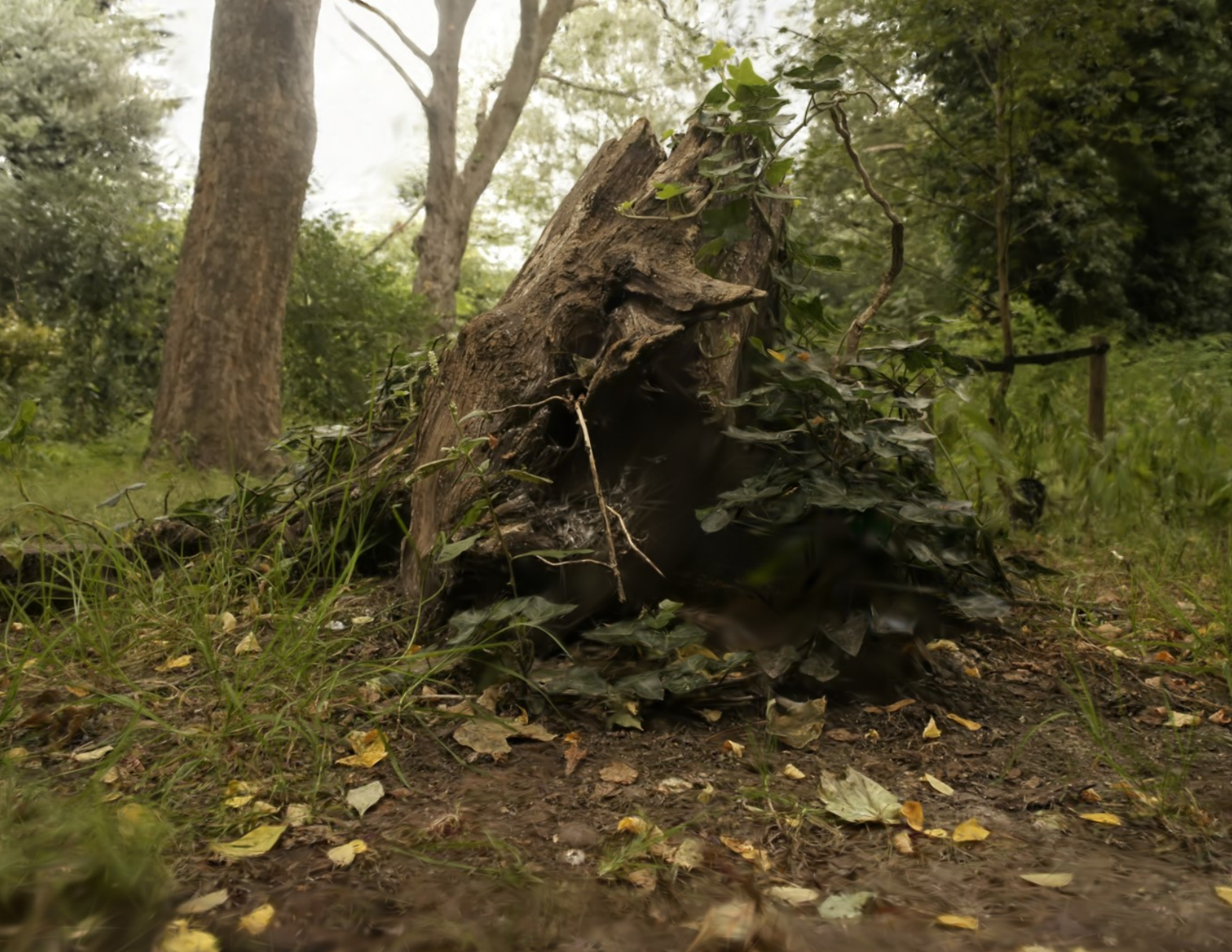}
    \end{subfigure}
    \begin{subfigure}[b]{0.23\textwidth}
        \includegraphics[width=\textwidth]{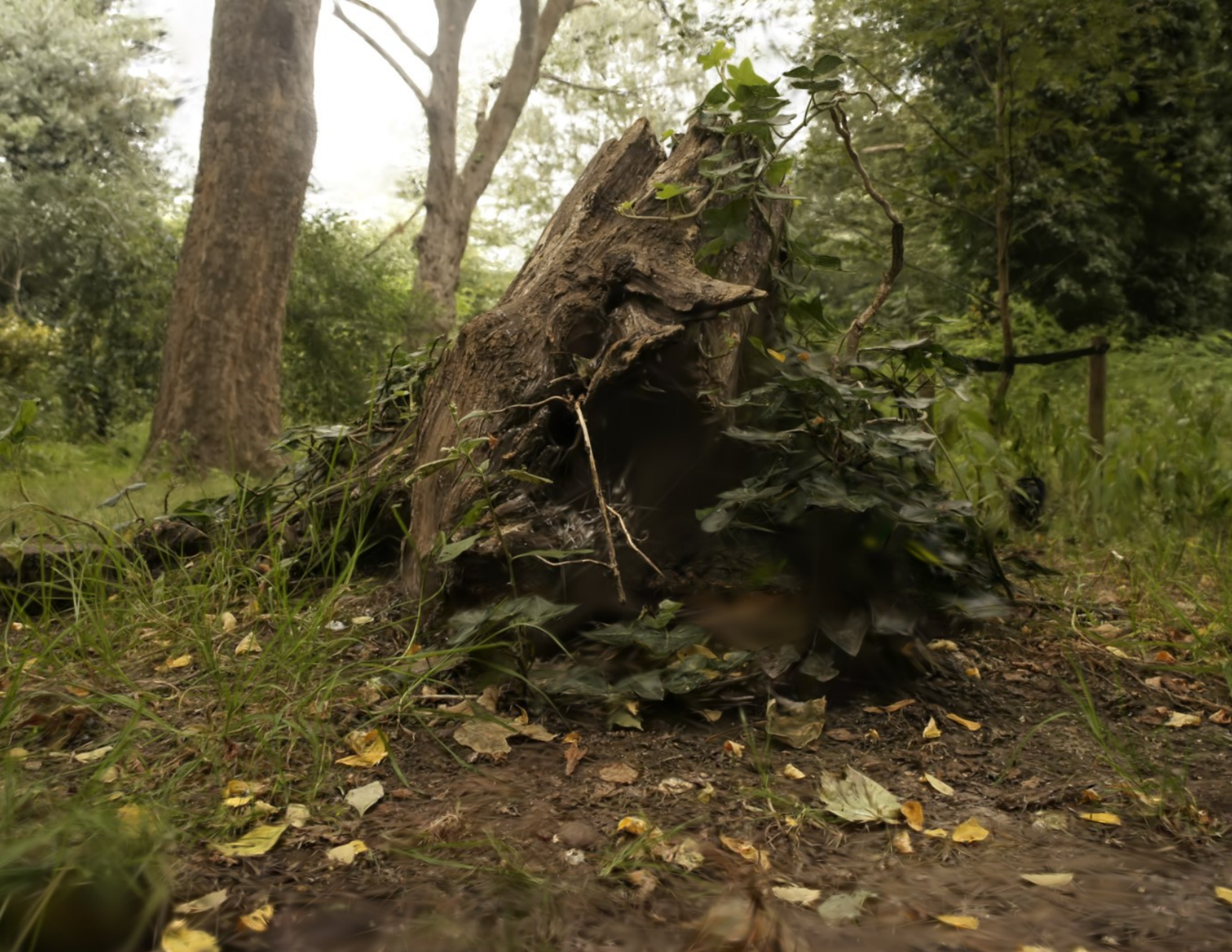}
    \end{subfigure}

    \rotatebox[origin=t, y=1.3cm]{90}{\textbf{Garden}}
    \begin{subfigure}[b]{0.23\textwidth}
        \includegraphics[width=\textwidth]{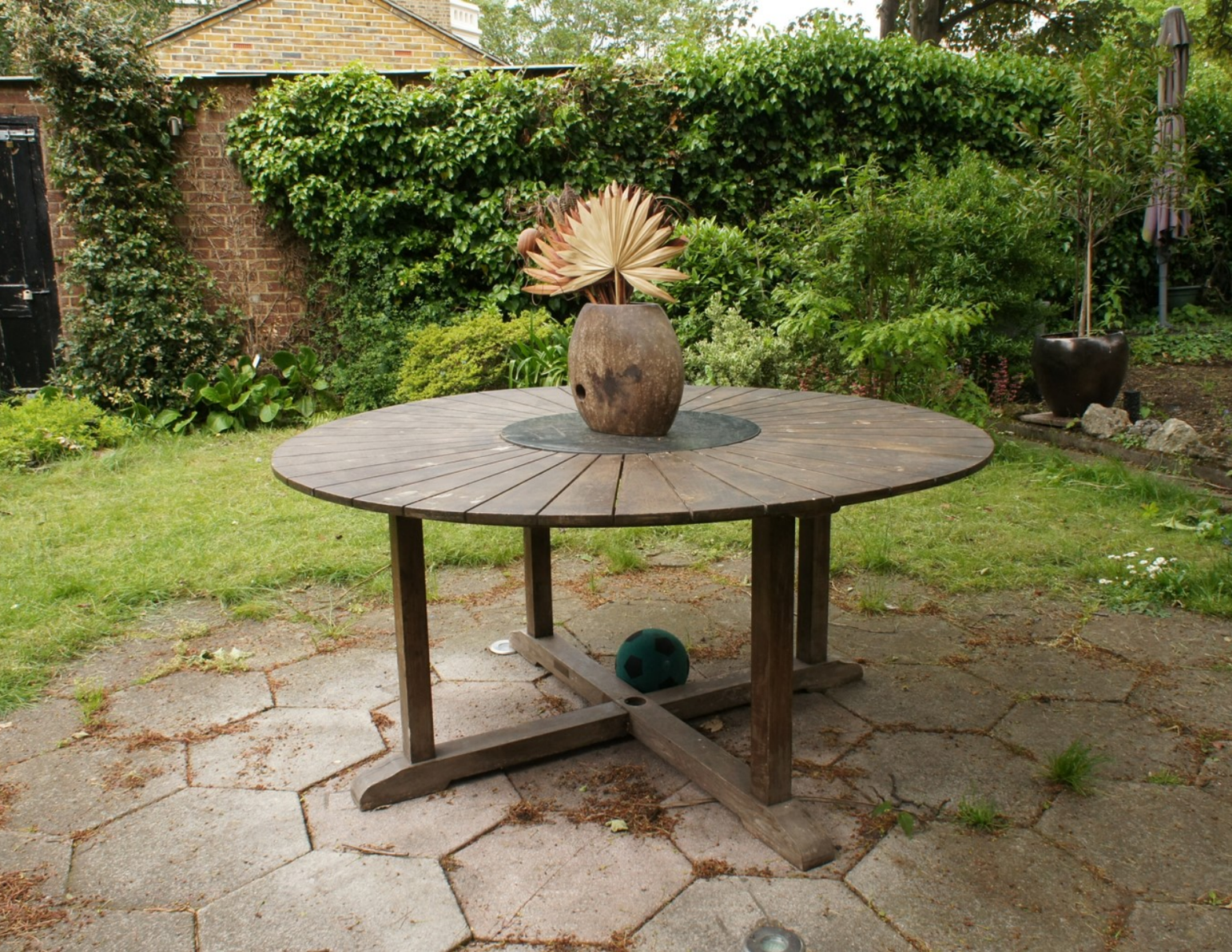}
    \end{subfigure}
    \begin{subfigure}[b]{0.23\textwidth}
        \includegraphics[width=\textwidth]{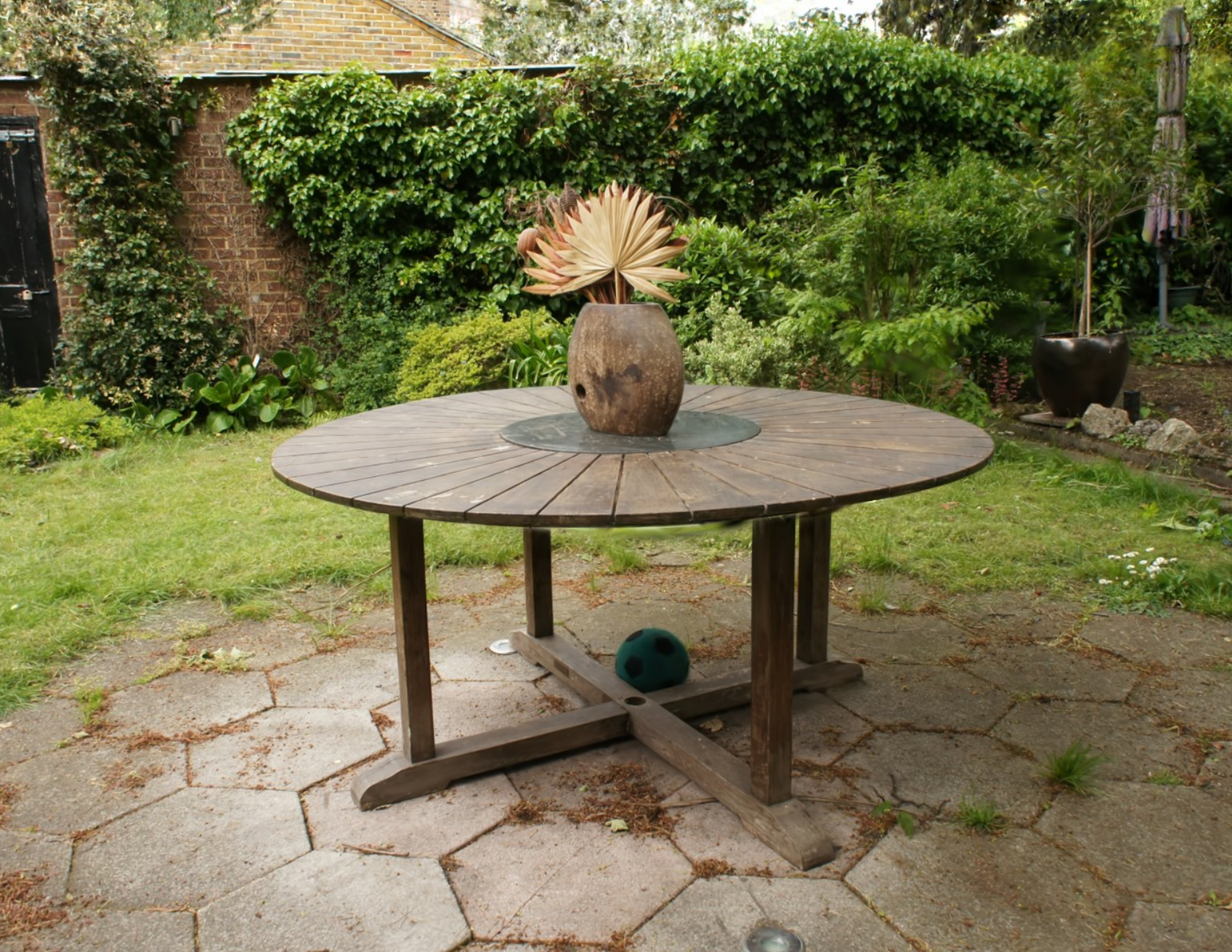}
    \end{subfigure}
    \begin{subfigure}[b]{0.23\textwidth}
        \includegraphics[width=\textwidth]{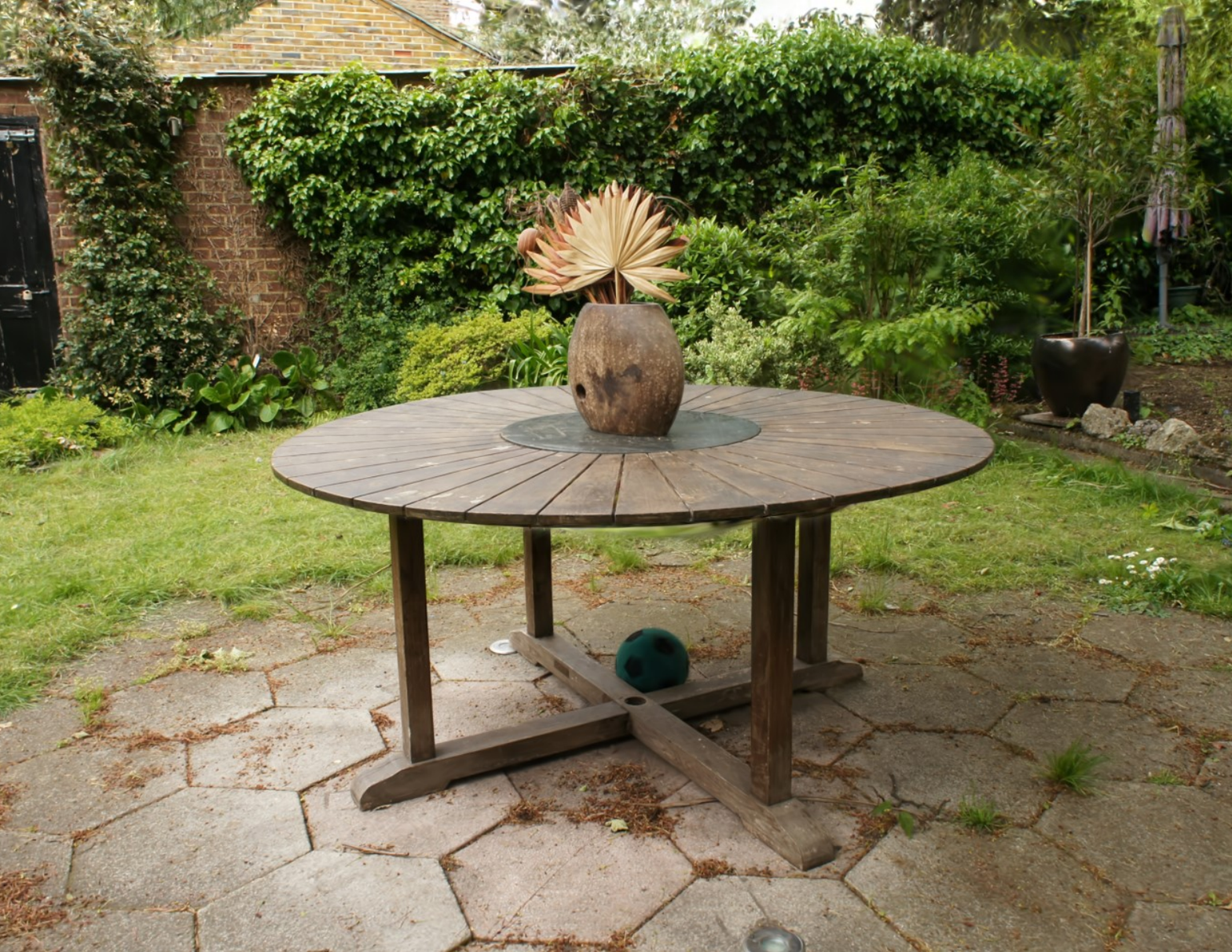}
    \end{subfigure}
    \begin{subfigure}[b]{0.23\textwidth}
        \includegraphics[width=\textwidth]{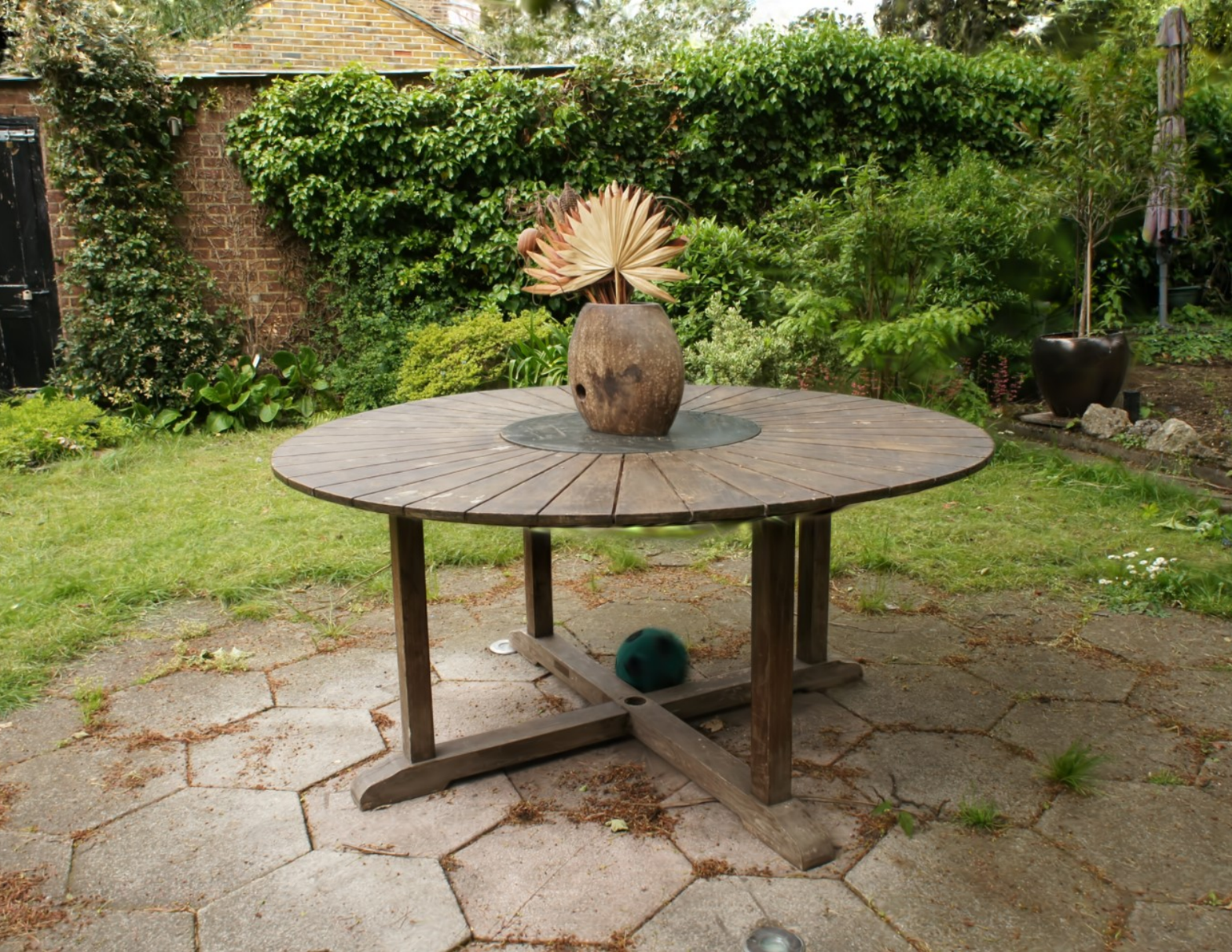}
    \end{subfigure}
    \par\medskip
    \end{figure}
    
    \clearpage
    
    \begin{figure}[t!]\ContinuedFloat
    \rotatebox[origin=t, y=1.3cm]{90}{\textbf{Flowers}}
    \begin{subfigure}[b]{0.23\textwidth}
        \includegraphics[width=\textwidth]{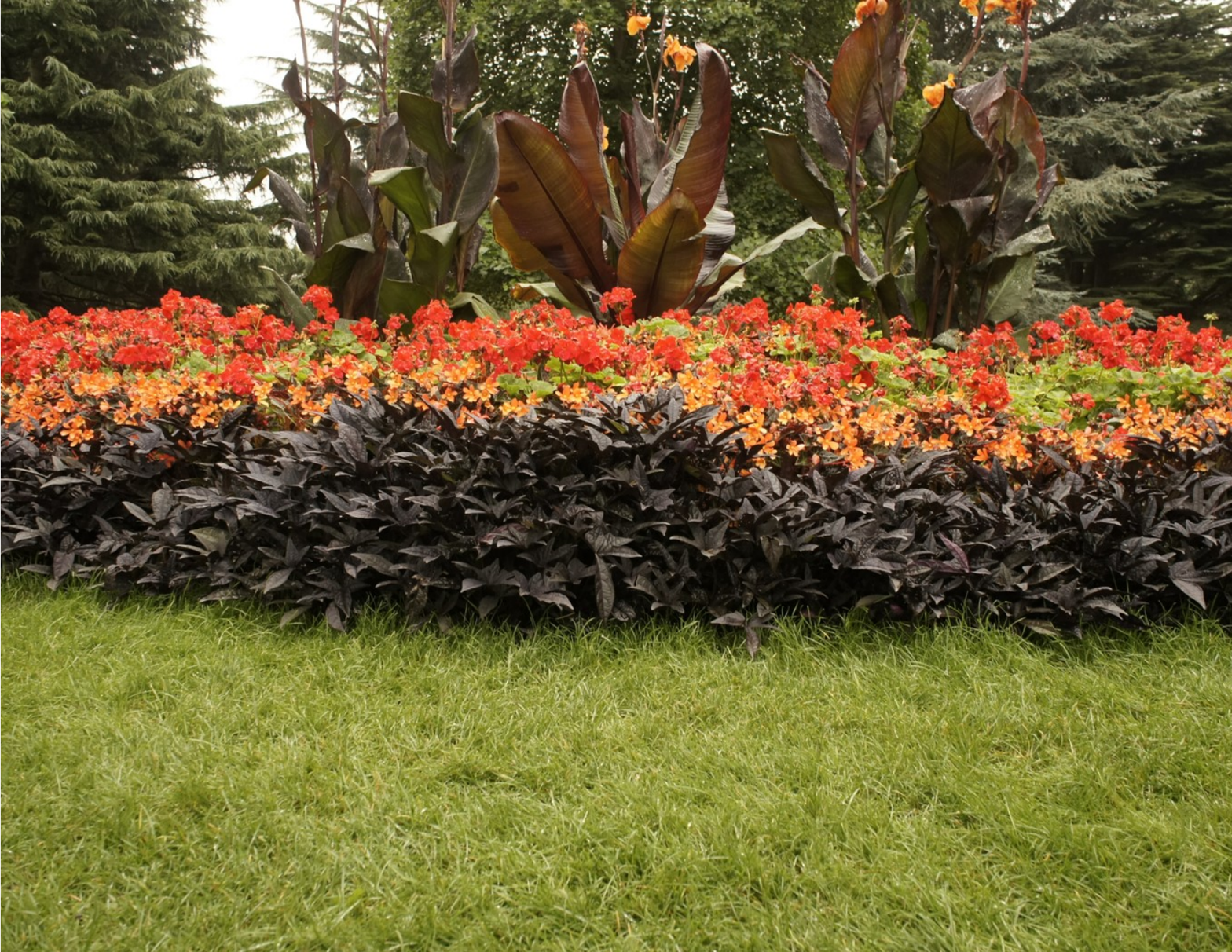}
    \end{subfigure}
    \begin{subfigure}[b]{0.23\textwidth}
        \includegraphics[width=\textwidth]{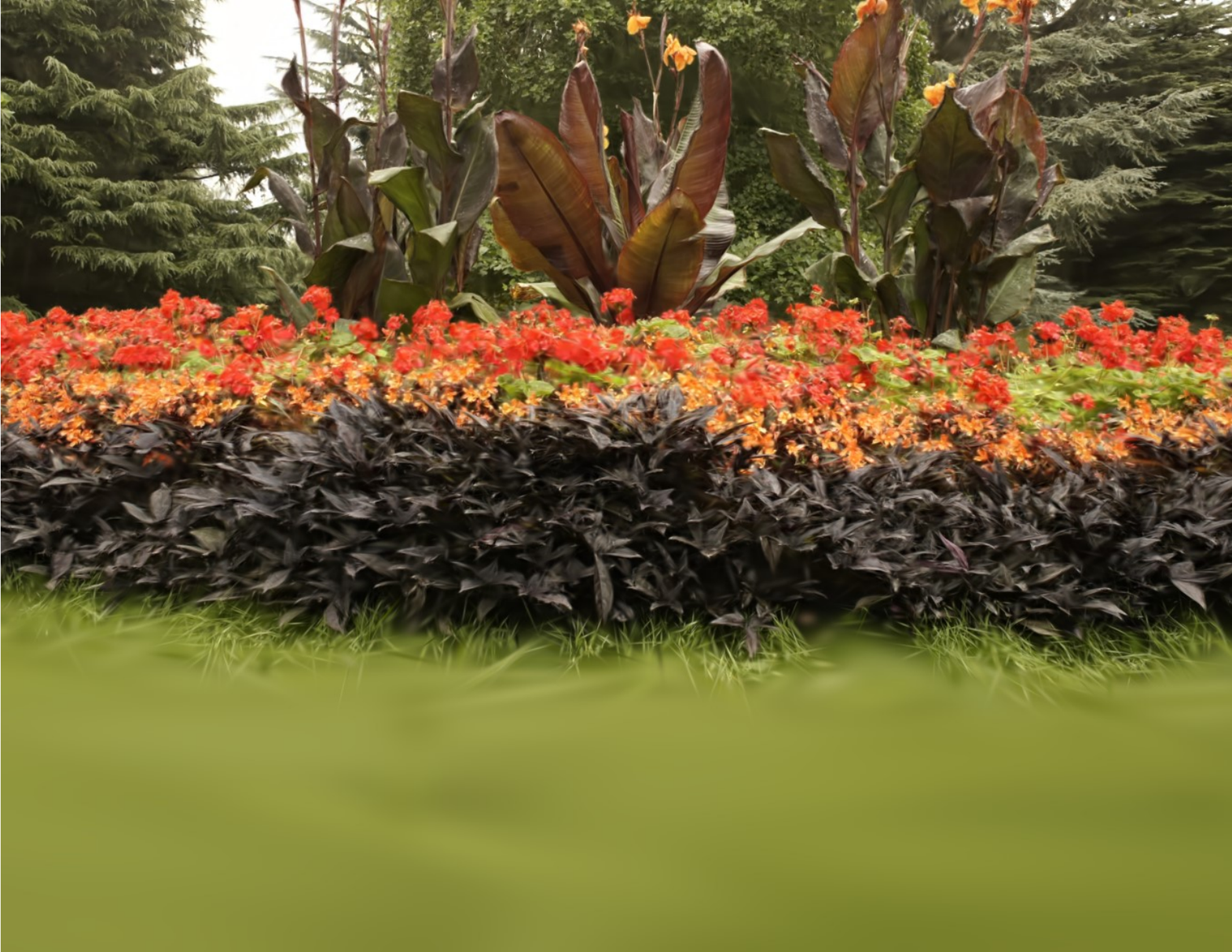}
    \end{subfigure}
    \begin{subfigure}[b]{0.23\textwidth}
        \includegraphics[width=\textwidth]{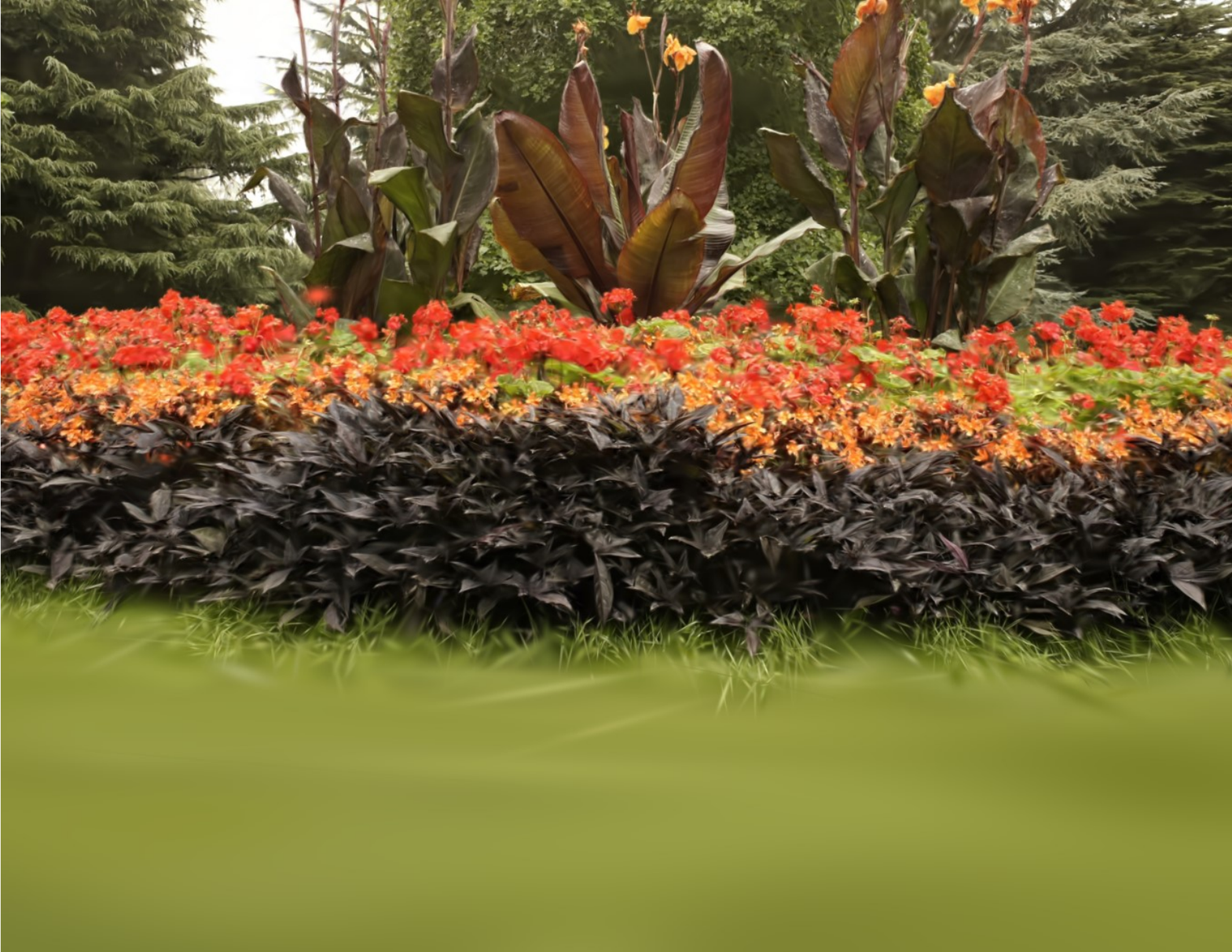}
    \end{subfigure}
    \begin{subfigure}[b]{0.23\textwidth}
        \includegraphics[width=\textwidth]{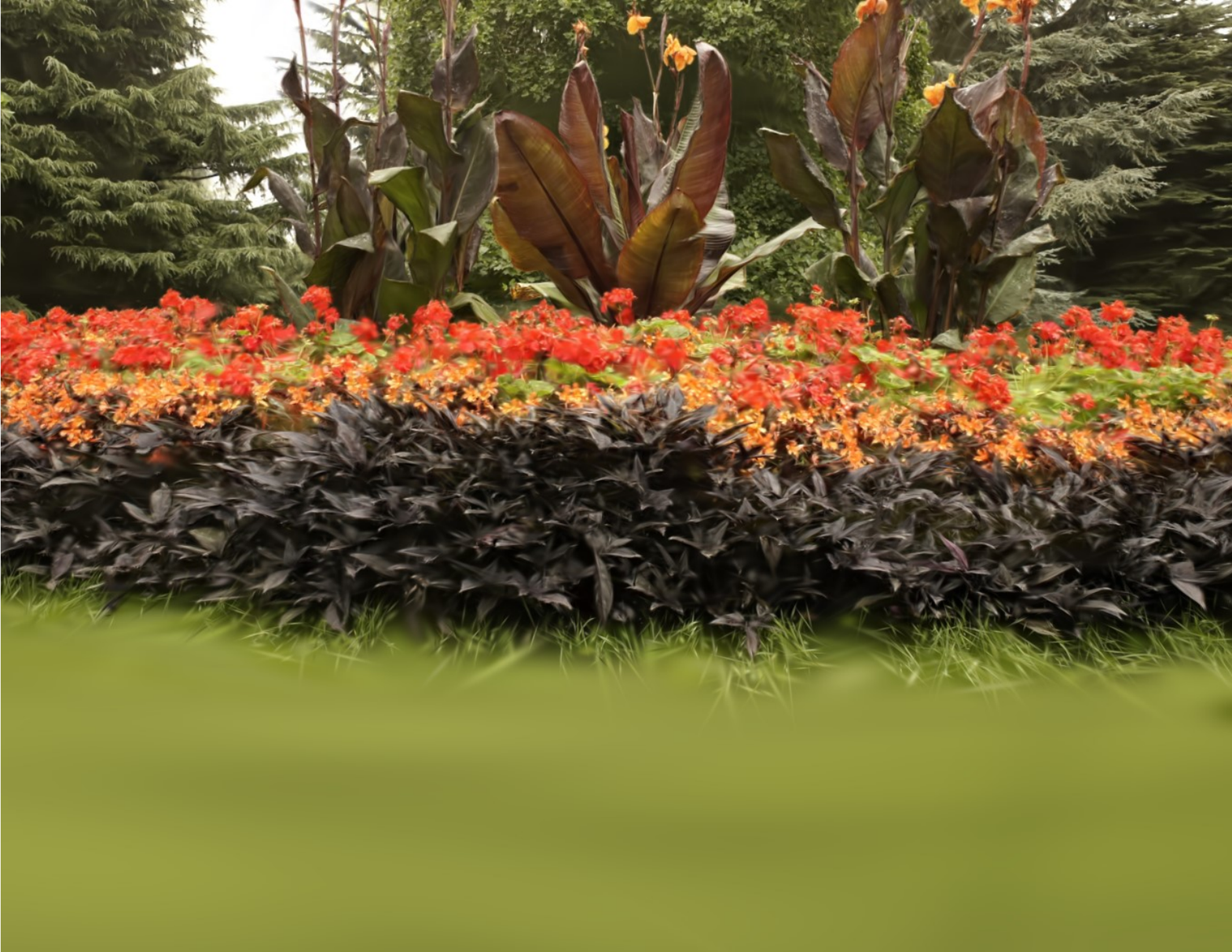}
    \end{subfigure}

    \rotatebox[origin=t, y=1.3cm]{90}{\textbf{Treehill}}
    \begin{subfigure}[b]{0.23\textwidth}
        \includegraphics[width=\textwidth]{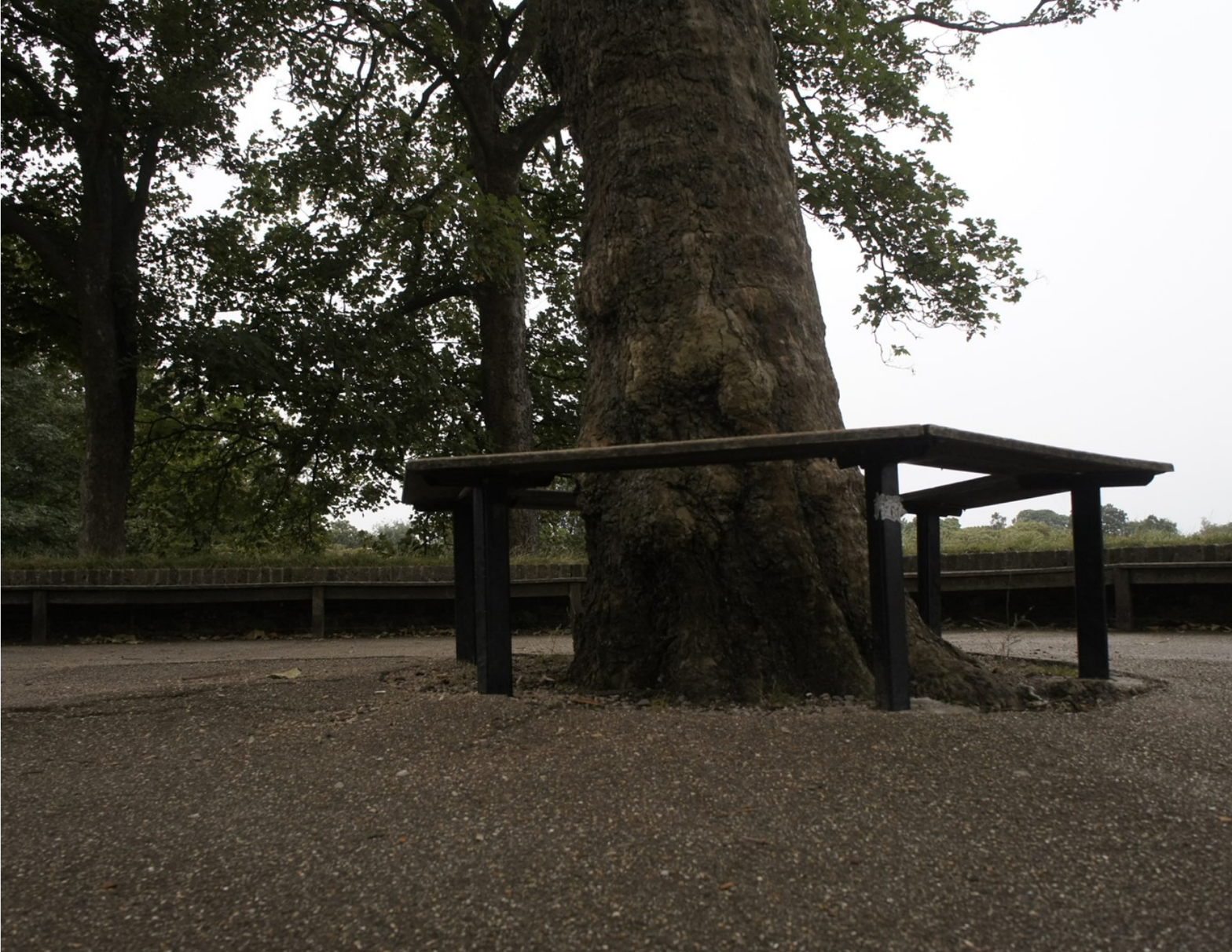}
    \end{subfigure}
    \begin{subfigure}[b]{0.23\textwidth}
        \includegraphics[width=\textwidth]{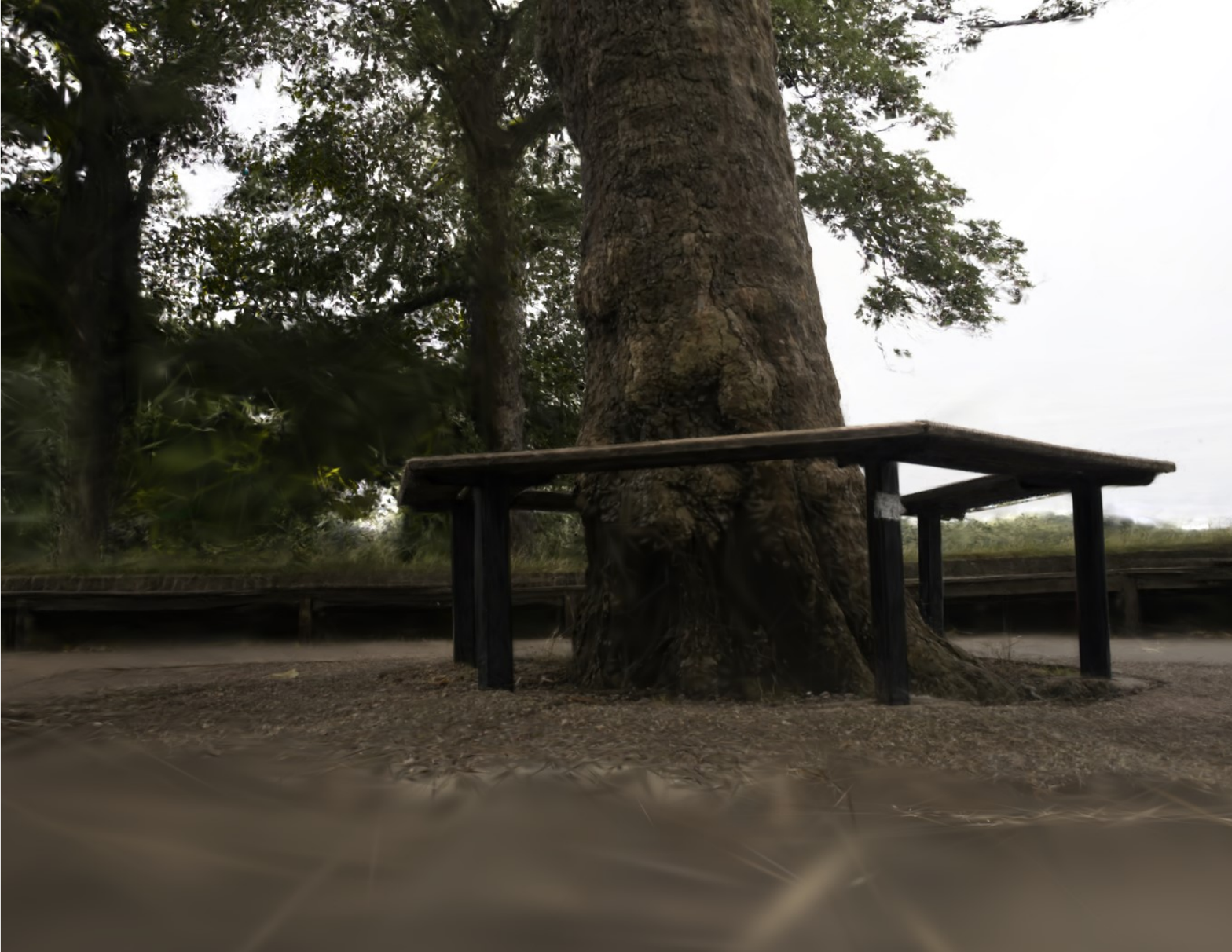}
    \end{subfigure}
    \begin{subfigure}[b]{0.23\textwidth}
        \includegraphics[width=\textwidth]{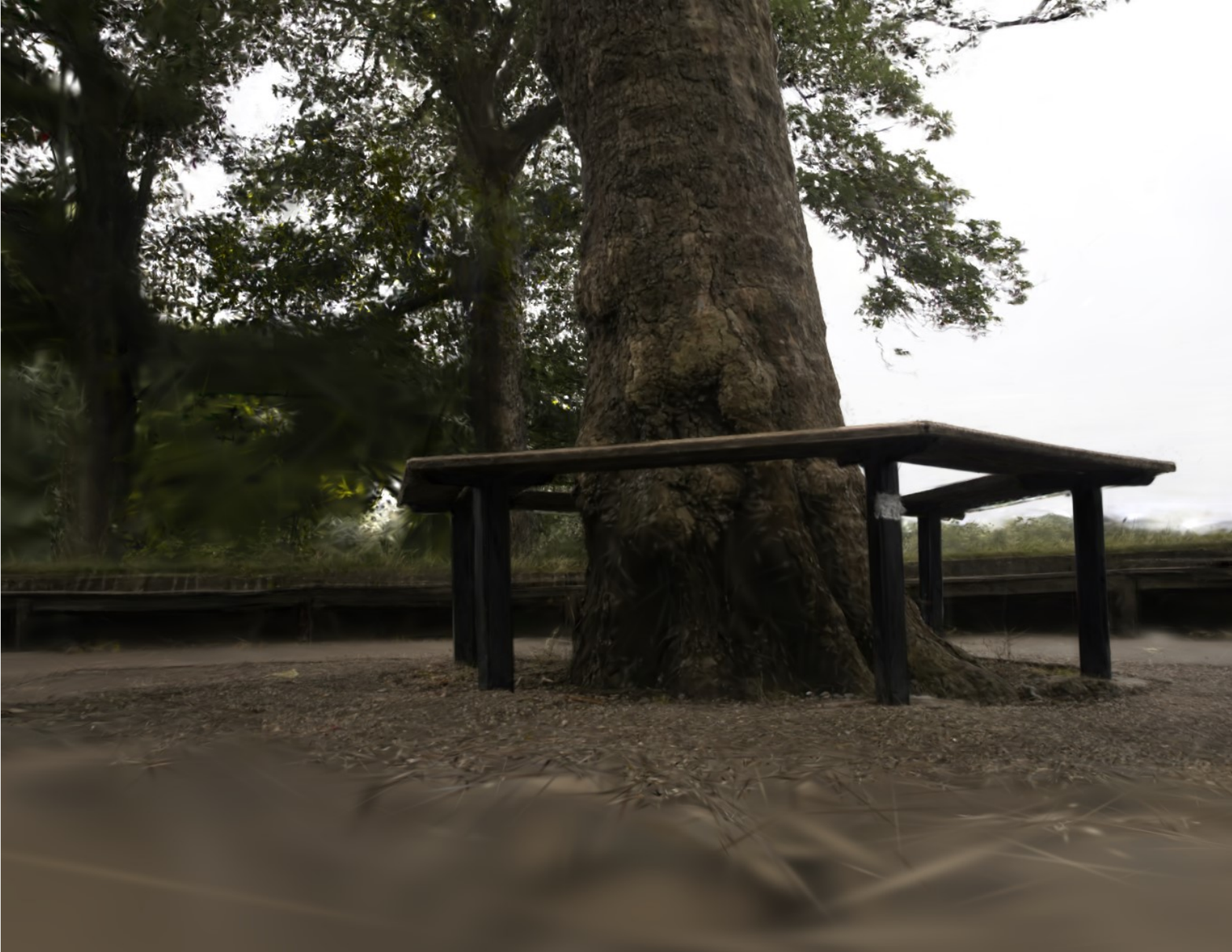}
    \end{subfigure}
    \begin{subfigure}[b]{0.23\textwidth}
        \includegraphics[width=\textwidth]{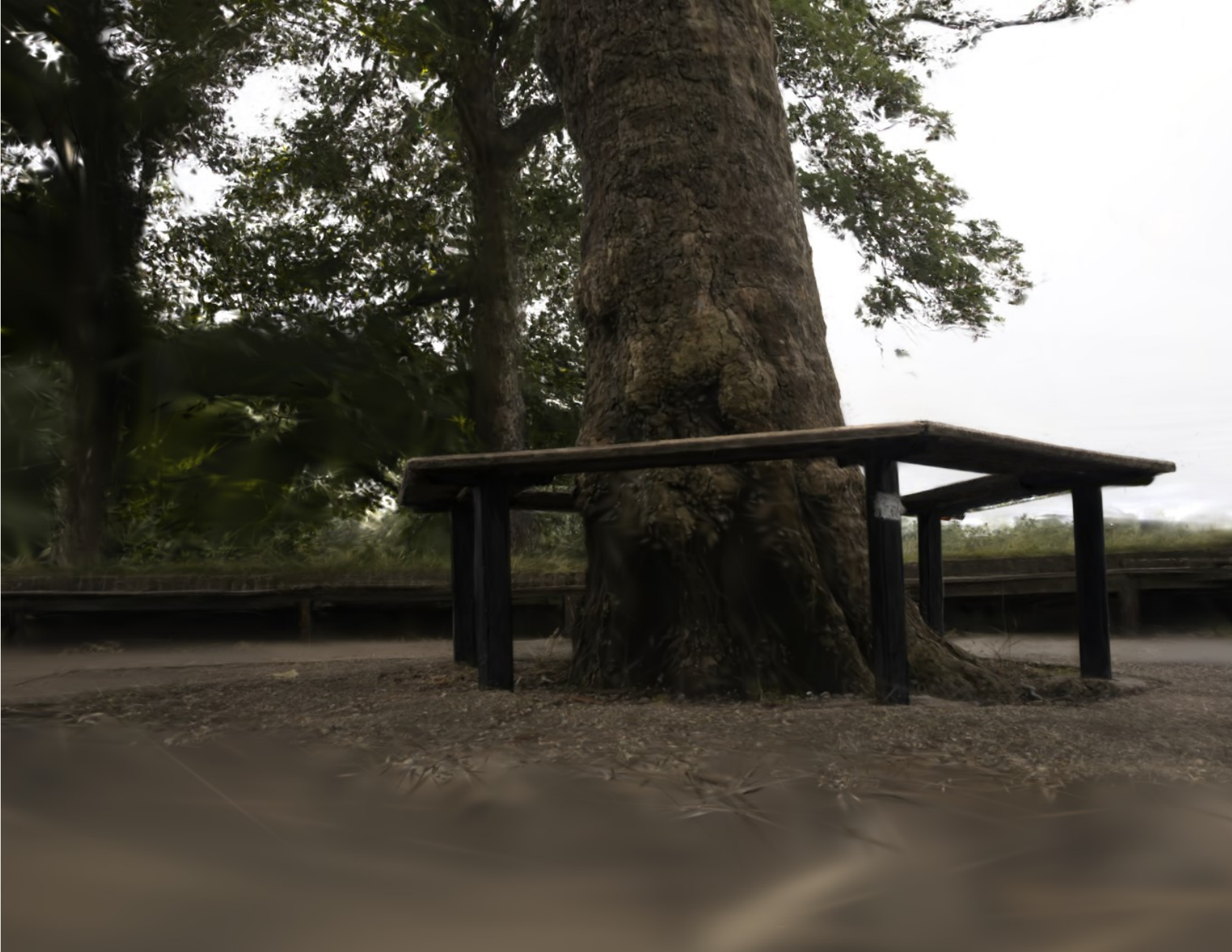}
    \end{subfigure}
\caption{Rendered images on MipNeRF360 Dataset}
\end{figure}


\begin{figure}[htbp]
    \centering

    \makebox[0.08\textwidth]{} 
    \makebox[0.29\textwidth]{\textbf{PSNR $\uparrow$}}
    \makebox[0.29\textwidth]{\textbf{SSIM $\uparrow$}}
    \makebox[0.29\textwidth]{\textbf{LPIPS $\downarrow$}}
    \par\medskip

    \begin{minipage}[b]{0.05\textwidth}
        \rotatebox[origin=t, y=2.8cm]{90}{\textbf{Bicycle}}
    \end{minipage}
    \begin{minipage}[b]{0.90\textwidth}
    \rotatebox[origin=t, y=1.3cm]{90}{\textbf{RadSplat}}
    \begin{subfigure}[b]{0.31\textwidth}
        \includegraphics[width=\textwidth]{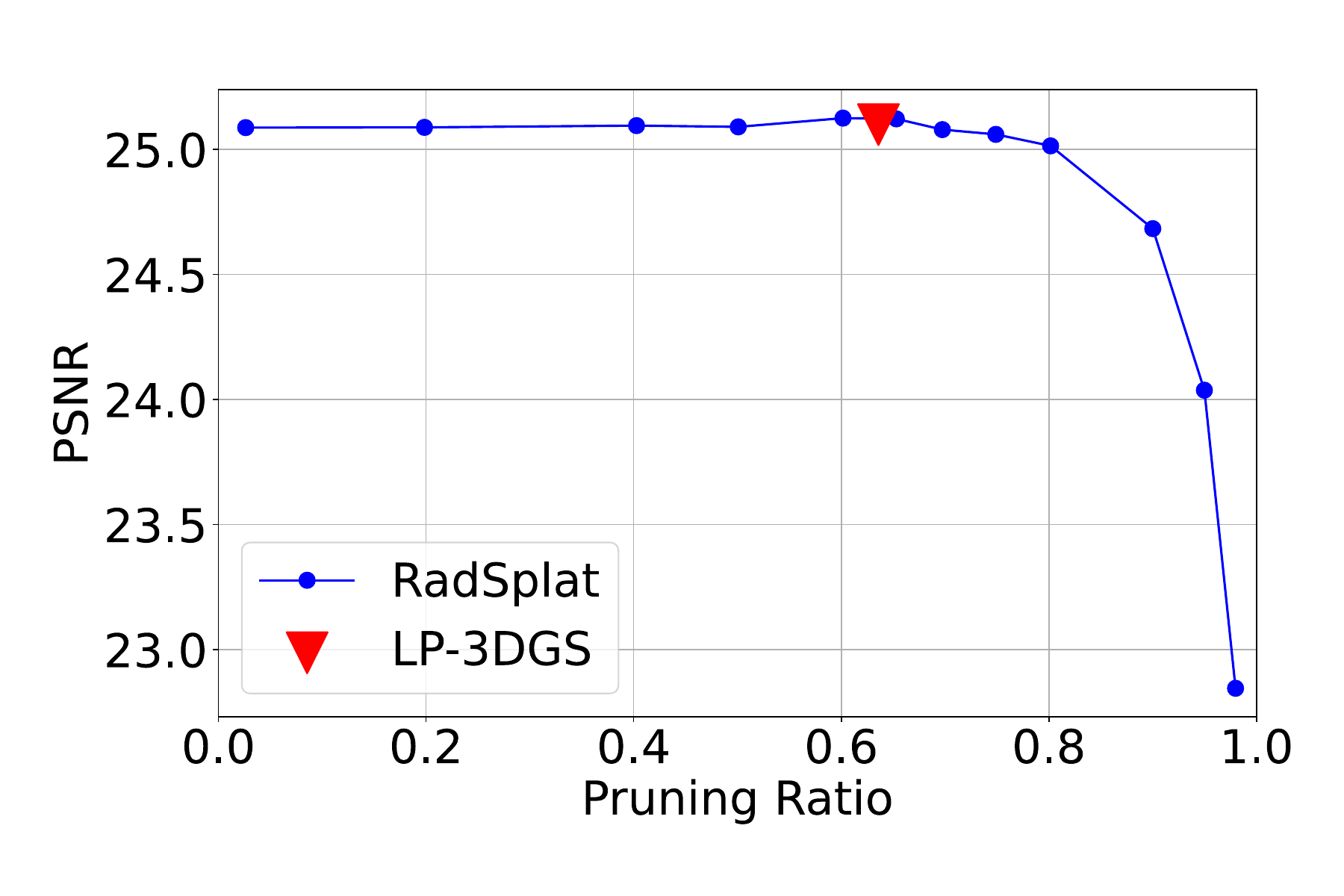}
    \end{subfigure}
    \begin{subfigure}[b]{0.31\textwidth}
        \includegraphics[width=\textwidth]{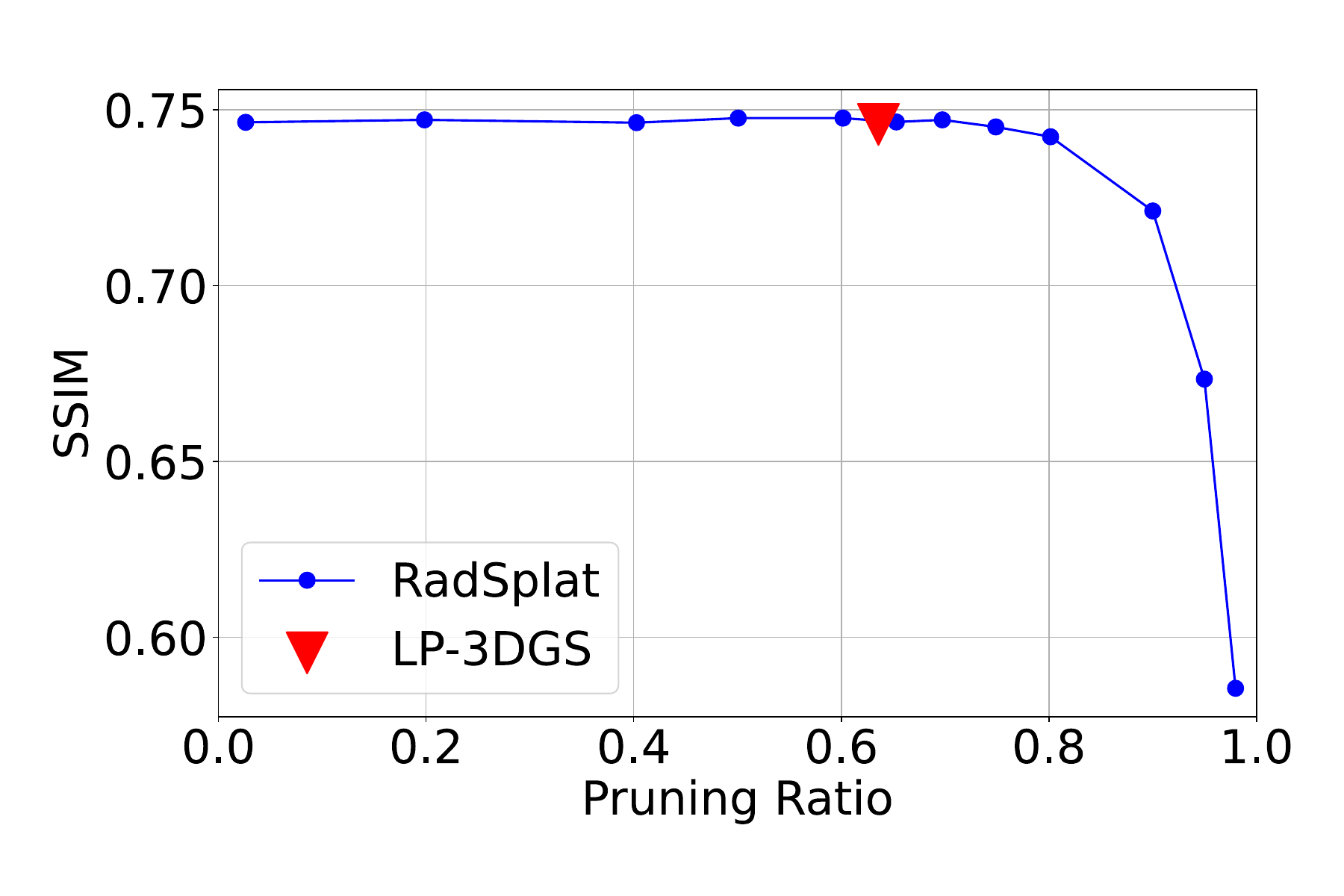}
    \end{subfigure}
    \begin{subfigure}[b]{0.31\textwidth}
        \includegraphics[width=\textwidth]{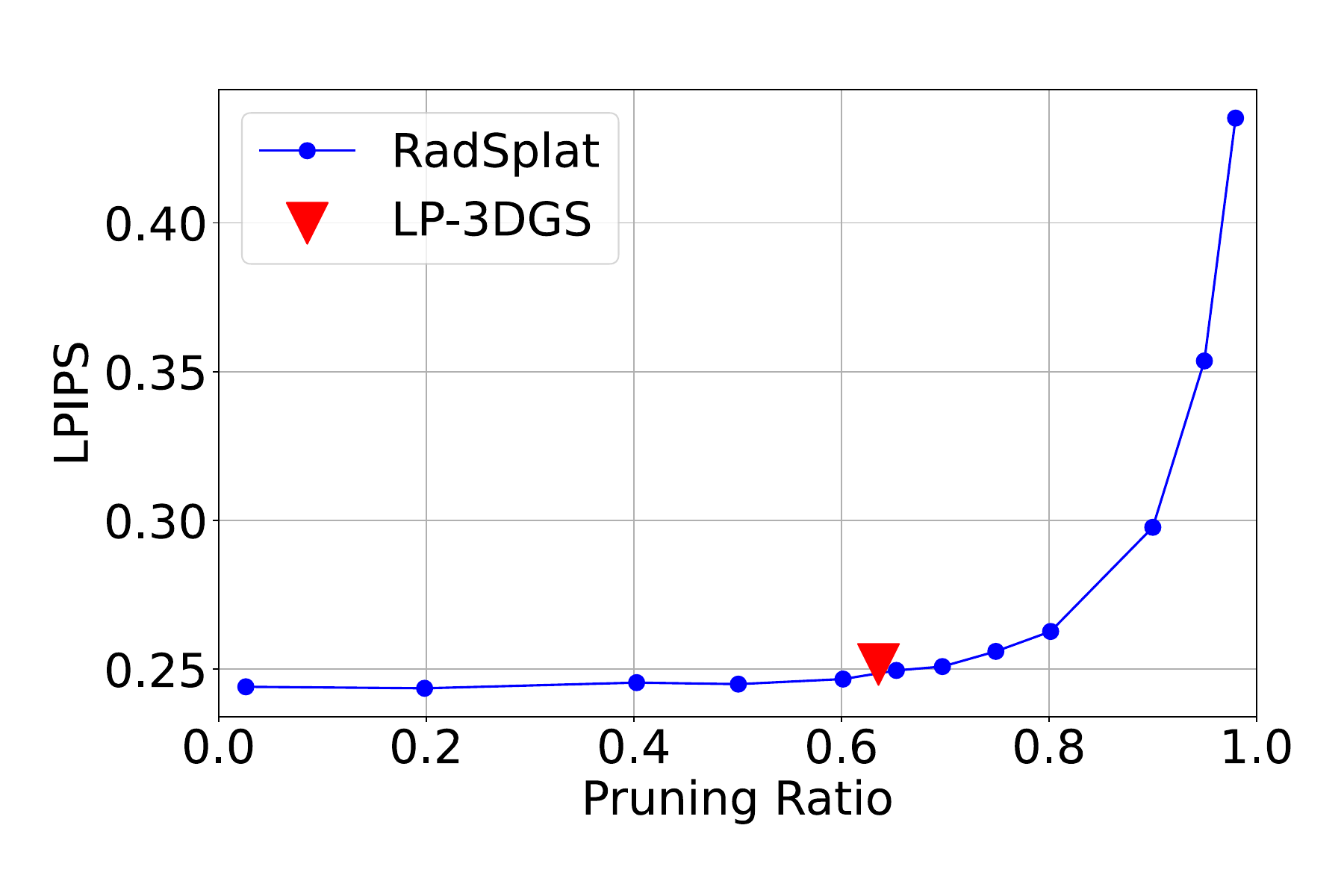}
    \end{subfigure}

    \rotatebox[origin=t, y=1.4cm]{90}{\textbf{Mini-Splatting}}
    \begin{subfigure}[b]{0.31\textwidth}
        \includegraphics[width=\textwidth]{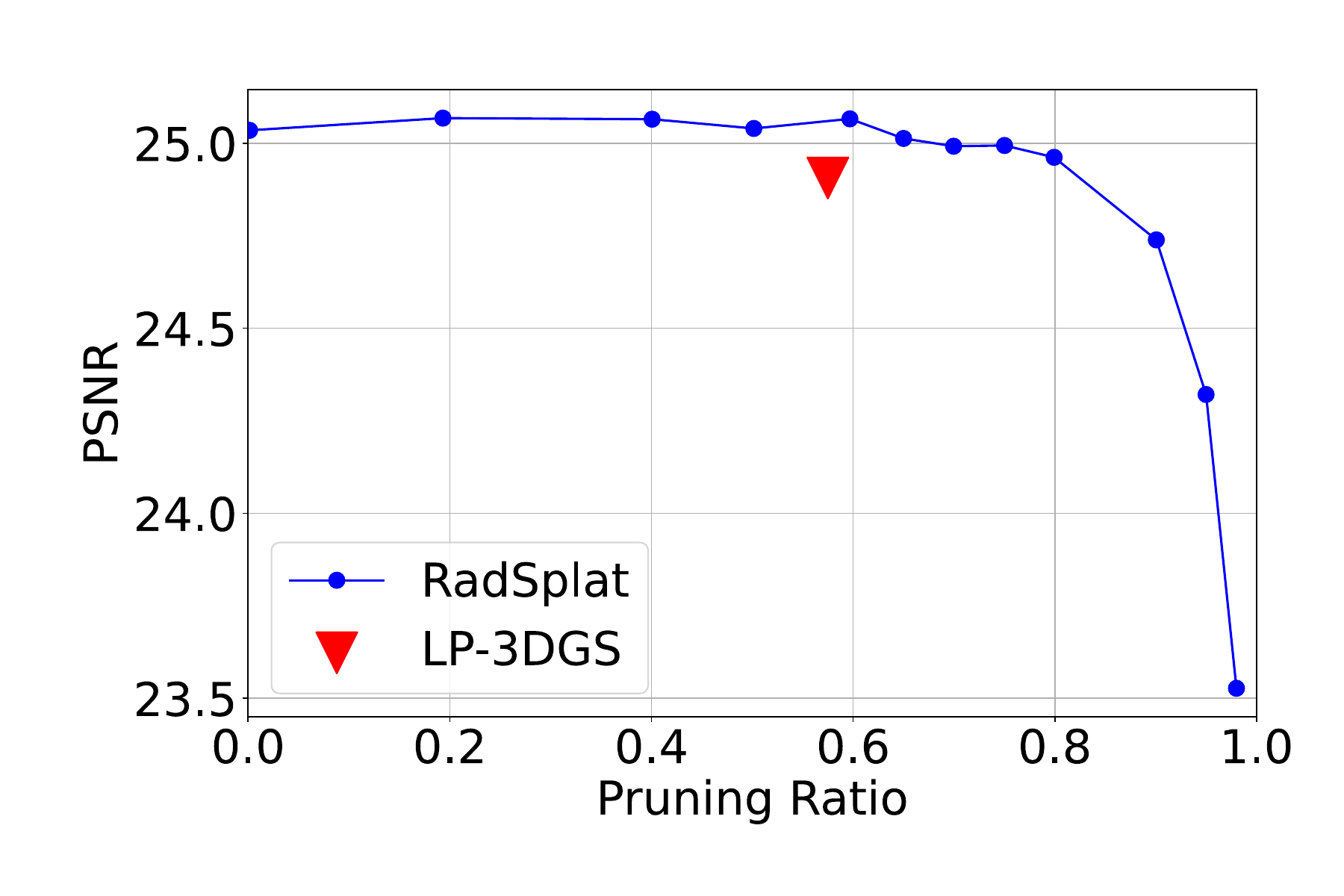}
    \end{subfigure}
    \begin{subfigure}[b]{0.31\textwidth}
        \includegraphics[width=\textwidth]{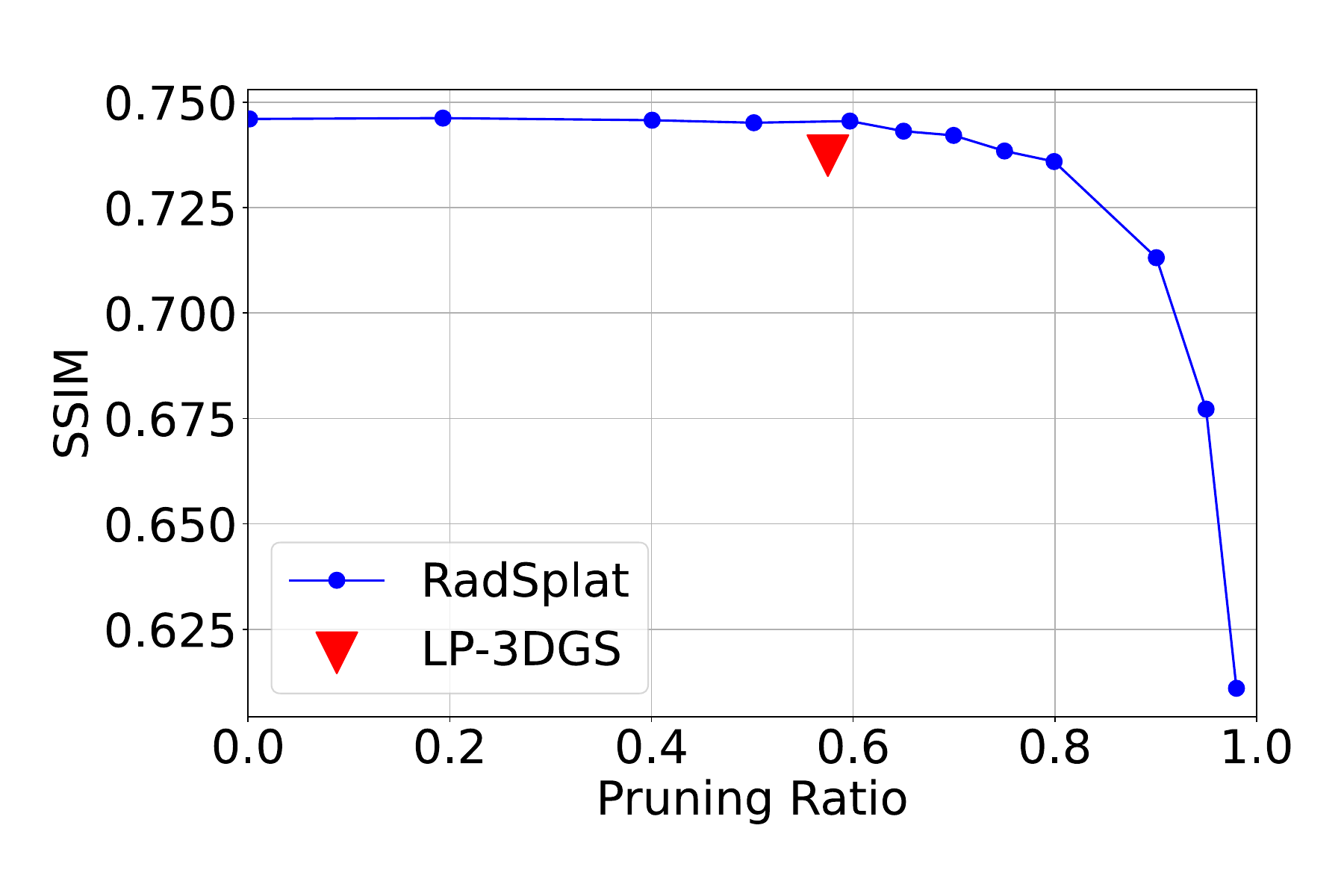}
    \end{subfigure}
    \begin{subfigure}[b]{0.31\textwidth}
        \includegraphics[width=\textwidth]{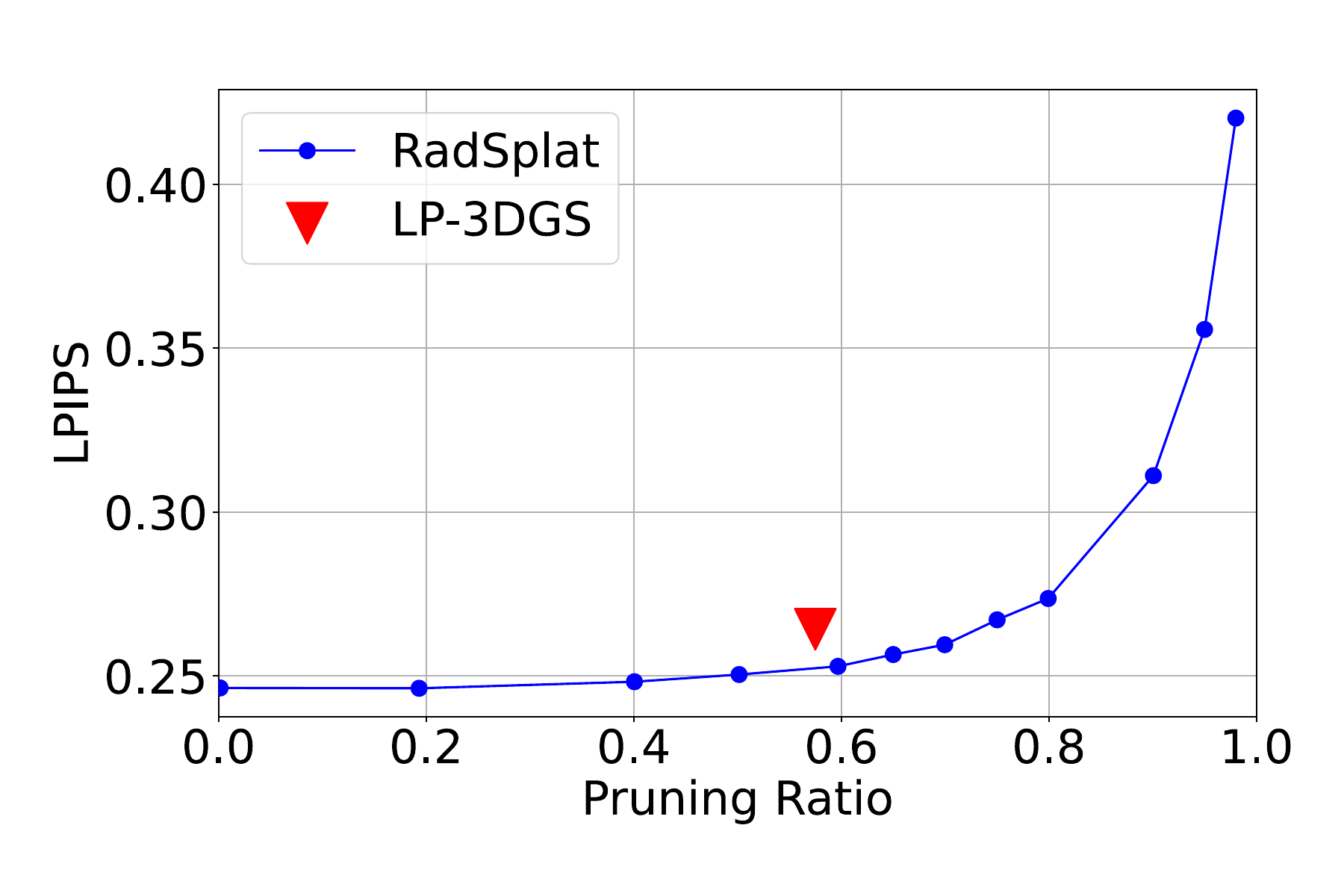}
    \end{subfigure}
    \end{minipage}

    \begin{minipage}[b]{0.05\textwidth}
        \rotatebox[origin=t, y=2.8cm]{90}{\textbf{Bonsai}}
    \end{minipage}
    \begin{minipage}[b]{0.90\textwidth}
    \rotatebox[origin=t, y=1.3cm]{90}{\textbf{RadSplat}}
    \begin{subfigure}[b]{0.31\textwidth}
        \includegraphics[width=\textwidth]{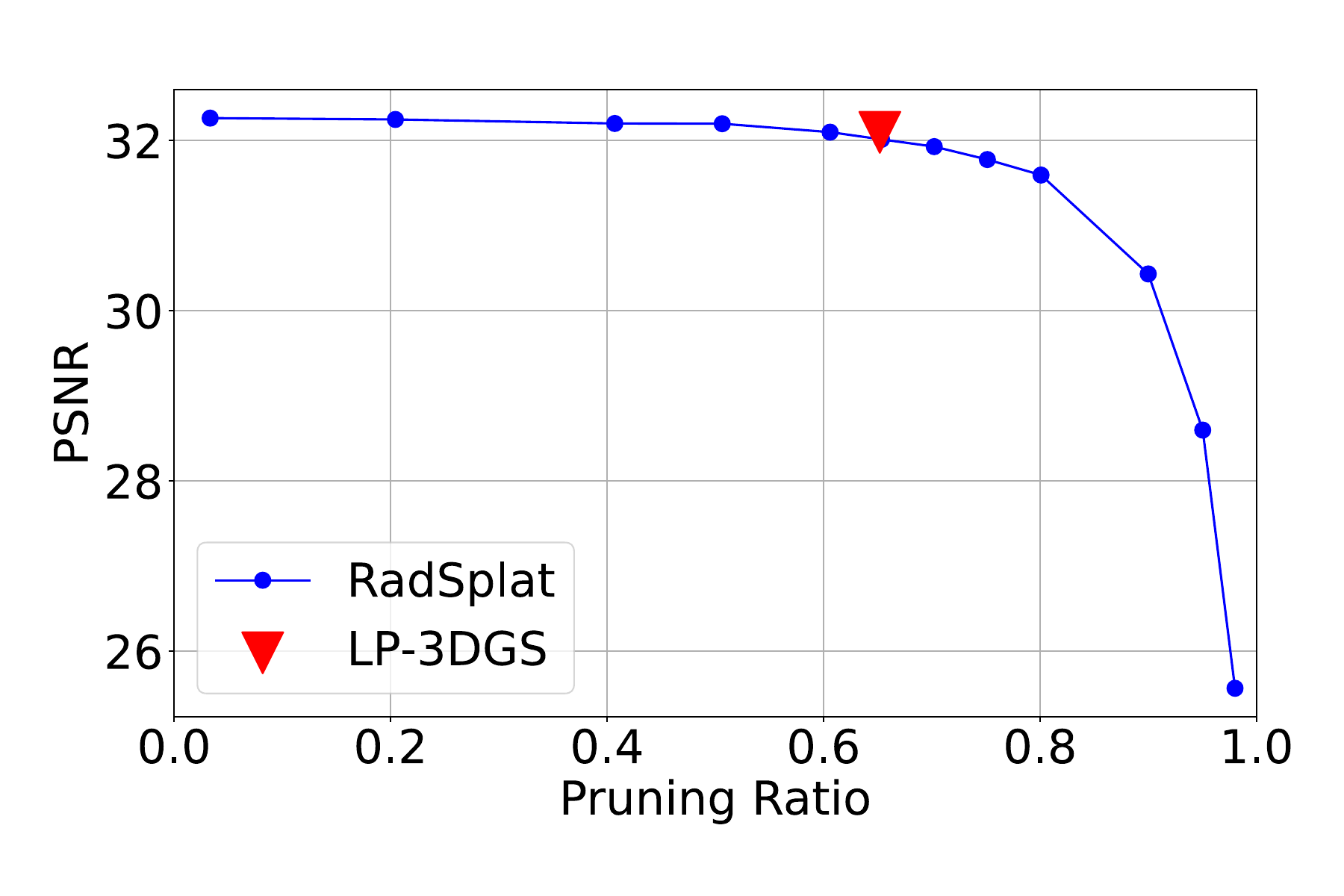}
    \end{subfigure}
    \begin{subfigure}[b]{0.31\textwidth}
        \includegraphics[width=\textwidth]{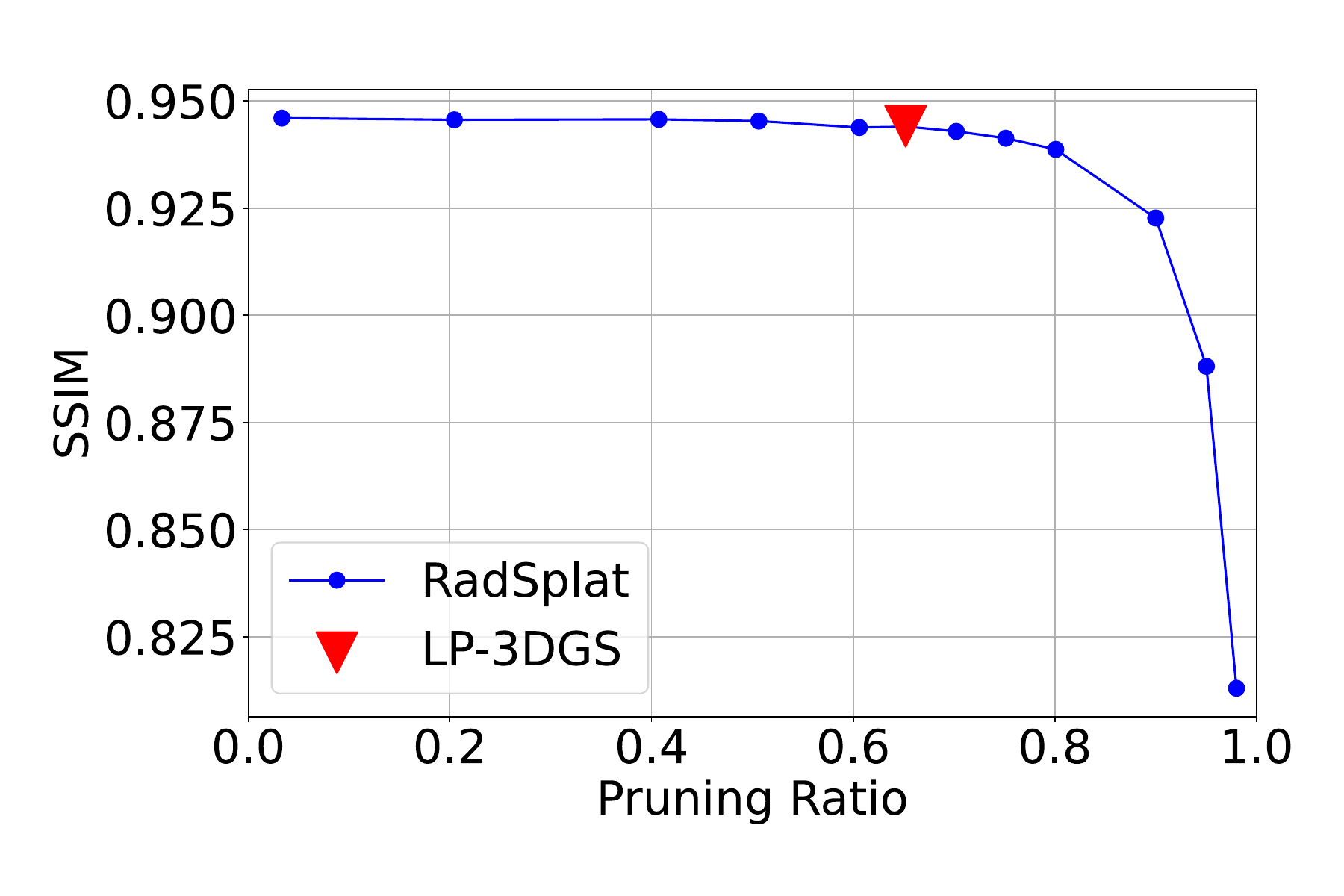}
    \end{subfigure}
    \begin{subfigure}[b]{0.31\textwidth}
        \includegraphics[width=\textwidth]{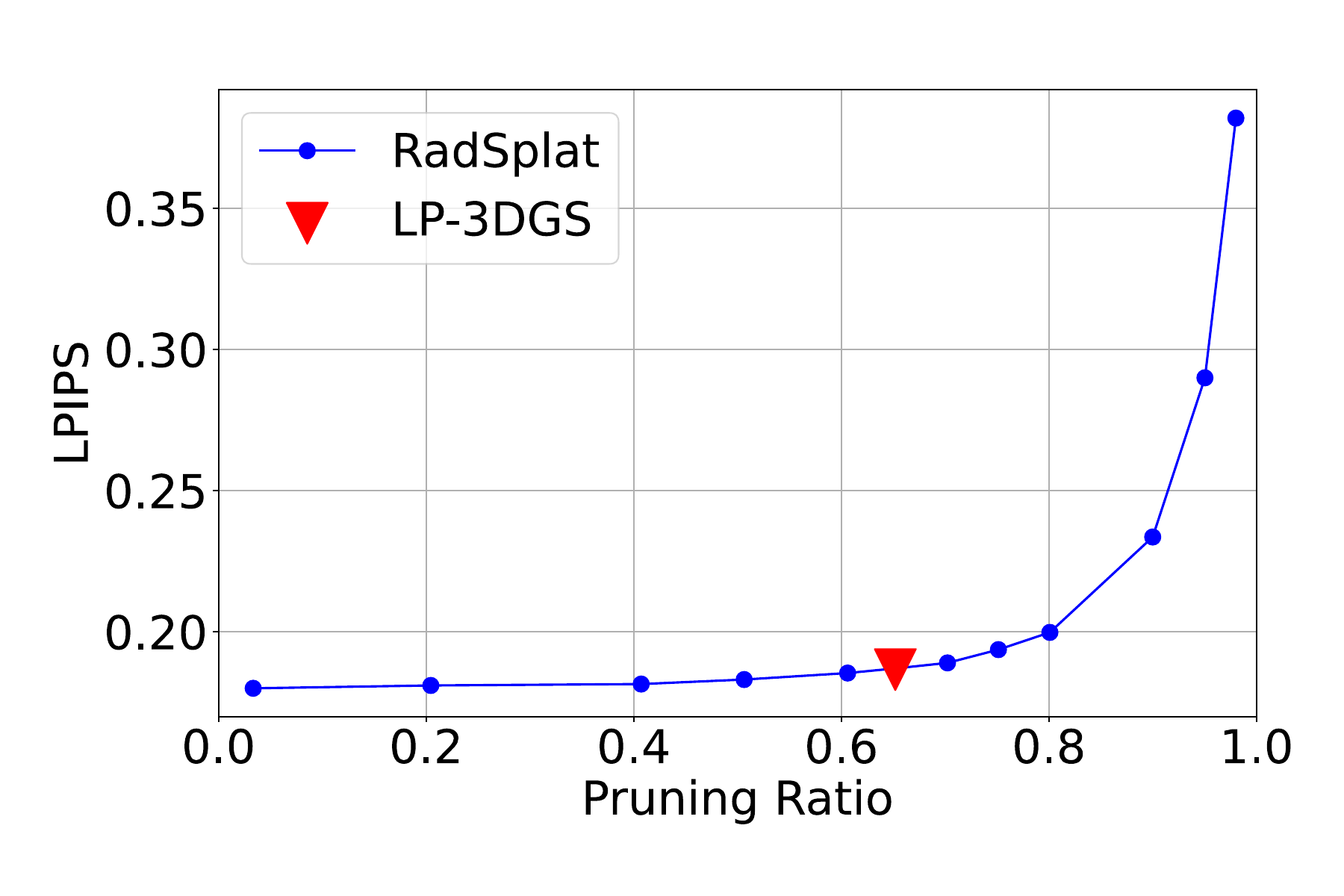}
    \end{subfigure}

    \rotatebox[origin=t, y=1.4cm]{90}{\textbf{Mini-Splatting}}
    \begin{subfigure}[b]{0.31\textwidth}
        \includegraphics[width=\textwidth]{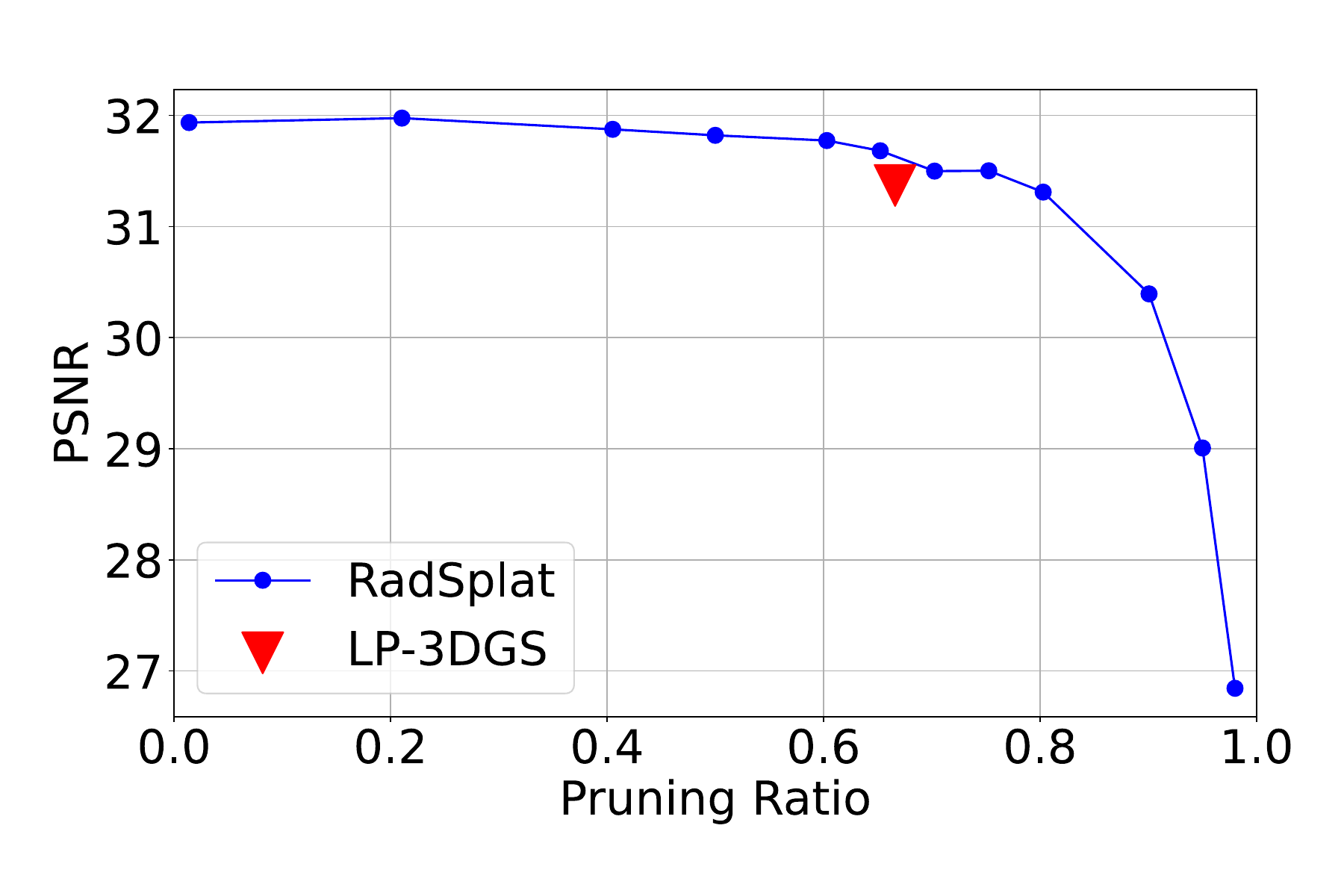}
    \end{subfigure}
    \begin{subfigure}[b]{0.31\textwidth}
        \includegraphics[width=\textwidth]{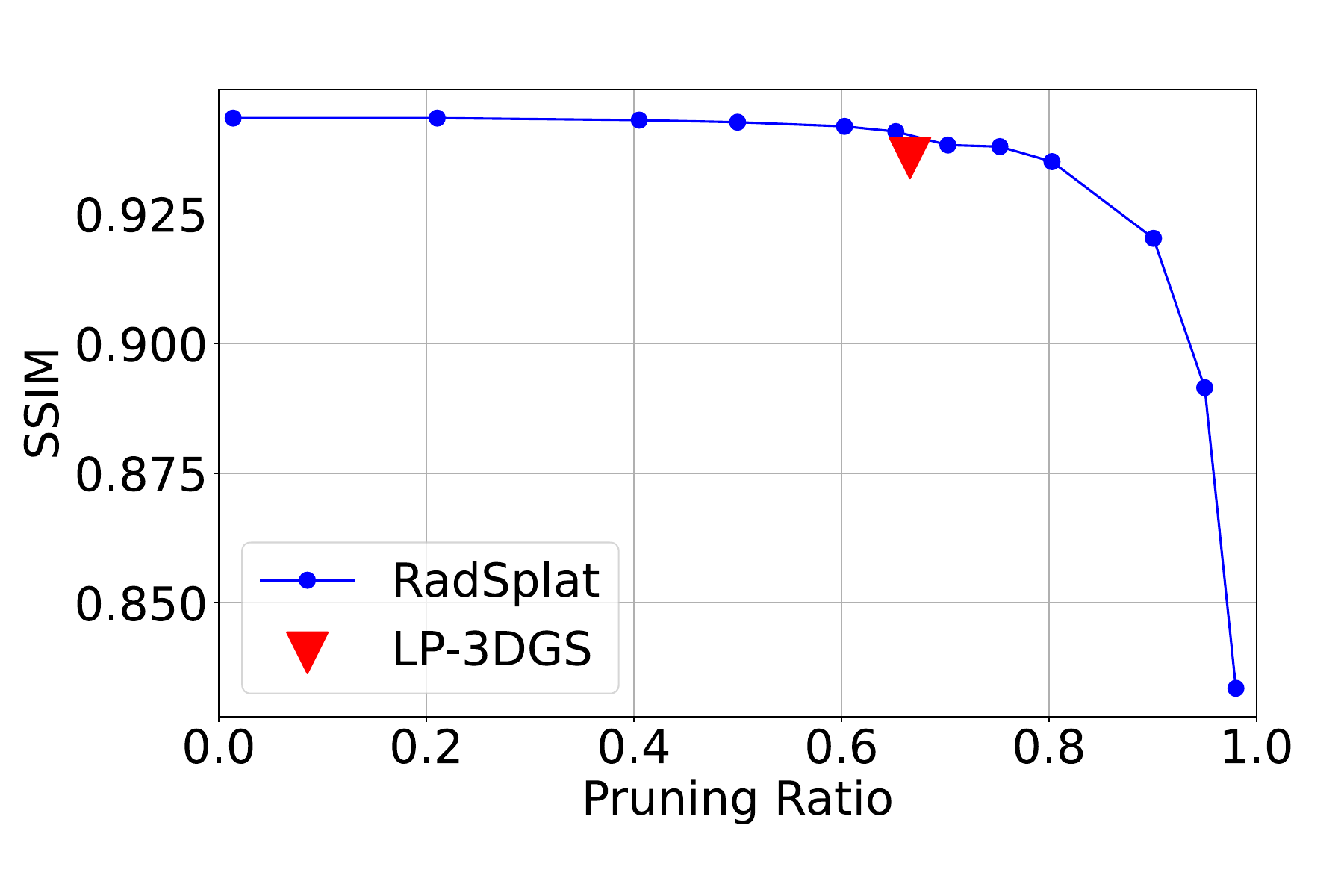}
    \end{subfigure}
    \begin{subfigure}[b]{0.31\textwidth}
        \includegraphics[width=\textwidth]{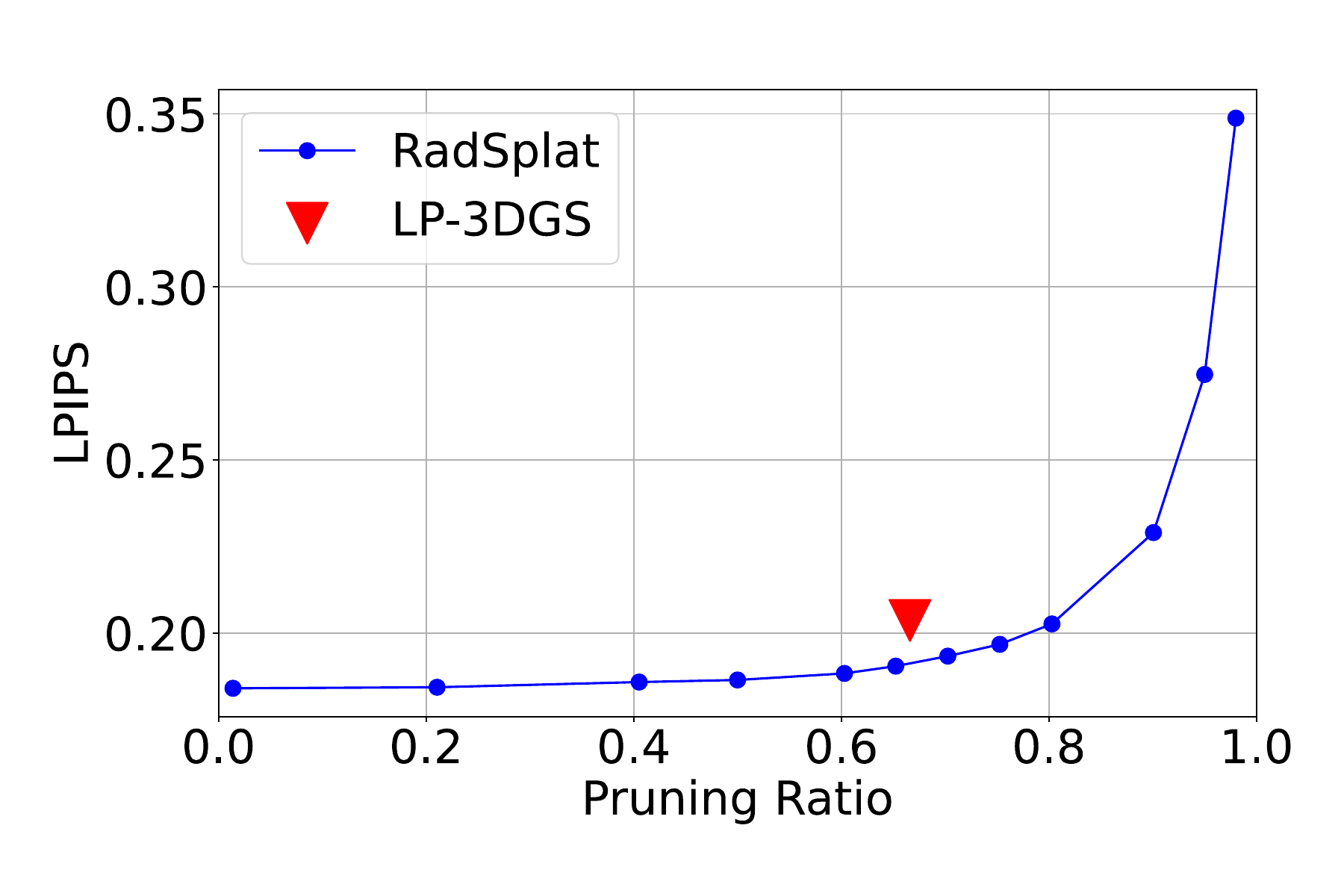}
    \end{subfigure}
    \end{minipage}
\end{figure}
\clearpage
\begin{figure}[htbp]\ContinuedFloat
    \centering

    \begin{minipage}[b]{0.05\textwidth}
        \rotatebox[origin=t, y=2.8cm]{90}{\textbf{Counter}}
    \end{minipage}
    \begin{minipage}[b]{0.90\textwidth}
    \rotatebox[origin=t, y=1.3cm]{90}{\textbf{RadSplat}}
    \begin{subfigure}[b]{0.31\textwidth}
        \includegraphics[width=\textwidth]{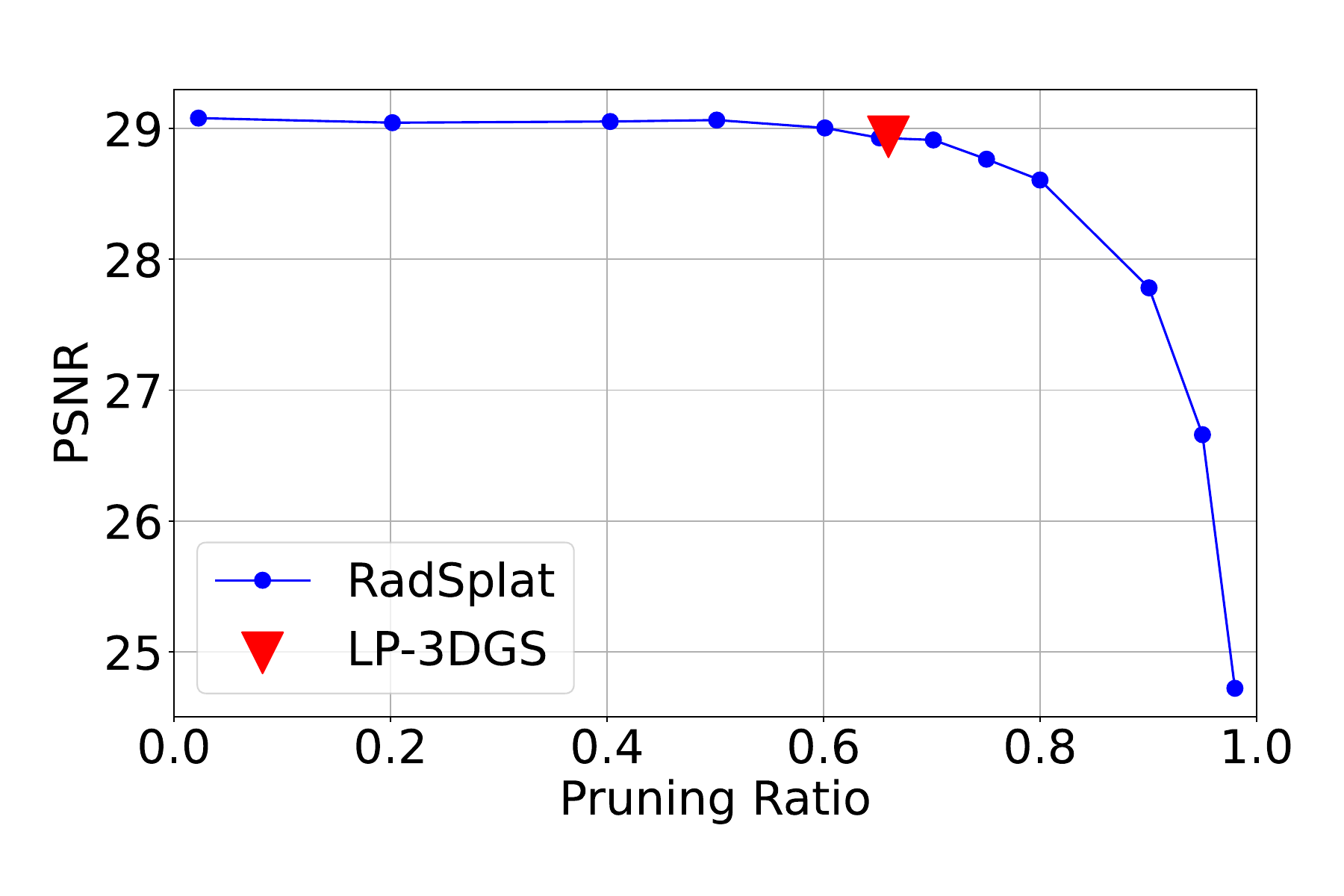}
    \end{subfigure}
    \begin{subfigure}[b]{0.31\textwidth}
        \includegraphics[width=\textwidth]{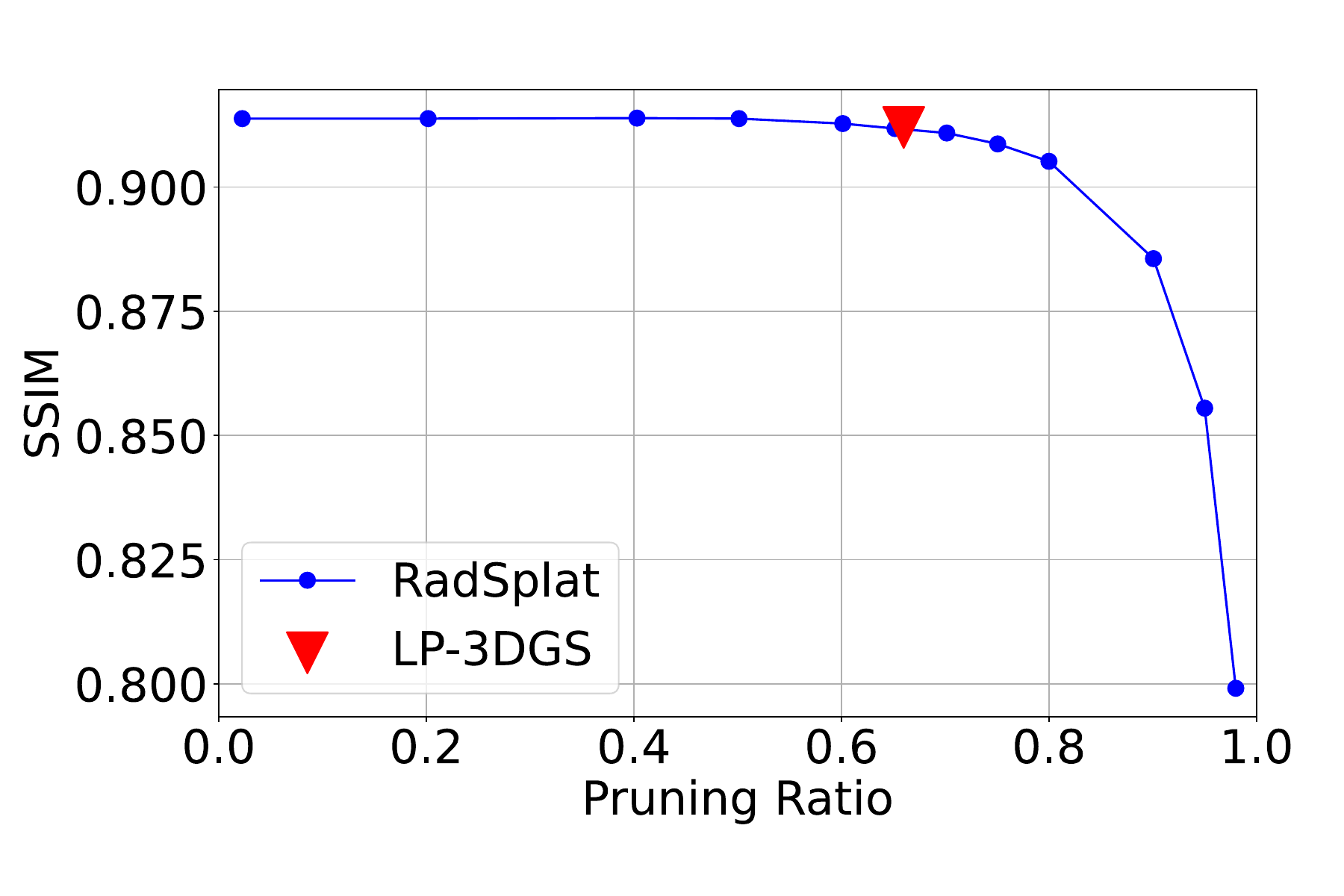}
    \end{subfigure}
    \begin{subfigure}[b]{0.31\textwidth}
        \includegraphics[width=\textwidth]{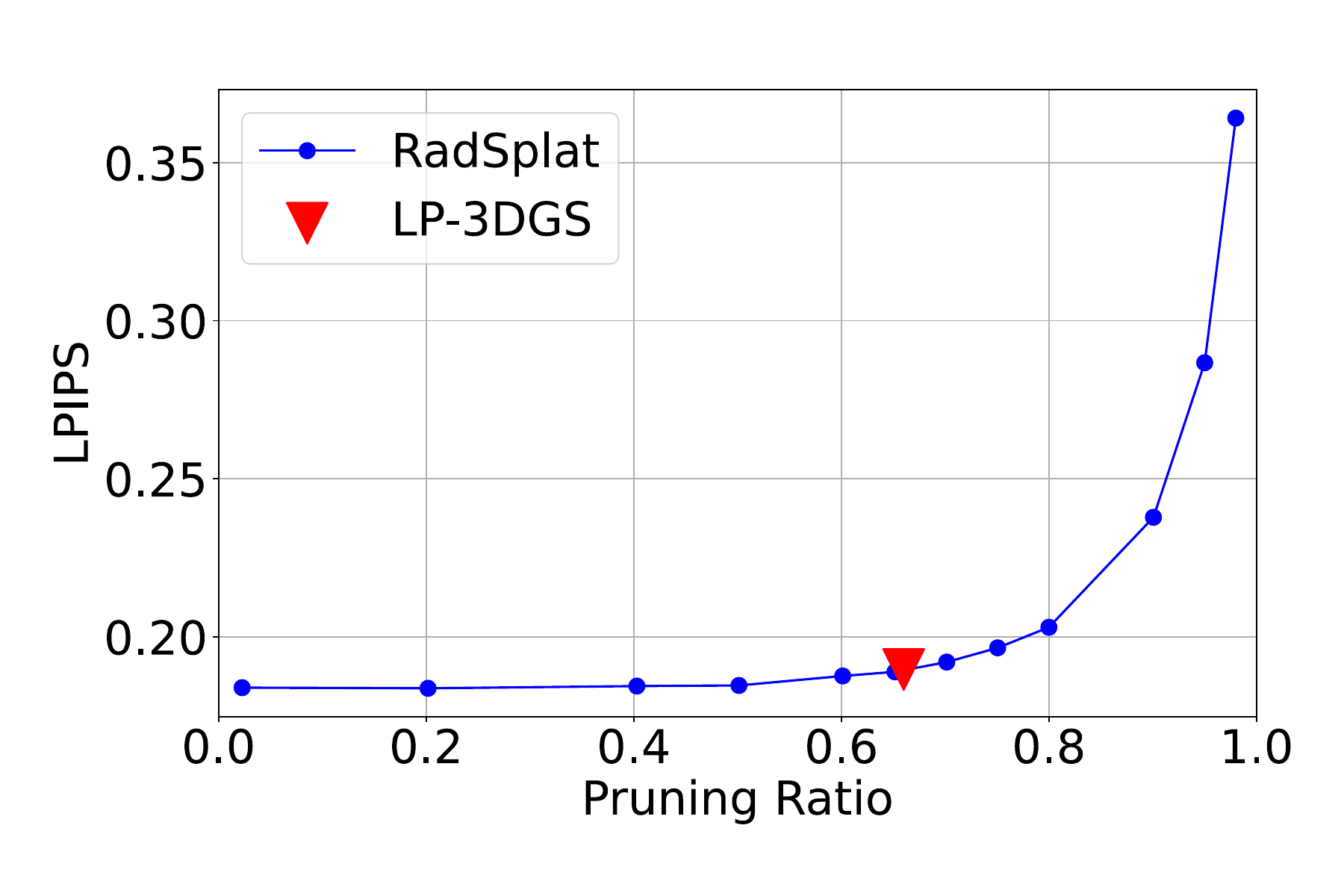}
    \end{subfigure}

    \rotatebox[origin=t, y=1.4cm]{90}{\textbf{Mini-Splatting}}
    \begin{subfigure}[b]{0.31\textwidth}
        \includegraphics[width=\textwidth]{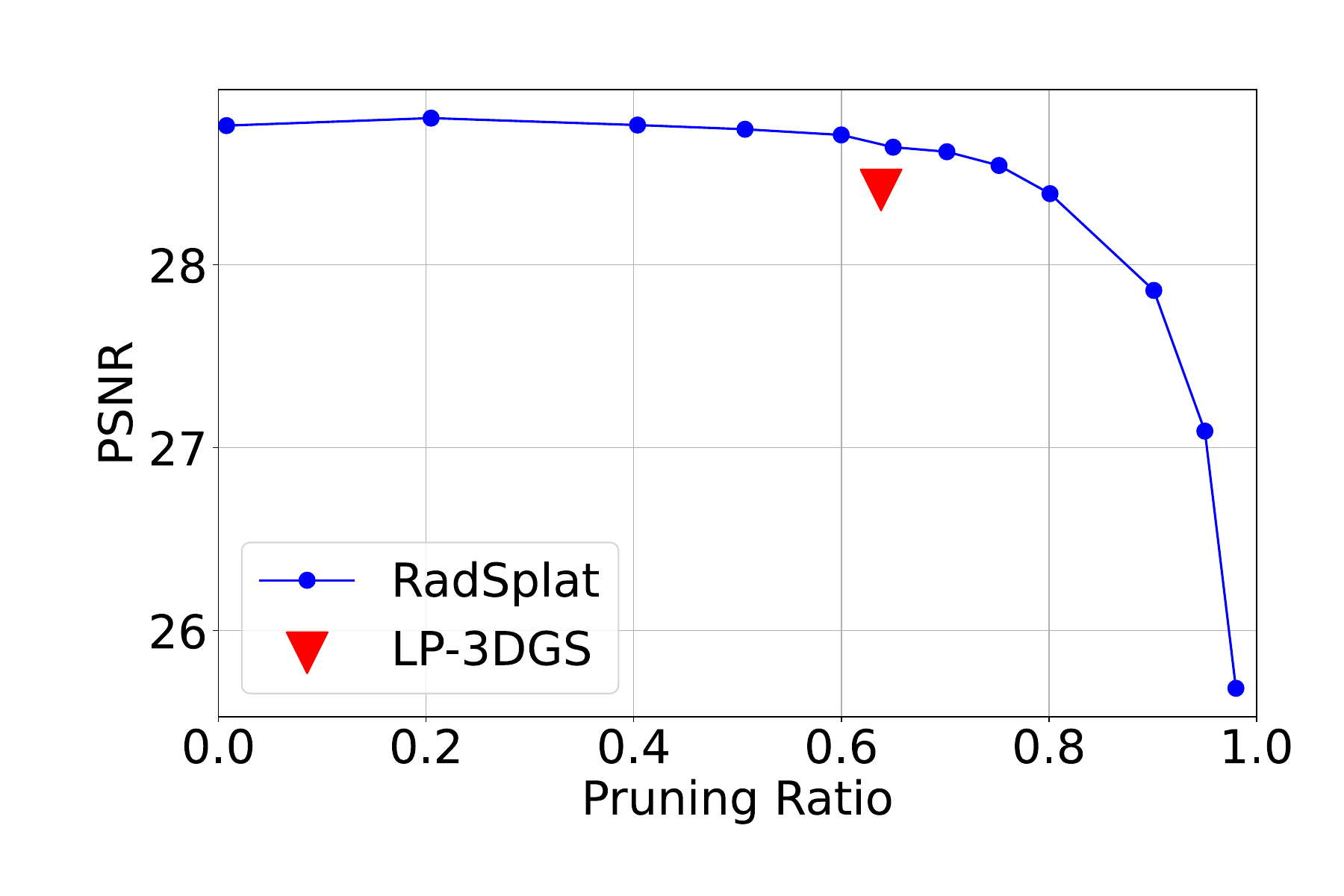}
    \end{subfigure}
    \begin{subfigure}[b]{0.31\textwidth}
        \includegraphics[width=\textwidth]{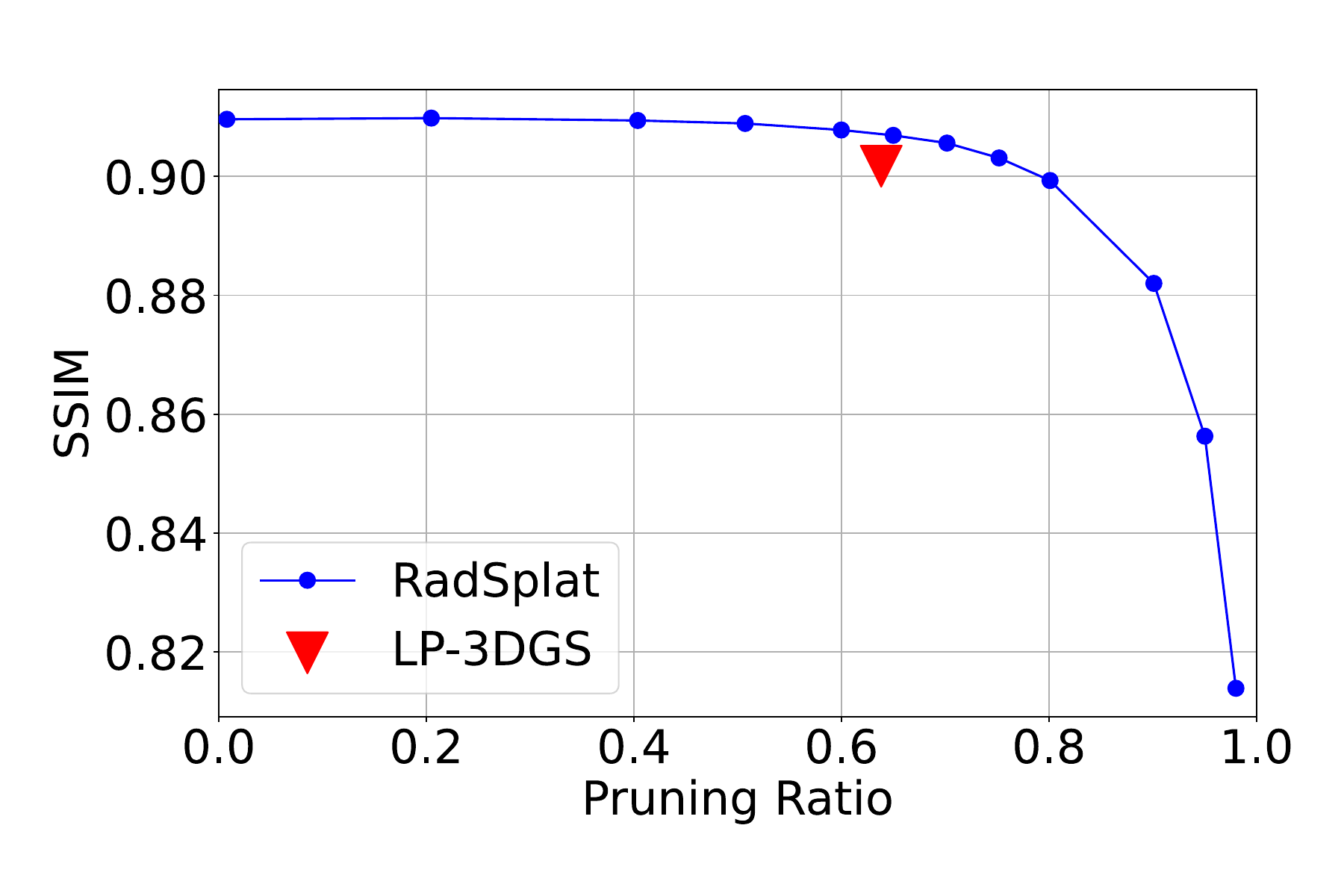}
    \end{subfigure}
    \begin{subfigure}[b]{0.31\textwidth}
        \includegraphics[width=\textwidth]{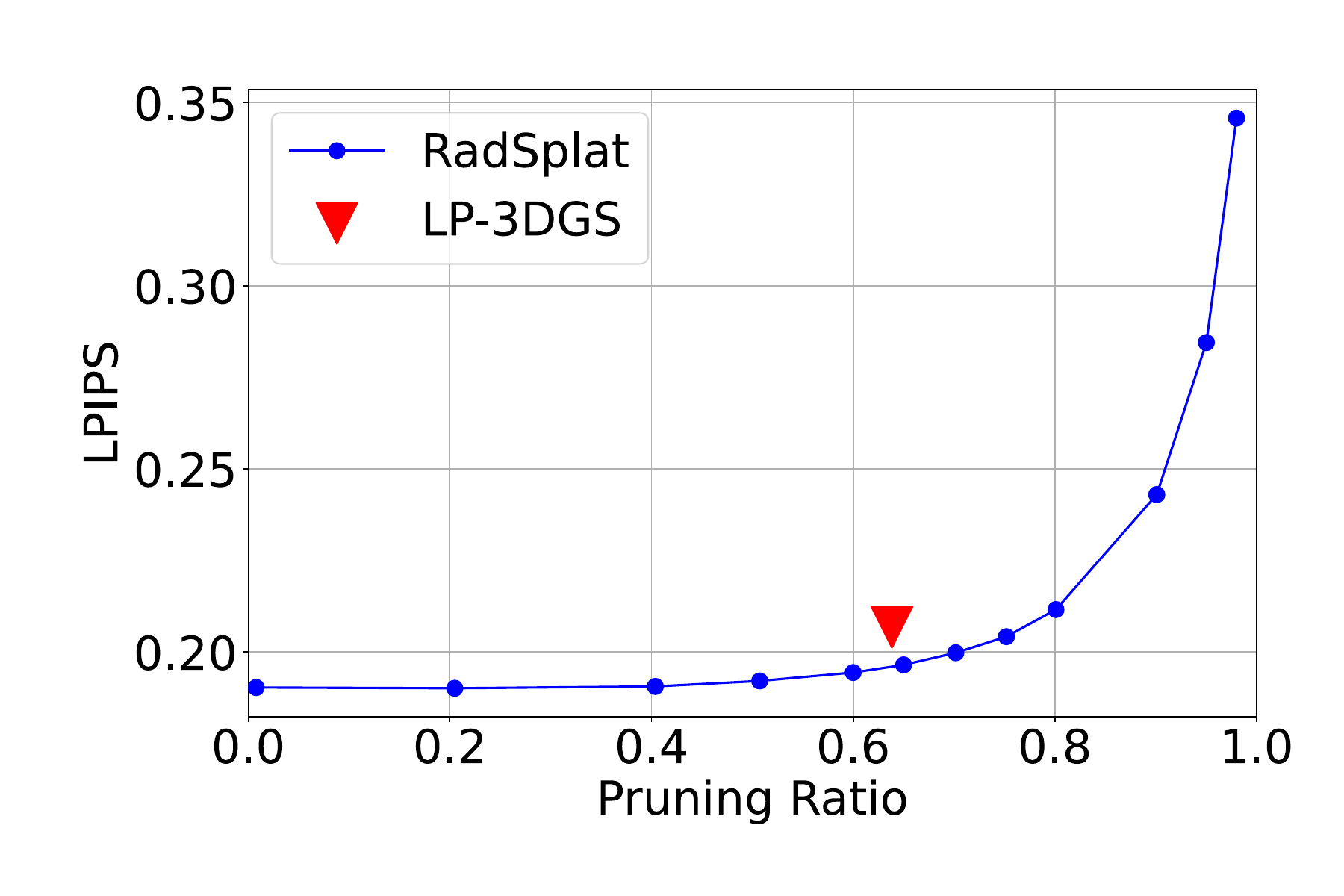}
    \end{subfigure}
    \end{minipage}

    \begin{minipage}[b]{0.05\textwidth}
        \rotatebox[origin=t, y=2.8cm]{90}{\textbf{Stump}}
    \end{minipage}
    \begin{minipage}[b]{0.90\textwidth}
    \rotatebox[origin=t, y=1.3cm]{90}{\textbf{RadSplat}}
    \begin{subfigure}[b]{0.31\textwidth}
        \includegraphics[width=\textwidth]{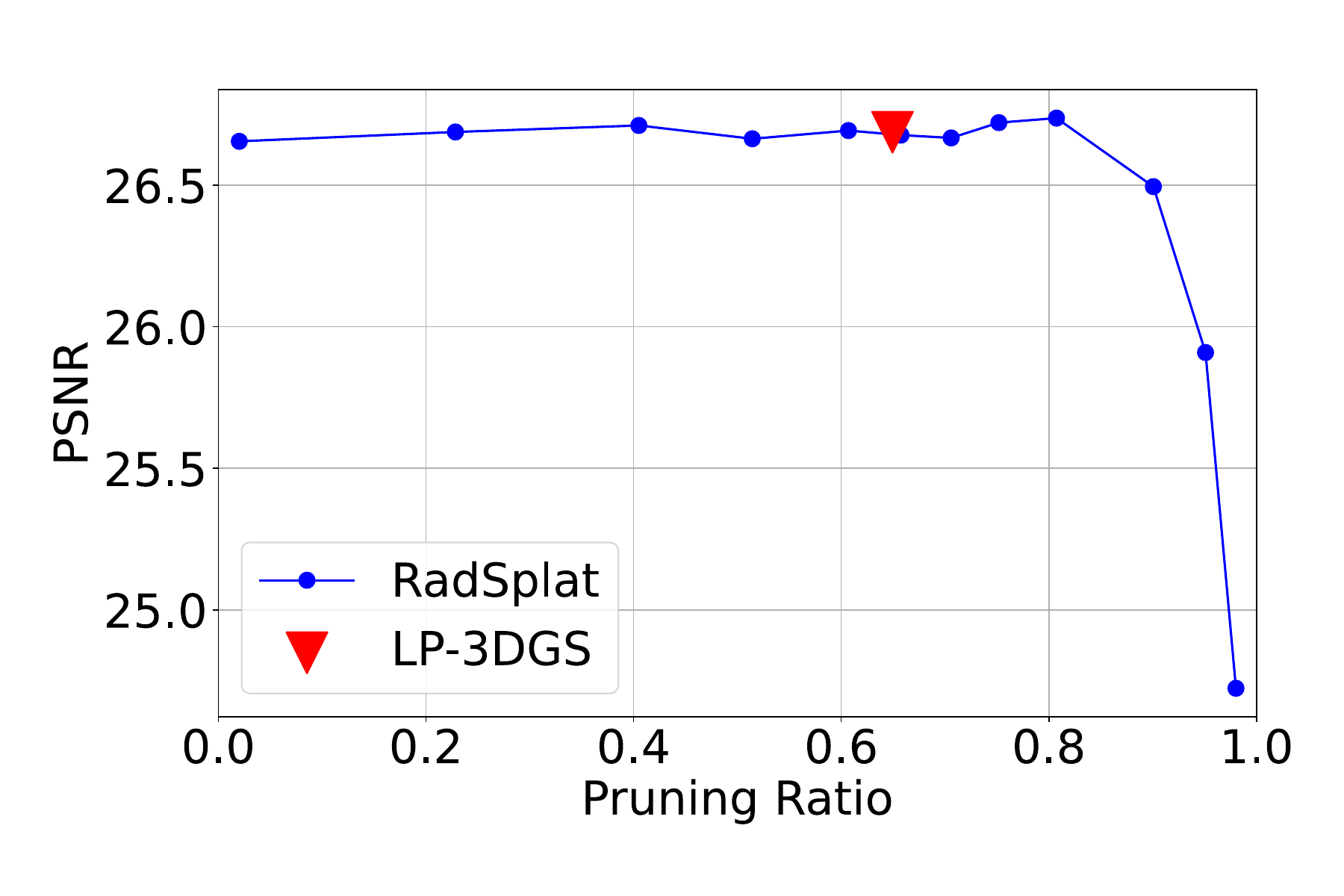}
    \end{subfigure}
    \begin{subfigure}[b]{0.31\textwidth}
        \includegraphics[width=\textwidth]{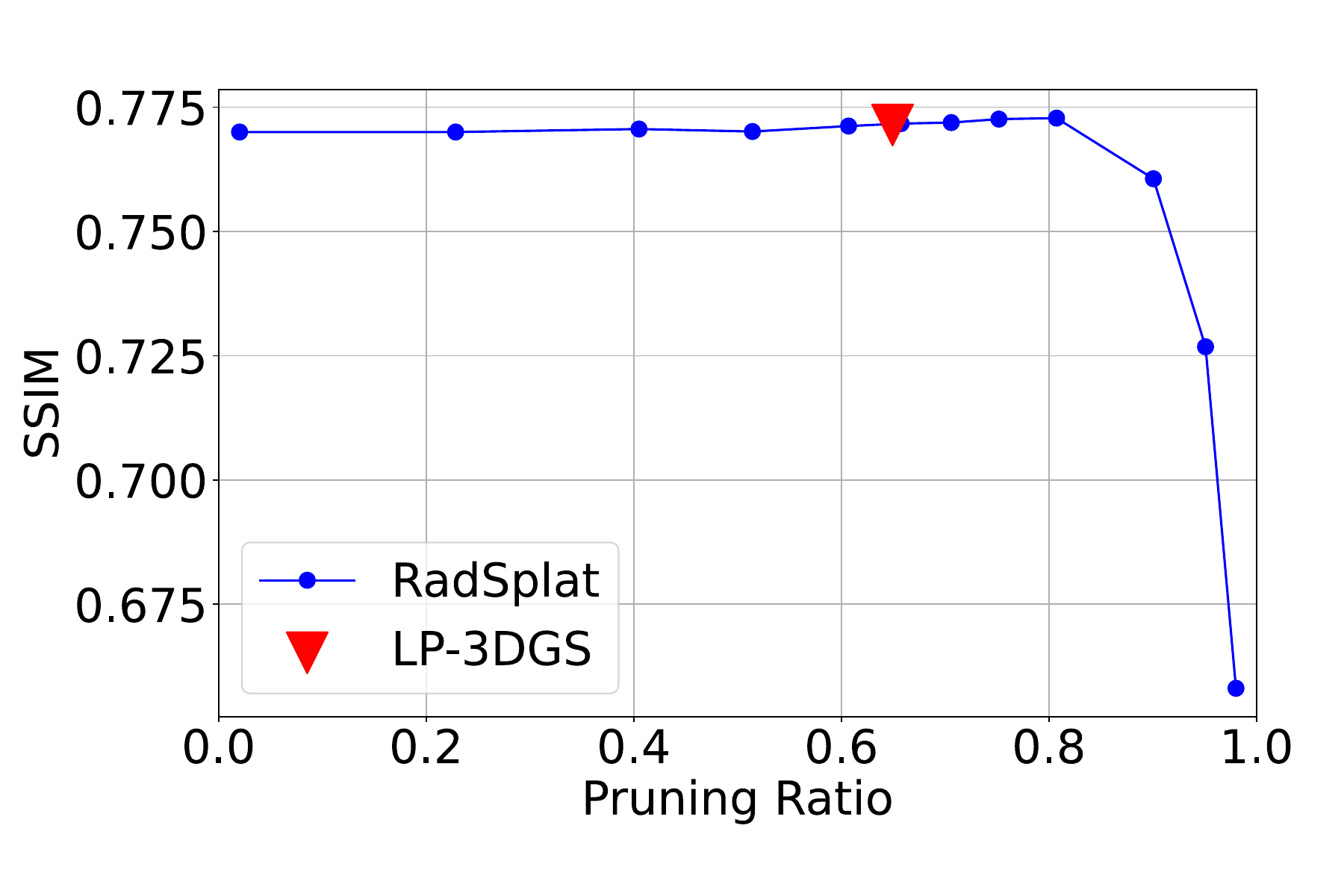}
    \end{subfigure}
    \begin{subfigure}[b]{0.31\textwidth}
        \includegraphics[width=\textwidth]{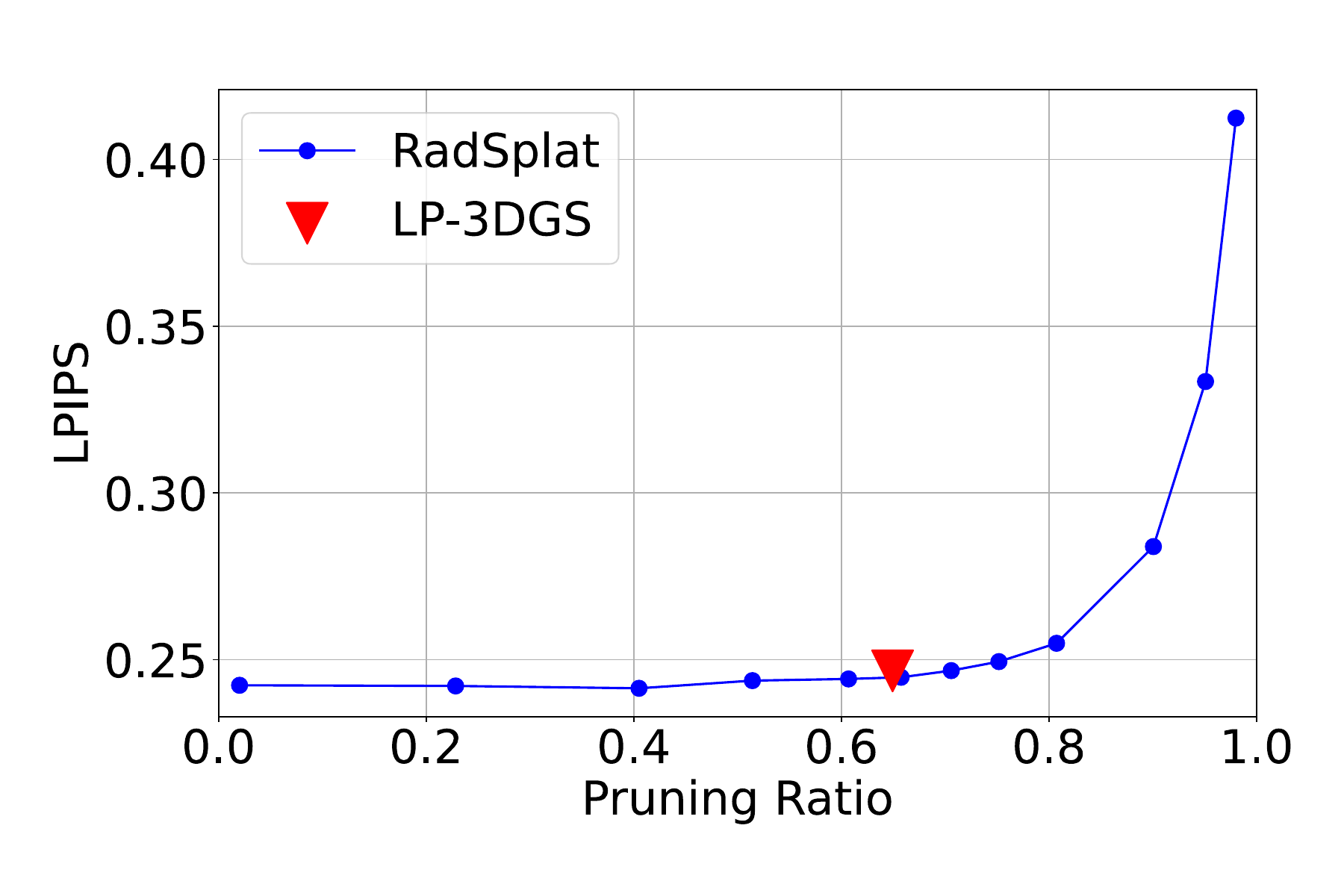}
    \end{subfigure}

    \rotatebox[origin=t, y=1.4cm]{90}{\textbf{Mini-Splatting}}
    \begin{subfigure}[b]{0.31\textwidth}
        \includegraphics[width=\textwidth]{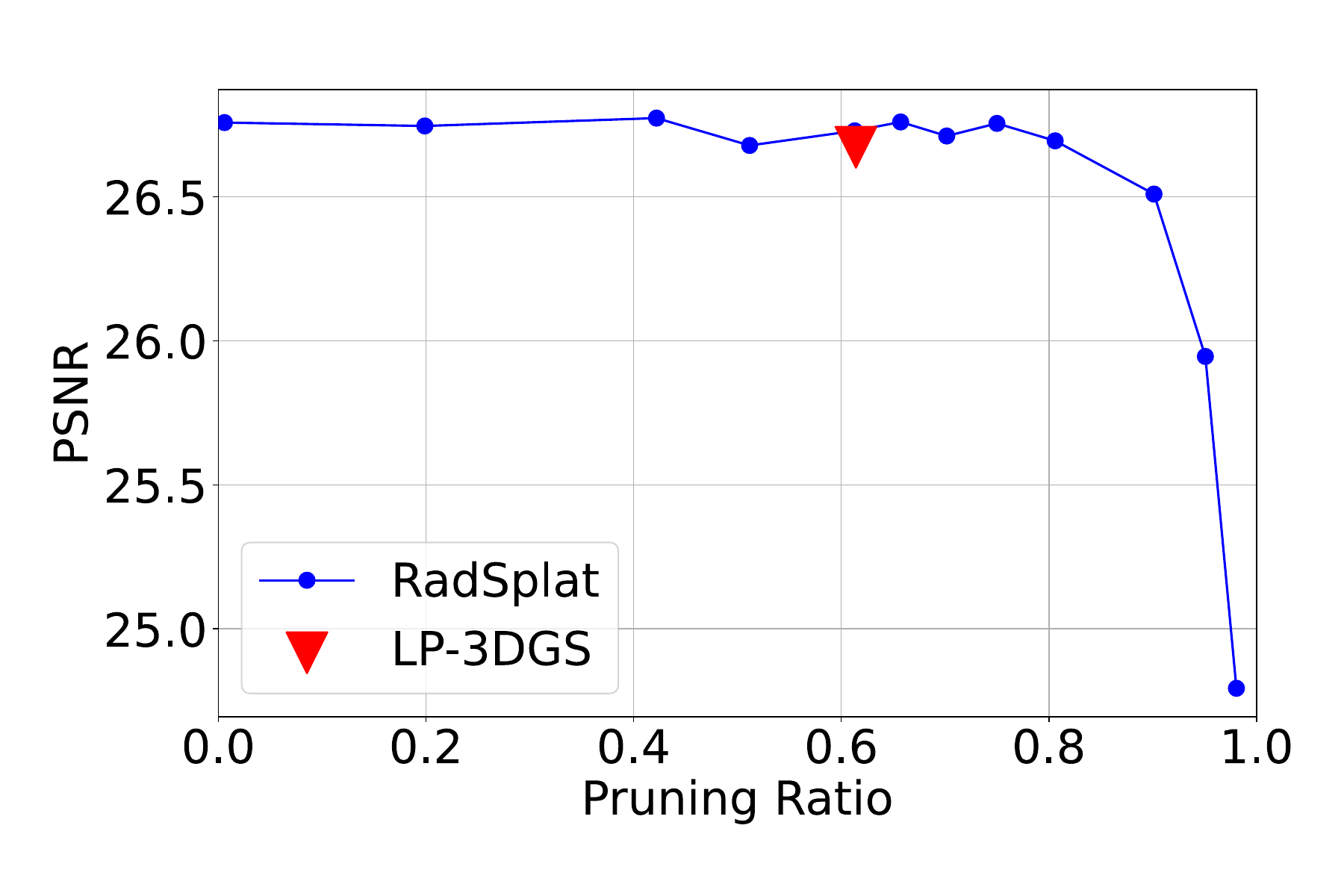}
    \end{subfigure}
    \begin{subfigure}[b]{0.31\textwidth}
        \includegraphics[width=\textwidth]{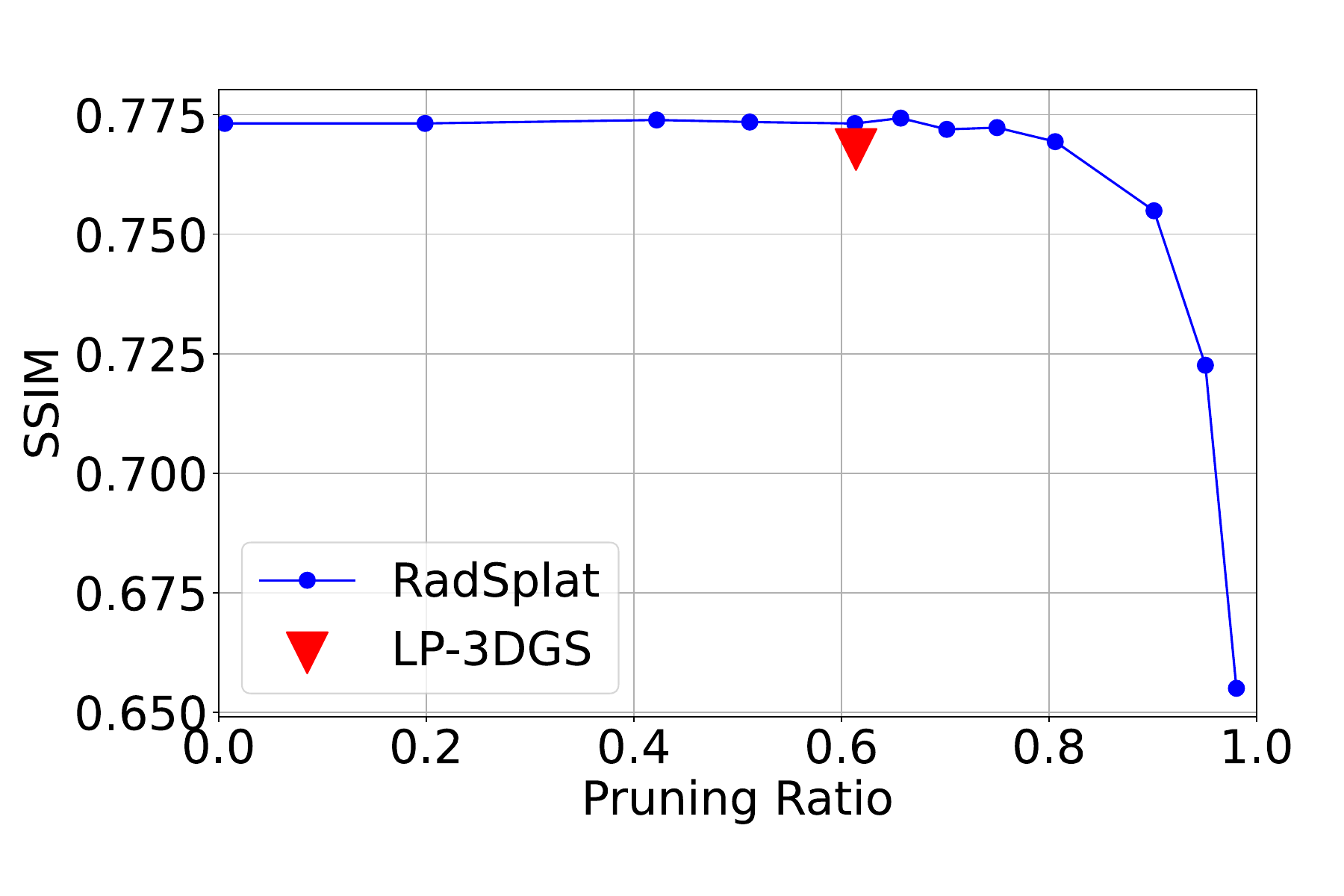}
    \end{subfigure}
    \begin{subfigure}[b]{0.31\textwidth}
        \includegraphics[width=\textwidth]{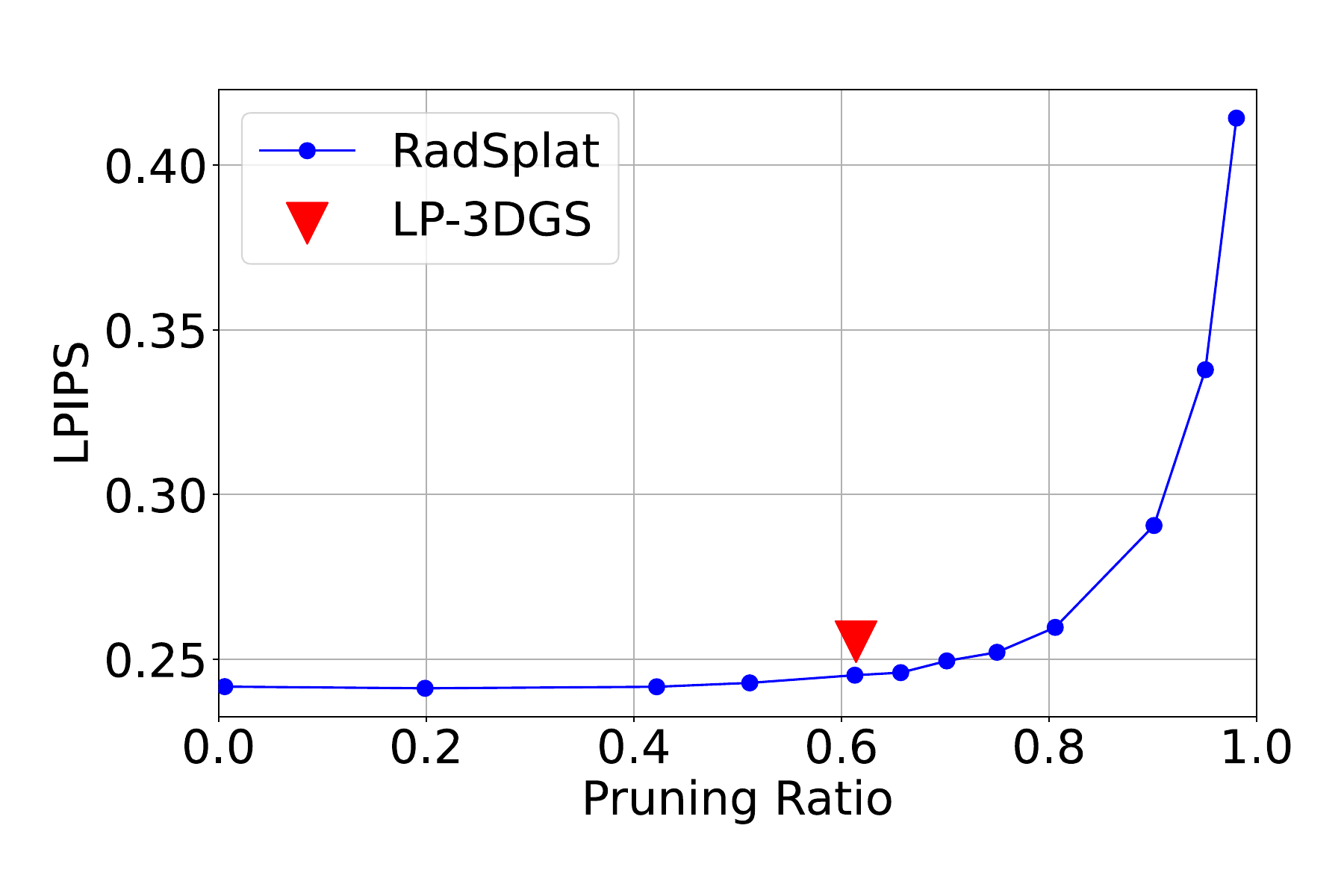}
    \end{subfigure}
    \end{minipage}

    \begin{minipage}[b]{0.05\textwidth}
        \rotatebox[origin=t, y=2.8cm]{90}{\textbf{Flowers}}
    \end{minipage}
    \begin{minipage}[b]{0.90\textwidth}
    \rotatebox[origin=t, y=1.3cm]{90}{\textbf{RadSplat}}
    \begin{subfigure}[b]{0.31\textwidth}
        \includegraphics[width=\textwidth]{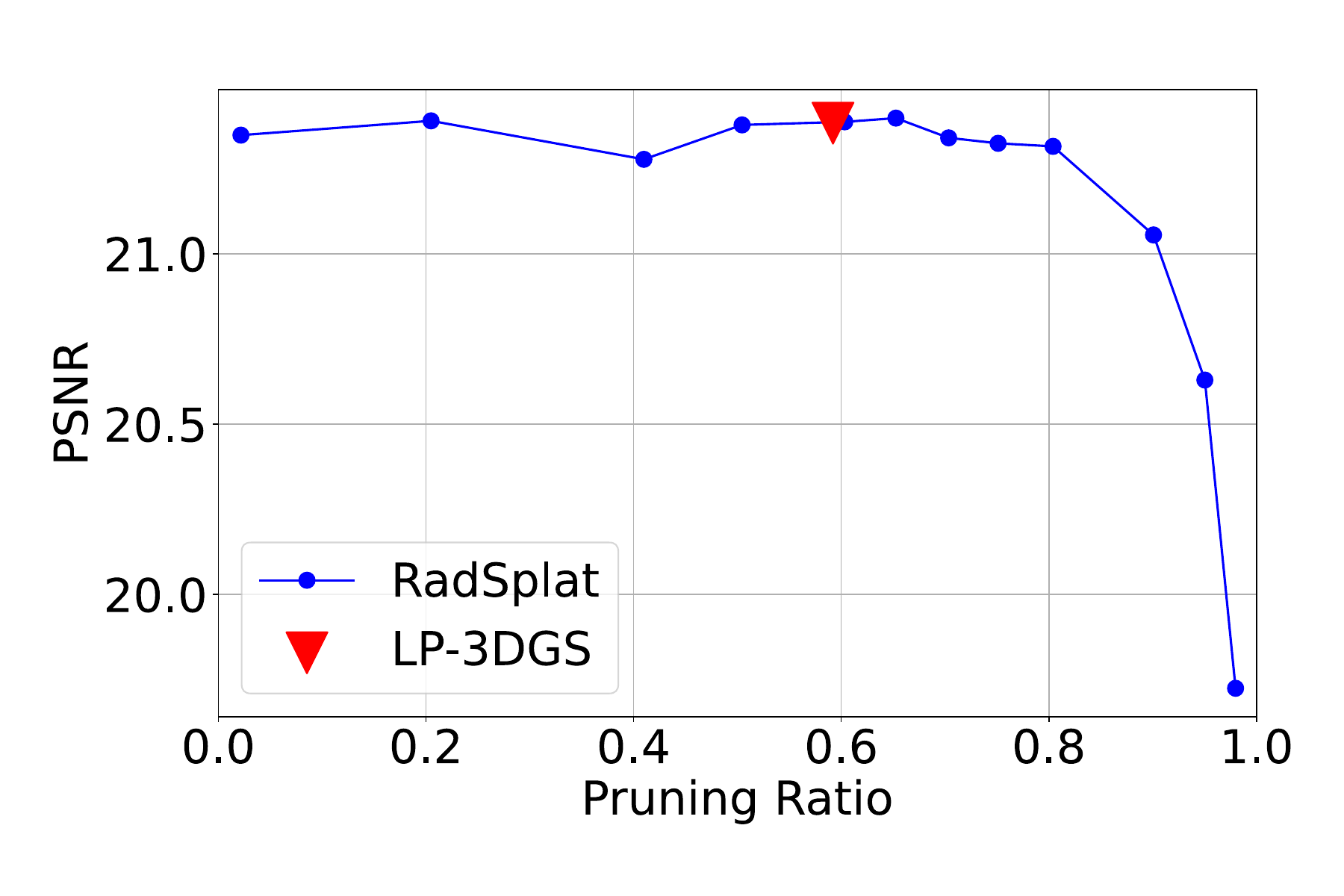}
    \end{subfigure}
    \begin{subfigure}[b]{0.31\textwidth}
        \includegraphics[width=\textwidth]{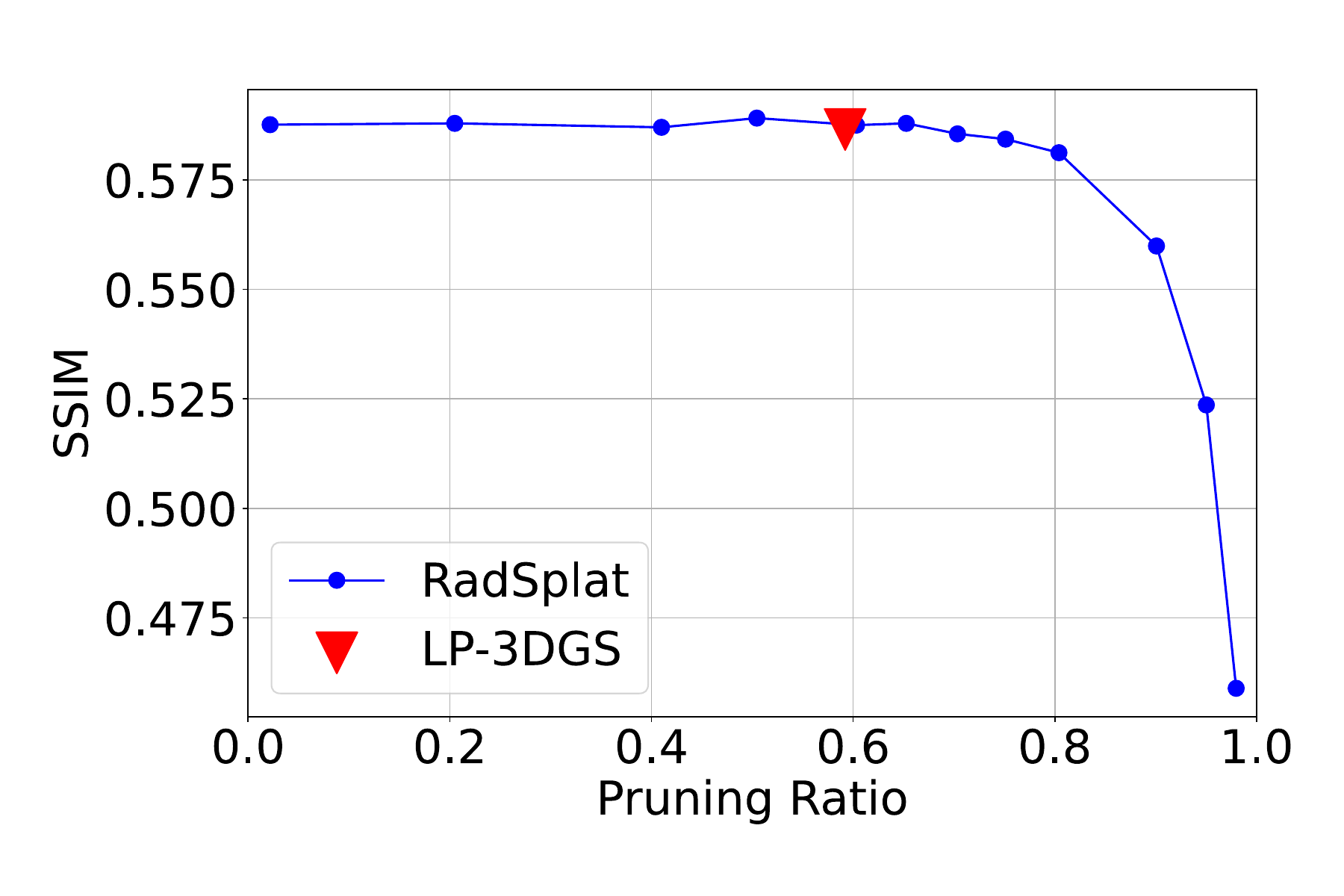}
    \end{subfigure}
    \begin{subfigure}[b]{0.31\textwidth}
        \includegraphics[width=\textwidth]{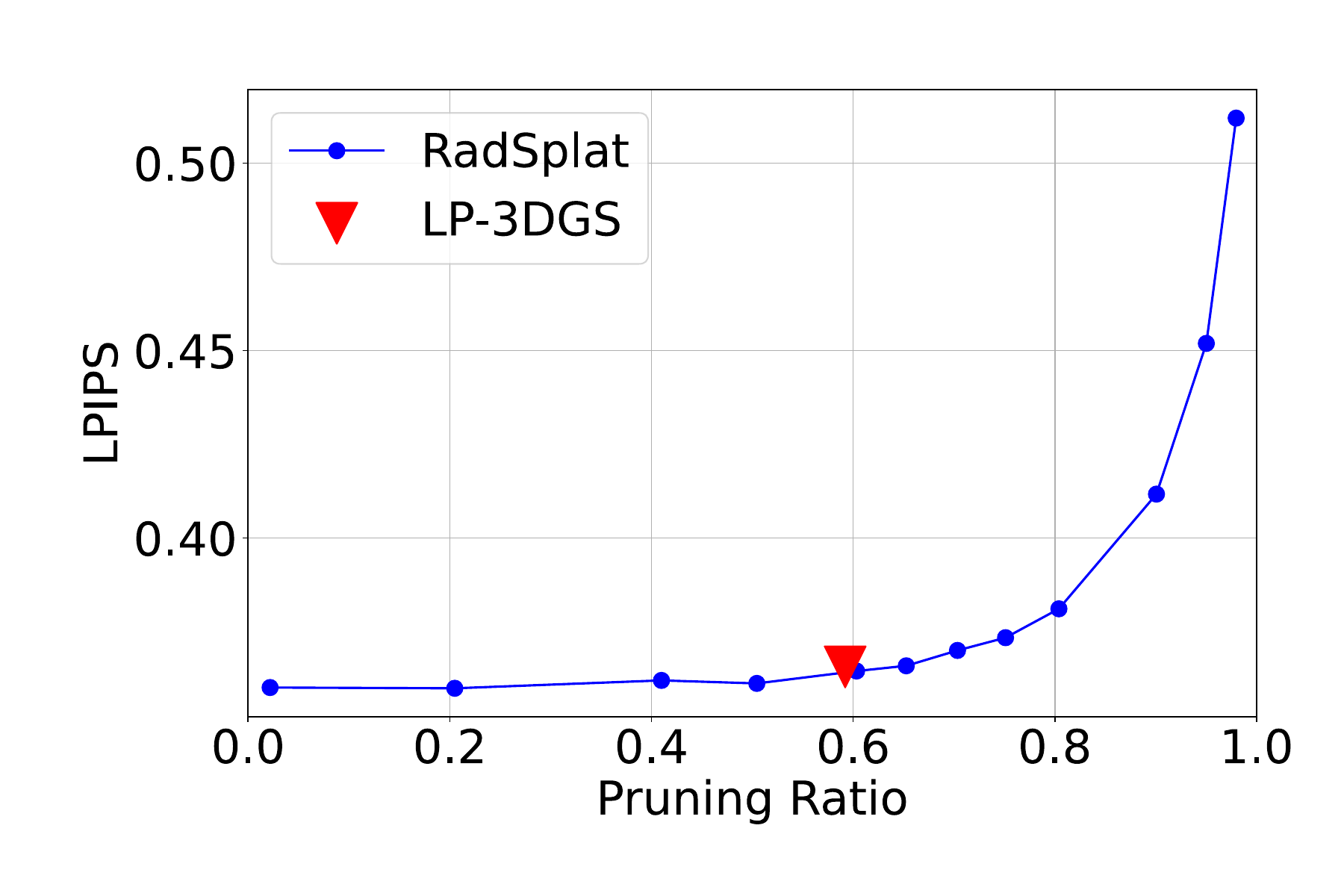}
    \end{subfigure}

    \rotatebox[origin=t, y=1.4cm]{90}{\textbf{Mini-Splatting}}
    \begin{subfigure}[b]{0.31\textwidth}
        \includegraphics[width=\textwidth]{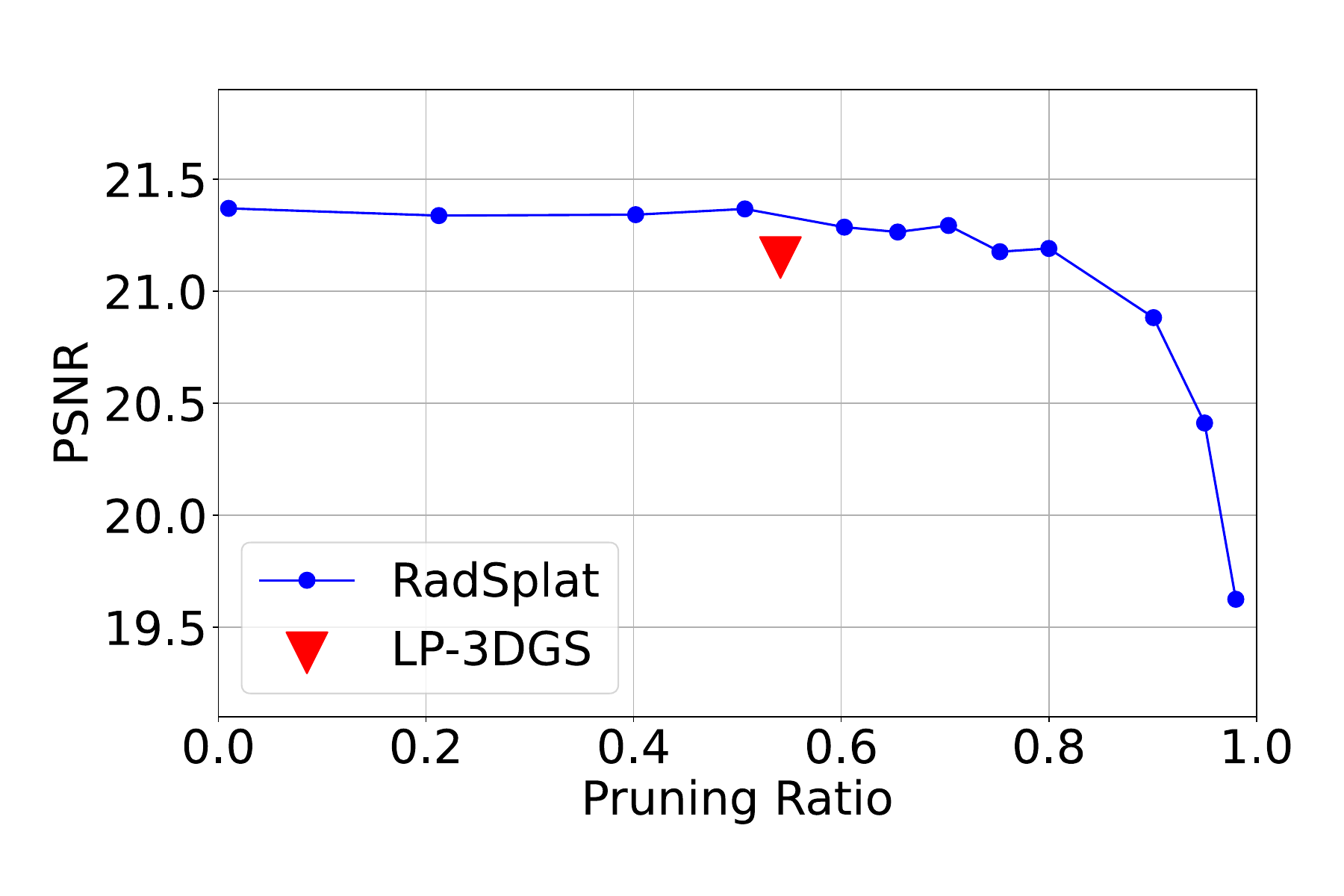}
    \end{subfigure}
    \begin{subfigure}[b]{0.31\textwidth}
        \includegraphics[width=\textwidth]{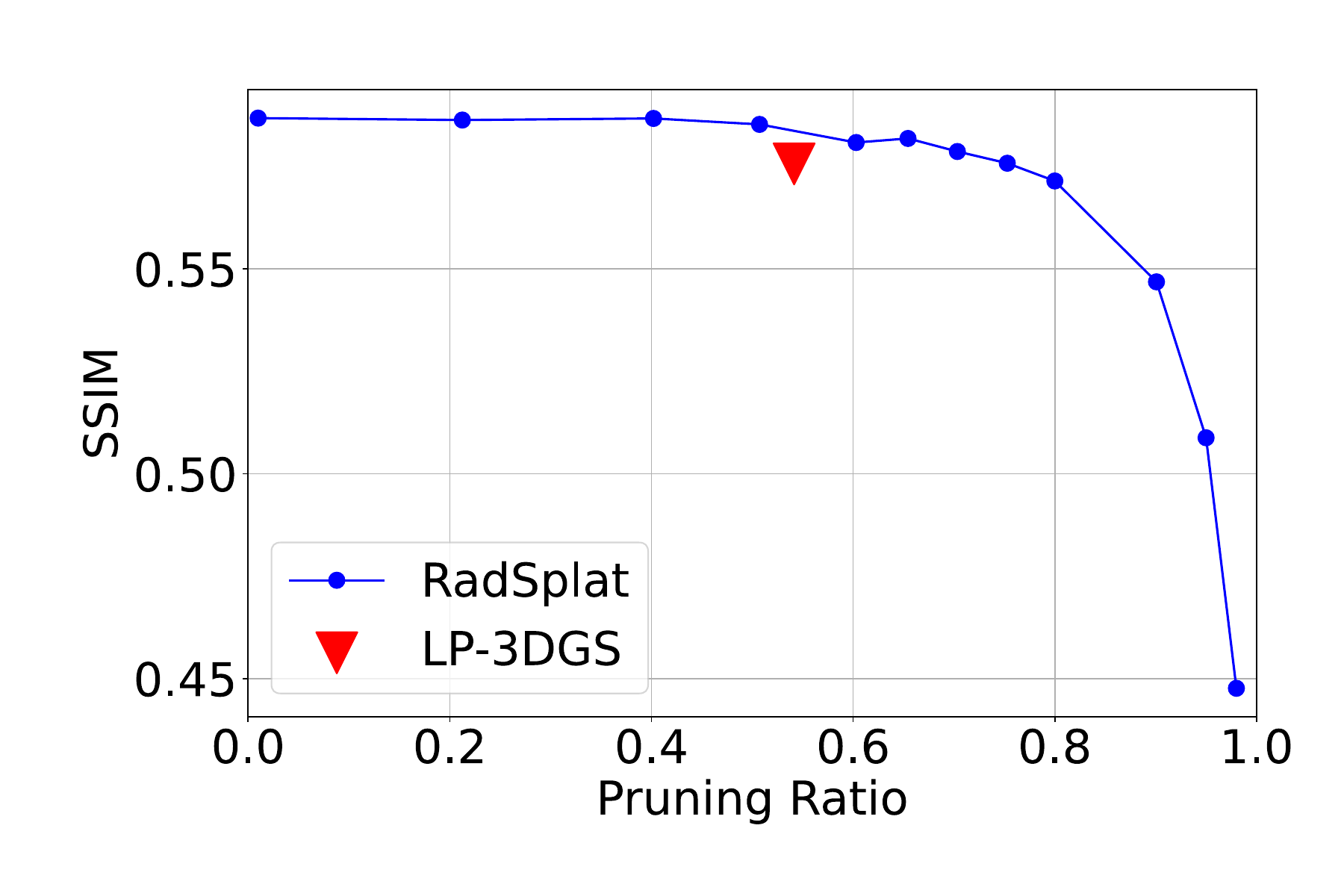}
    \end{subfigure}
    \begin{subfigure}[b]{0.31\textwidth}
        \includegraphics[width=\textwidth]{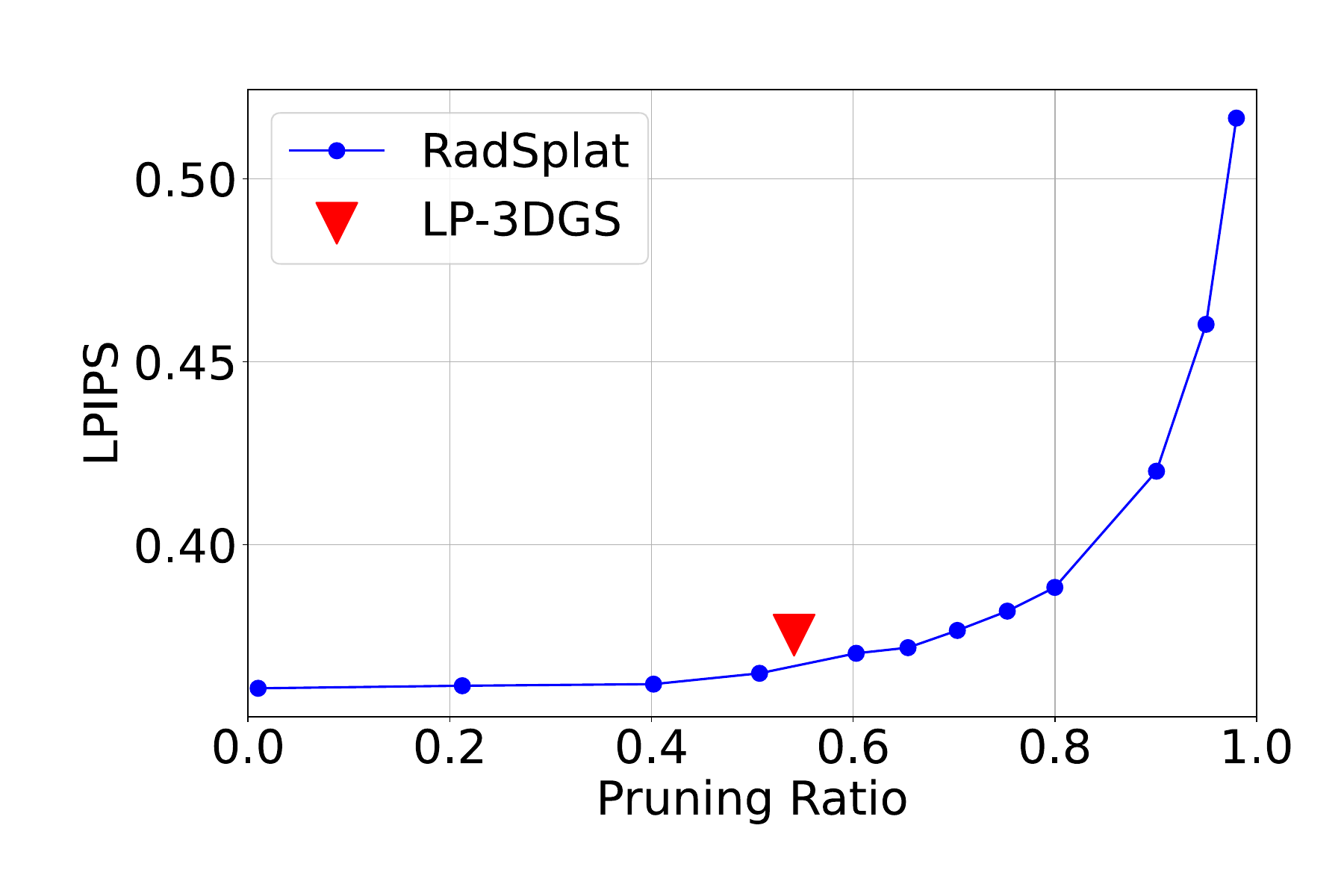}
    \end{subfigure}
    \end{minipage}

    \begin{minipage}[b]{0.05\textwidth}
        \rotatebox[origin=t, y=2.8cm]{90}{\textbf{Treehill}}
    \end{minipage}
    \begin{minipage}[b]{0.90\textwidth}
    \rotatebox[origin=t, y=1.3cm]{90}{\textbf{RadSplat}}
    \begin{subfigure}[b]{0.31\textwidth}
        \includegraphics[width=\textwidth]{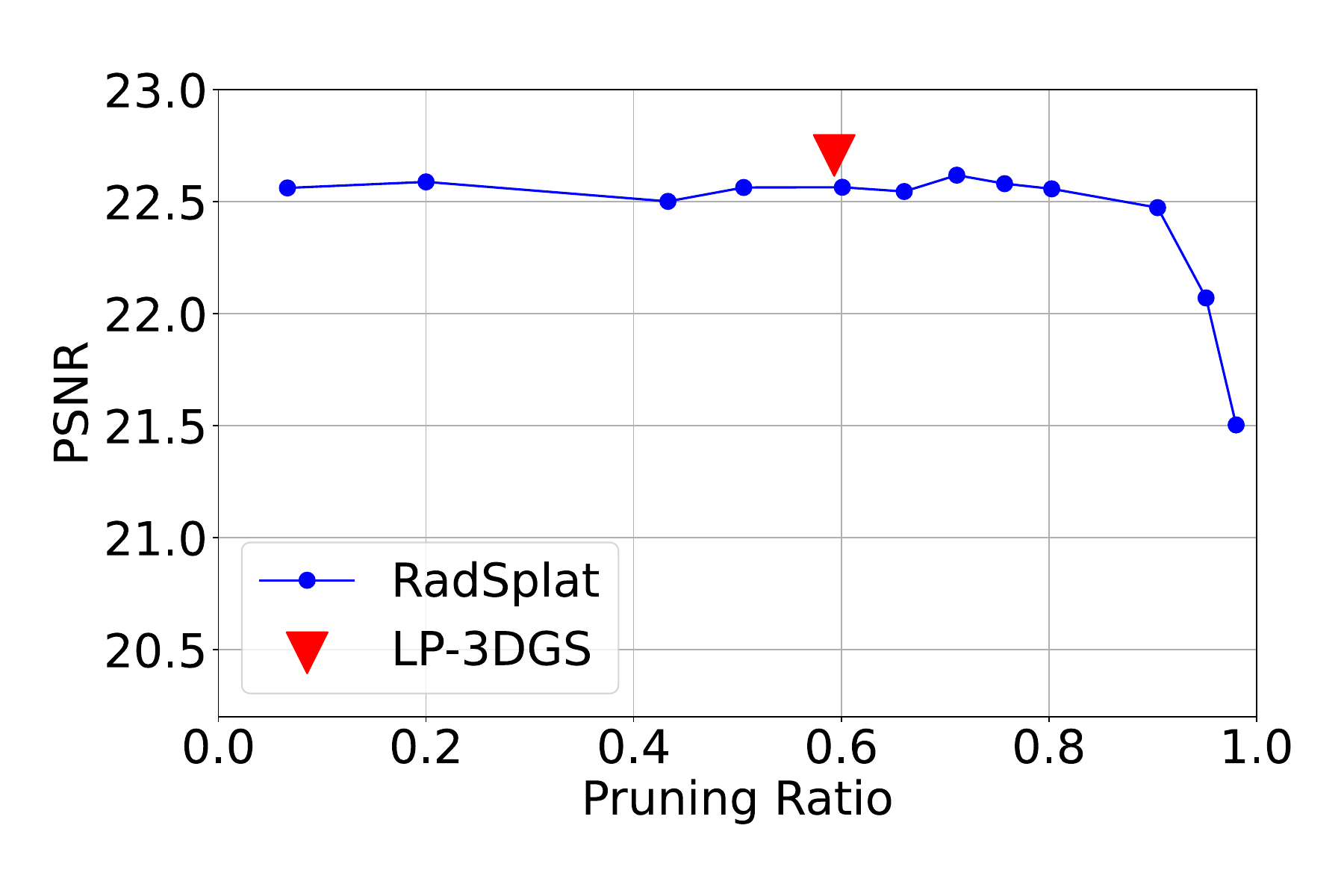}
    \end{subfigure}
    \begin{subfigure}[b]{0.31\textwidth}
        \includegraphics[width=\textwidth]{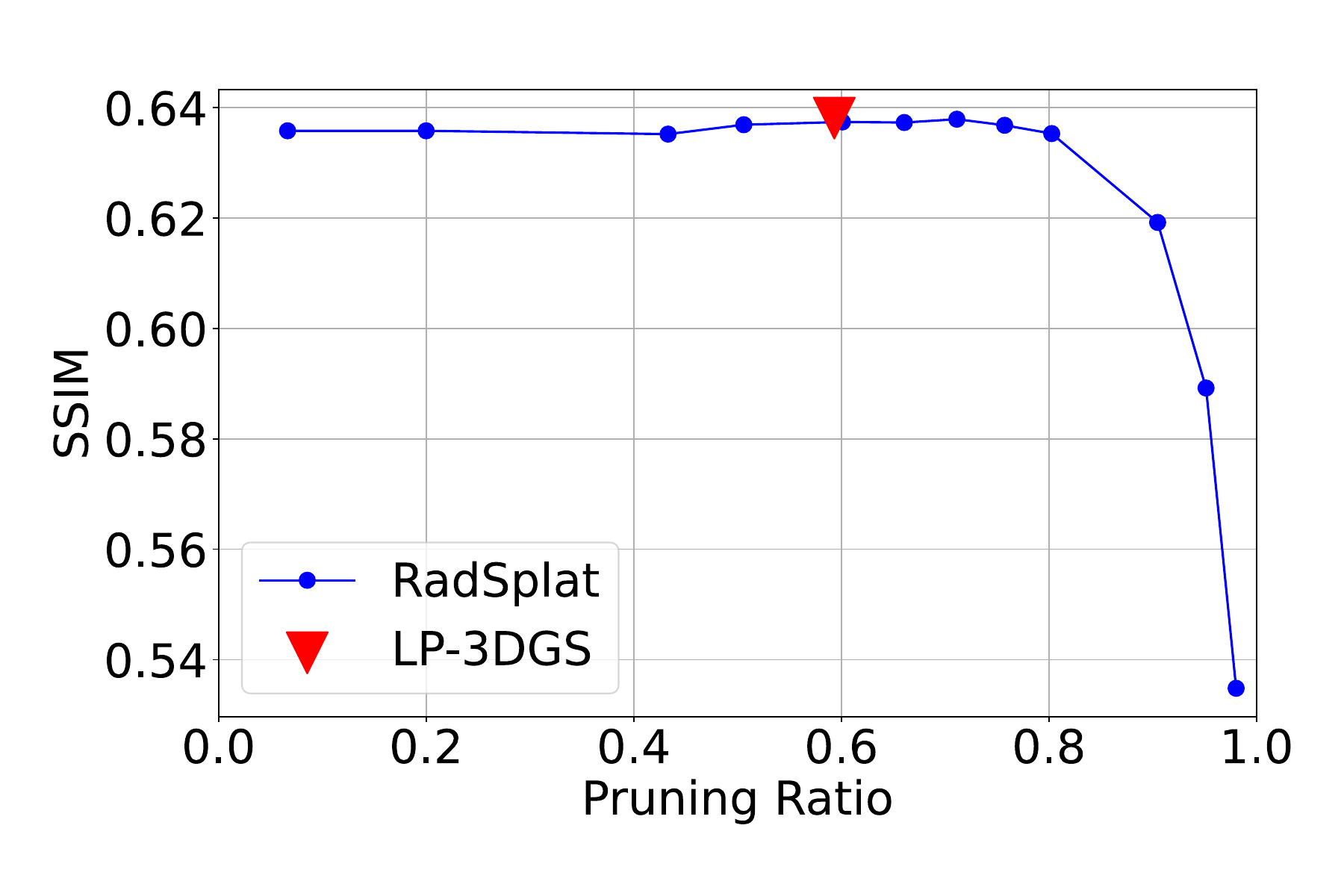}
    \end{subfigure}
    \begin{subfigure}[b]{0.31\textwidth}
        \includegraphics[width=\textwidth]{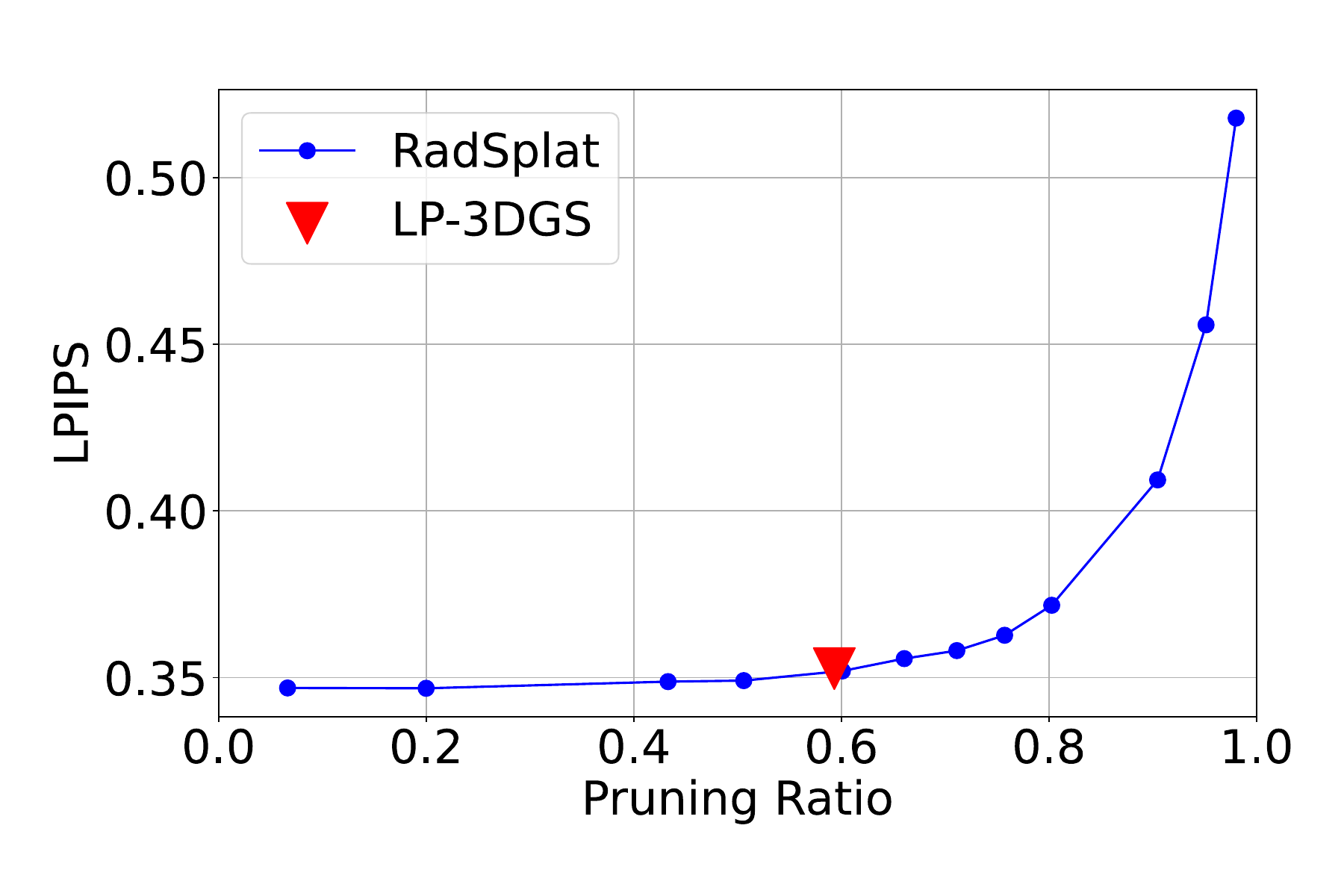}
    \end{subfigure}

    \rotatebox[origin=t, y=1.4cm]{90}{\textbf{Mini-Splatting}}
    \begin{subfigure}[b]{0.31\textwidth}
        \includegraphics[width=\textwidth]{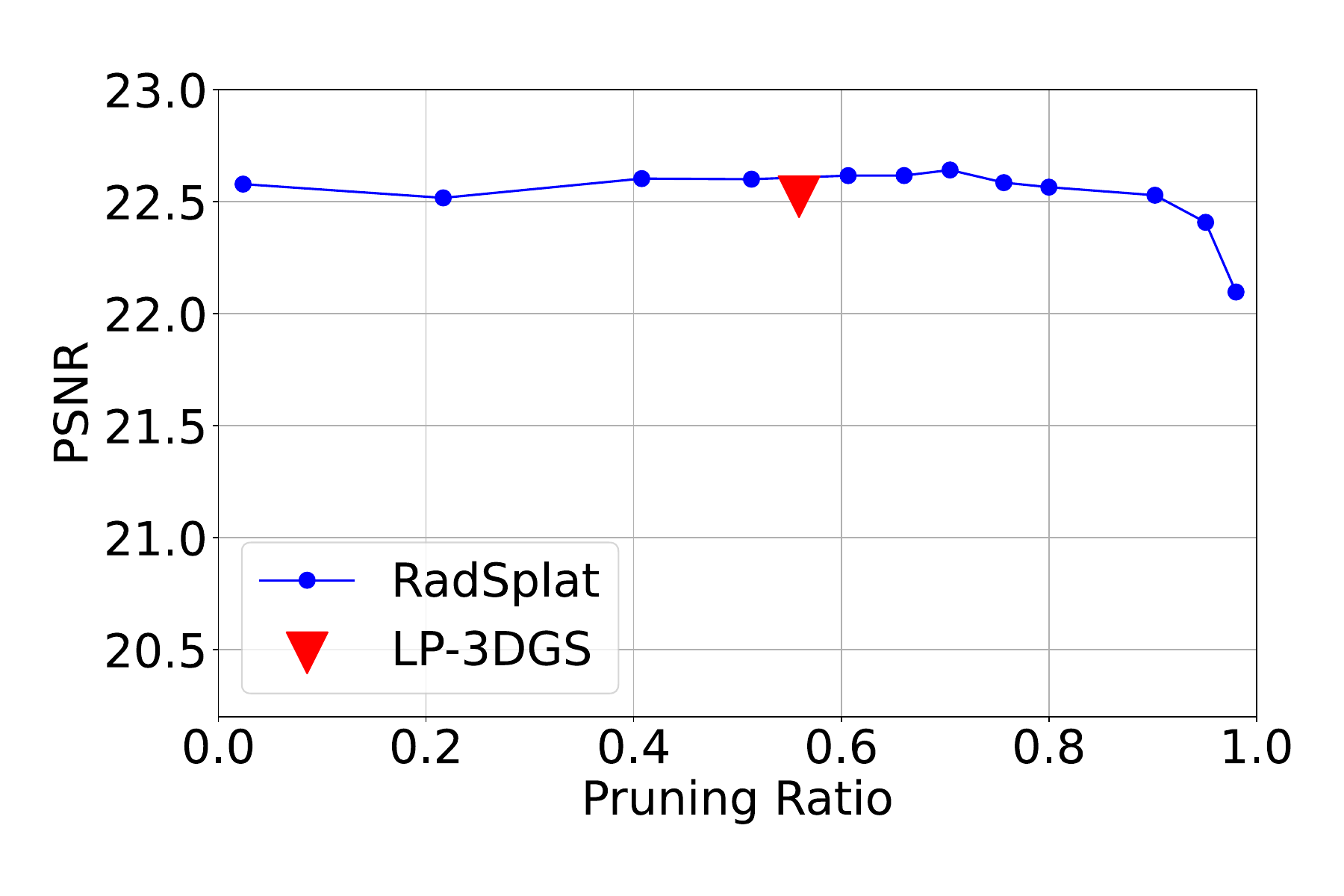}
    \end{subfigure}
    \begin{subfigure}[b]{0.31\textwidth}
        \includegraphics[width=\textwidth]{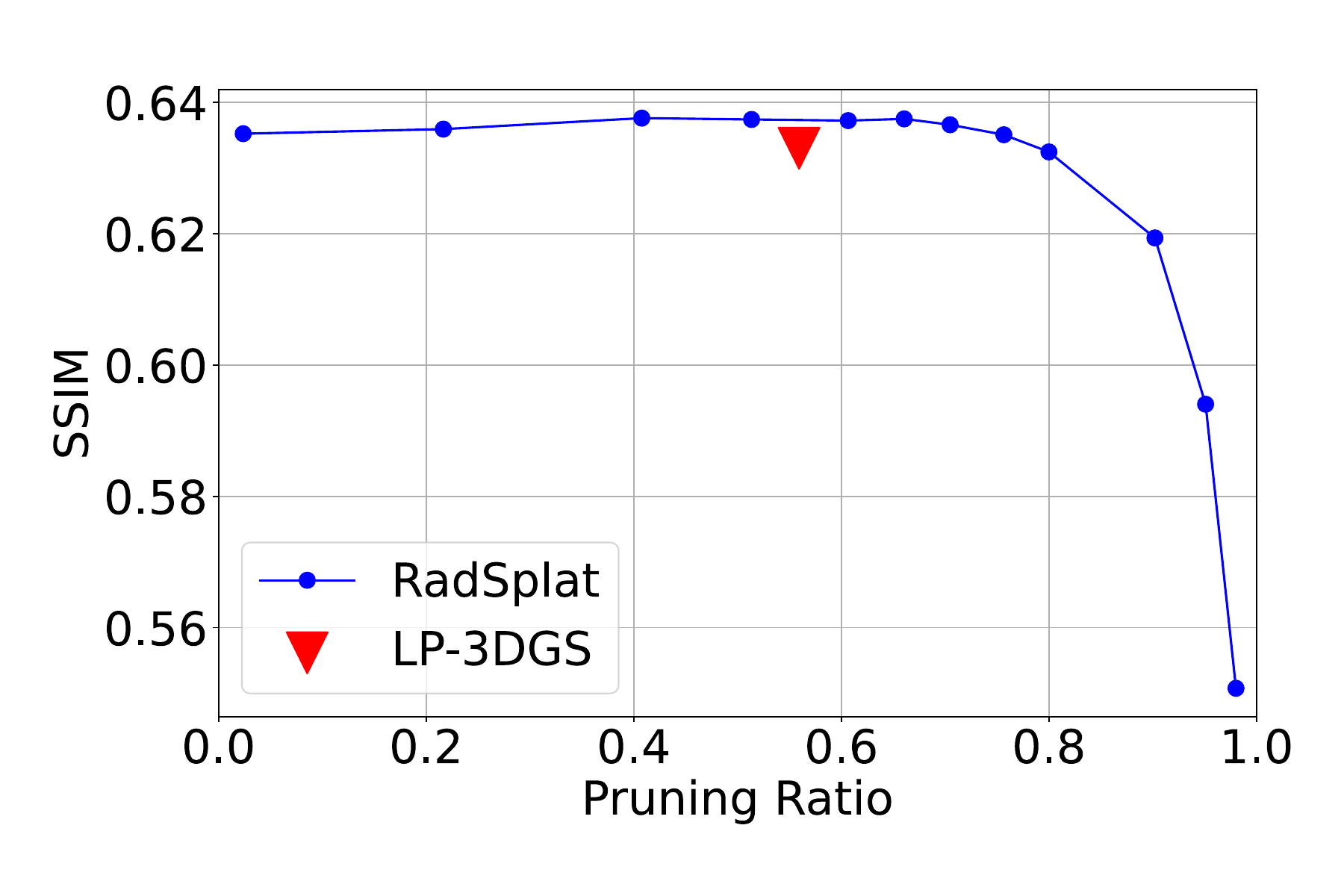}
    \end{subfigure}
    \begin{subfigure}[b]{0.31\textwidth}
        \includegraphics[width=\textwidth]{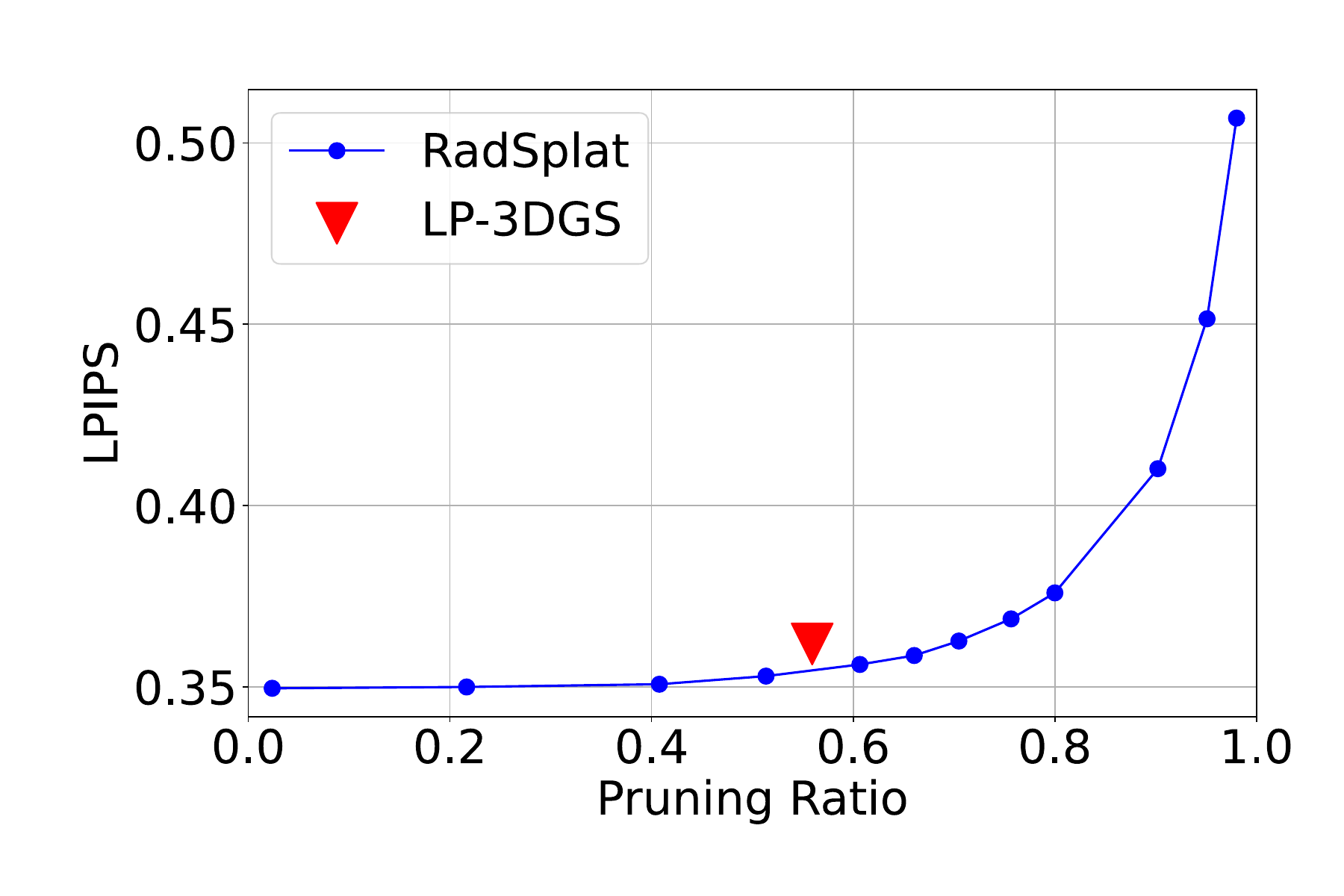}
    \end{subfigure}
    \end{minipage}

    \caption{The performance changes with the pruning ratio in different scenes}
    \label{fig:mip360curve}
\end{figure}

\clearpage
\subsection{Experiment Results on NeRF Synthetic Dataset}

\begin{table}[htb]
\caption{The quantitative results on NeRF Synthetic Dataset}
\label{tab:quantitative_result_nerf_synthetic}
\begin{adjustbox}{max width=\linewidth}
\begin{tabular}{cccccccccc}
    \toprule
    Scene           & Chair       & Drums        & Ficus       & Hotdog       & Lego    & Materials   & Mic  & Ship  & AVG \\ 
    \midrule
    Baseline PSNR $\uparrow$   & 35.546  & 26.276 & 35.480  & 38.081  & 36.012 & 30.502 & 36.795 & 31.688  & 33.798  \\
    LP-3DGS (RadSplat Score)   & 35.496  & 26.221 & 35.442  & 37.976  & 35.990 & 30.374 & 36.589 & 31.584  & 33.709 \\
    LP-3DGS (Mini-Splatting Score) & 35.419  & 26.102 & 35.354  & 37.728  & 35.769 & 29.883 & 36.337 & 31.375  &  33.496 \\ \hline
    Baseline SSIM $\uparrow$   & 0.9877  & 0.9548 & 0.9870  & 0.9854  & 0.9825 & 0.9604 & 0.9926 & 0.9062  & 0.9696  \\
    LP-3DGS (RadSplat Score)   & 0.9878  & 0.9547 & 0.9867  & 0.9854  & 0.9825 & 0.9598 & 0.9924 & 0.9061  & 0.9694 \\
    LP-3DGS (Mini-Splatting Score) & 0.9874  & 0.9358 & 0.9867  & 0.9846  & 0.9817 & 0.9566 & 0.9919 & 0.9034  & 0.966 \\ \hline
    Baseline LPIPS $\downarrow$& 0.01046  & 0.03657 & 0.01775  & 0.01977  & 0.0161 & 0.03671 & 0.00635 & 0.1058  &0.03119\\
    LP-3DGS (RadSplat Score)   & 0.01091  & 0.03723 & 0.01213  & 0.02079  & 0.01675 & 0.03817 & 0.00680 & 0.1083  & 0.03139  \\
    LP-3DGS (Mini-Splatting Score) & 0.0111  & 0.03876 & 0.01217  & 0.02211  & 0.018 & 0.04323 & 0.00749 & 0.1151 & 0.0335  \\ \hline
    RadSplat Score pruning ratio       & 0.77      & 0.76   & 0.84    & 0.68    & 0.65   & 0.61   & 0.78   & 0.60  &   0.71     \\
    Mini-Splatting Score pruning ratio  & 0.63   & 0.65   & 0.65    & 0.58    & 0.58   & 0.80   & 0.60  & 0.50  &  0.62  \\
    \bottomrule
    \end{tabular}
\end{adjustbox}
\end{table}

\subsection{Experiment Results on Truck \& Train Scenes}
\begin{table}[htb]
\caption{The quantitative results on Tanks \& Temples Dataset}
\label{tab:quantitative_result_tt}
\begin{adjustbox}{max width=\linewidth}
\begin{tabular}{cccc}
    \toprule
    Scene           & Truck       & Train     & AVG \\ 
    \midrule
    Baseline PSNR $\uparrow$   & 25.263  & 22.025 &  23.644 \\
    LP-3DGS (RadSplat Score)   & 25.376  & 21.822 & 23.599   \\
    LP-3DGS (Mini-Splatting Score) & 25.152  & 21.675 & 23.414 \\ \hline
    Baseline SSIM $\uparrow$   & 0.8778  & 0.8118 & 0.8448   \\
    LP-3DGS (RadSplat Score)   & 0.8768  & 0.8072 & 0.8420  \\
    LP-3DGS (Mini-Splatting Score) & 0.8724  & 0.7963 & 0.8344  \\ \hline
    Baseline LPIPS $\downarrow$& 0.1482  & 0.2083 &  0.1783  \\
    LP-3DGS (RadSplat Score)   & 0.1541  & 0.2217 & 0.1879   \\
    LP-3DGS (Mini-Splatting Score) & 0.162  & 0.2343 & 0.1982 \\ \hline
    RadSplat Score pruning ratio       & 0.72      & 0.63   & 0.68    \\
    Mini-Splatting Score pruning ratio  & 0.65  & 0.57   & 0.61 \\
    \bottomrule
    \end{tabular}
\end{adjustbox}
\end{table}

\end{document}